\documentclass[journal=jacsat,manuscript=article]{achemso}

\usepackage[version=3]{mhchem}
\usepackage{multirow}
\usepackage{array}
\usepackage{tabularx} 
\usepackage{seqsplit} 
\usepackage{amsfonts}
\usepackage{booktabs}
\usepackage[utf8]{inputenc}
\usepackage{textcomp}
\DeclareUnicodeCharacter{2082}{\textsubscript{2}}
\DeclareUnicodeCharacter{0306}{\u{}}

\author{Srivathsan Badrinarayanan}
\affiliation[cheme]
{Department of Chemical Engineering, Carnegie Mellon University, 15213, USA}

\author{Rishikesh Magar}
\affiliation[meche]
{Department of Mechanical Engineering, Carnegie Mellon University, 15213, USA}

\author{Akshay Antony}
\affiliation[meche]
{Department of Mechanical Engineering, Carnegie Mellon University, 15213, USA}

\author{Radheesh Sharma Meda}
\affiliation[cheme]
{Department of Chemical Engineering, Carnegie Mellon University, 15213, USA}

\author{Amir Barati Farimani}
\email{barati@cmu.edu}
\affiliation[meche]
{Department of Mechanical Engineering, Carnegie Mellon University, 15213, USA}
\alsoaffiliation[biomed]
{Department of Biomedical Engineering, Carnegie Mellon University, 15213, USA}
\alsoaffiliation[cheme]
{Department of Chemical Engineering, Carnegie Mellon University, 15213, USA}
\alsoaffiliation[mld]
{Machine Learning Department, Carnegie Mellon University, 15213, USA}

\title[An \textsf{achemso} demo]
    {MOFGPT: Generative Design of Metal-Organic Frameworks using Language Models}

\abbreviations{MOF, GPT, RL, ML}

\begin{document}

\begin{tocentry}
    \centering
    \includegraphics[width=9cm, height=3.5cm]{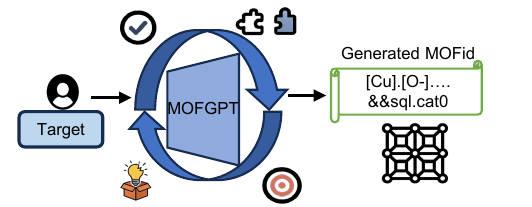}
\end{tocentry}

\begin{abstract}
The discovery of Metal-Organic Frameworks (MOFs) with application-specific properties remains a central challenge in materials chemistry, owing to the immense size and complexity of their structural design space. Conventional computational screening techniques such as molecular simulations and density functional theory (DFT), while accurate, are computationally prohibitive at scale. Machine learning offers an exciting alternative by leveraging data-driven approaches to accelerate materials discovery. The complexity of MOFs, with their extended periodic structures and diverse topologies, creates both opportunities and challenges for generative modeling approaches. To address these challenges, we present a reinforcement learning-enhanced, transformer-based framework for the de novo design of MOFs. Central to our approach is MOFid, a chemically-informed string representation encoding both connectivity and topology, enabling scalable generative modeling. Our pipeline comprises three components: (1) a generative GPT model trained on MOFid sequences, (2) MOFormer, a transformer-based property predictor, and (3) a reinforcement learning (RL) module that optimizes generated candidates via property-guided reward functions. By integrating property feedback into sequence generation, our method drives the model toward synthesizable, topologically valid MOFs with desired functional attributes. This work demonstrates the potential of large language models, when coupled with reinforcement learning, to accelerate inverse design in reticular chemistry and unlock new frontiers in computational MOF discovery.

\end{abstract}

\section{Introduction}

Metal-Organic Frameworks (MOFs) are porous crystalline materials consisting of inorganic metal clusters linked with organic ligands, known for their exceptional thermal stability, high porosity, and diverse applications in gas storage\cite{getman2012review,boyd2019data}, water treatment\cite{cao2019water,dutta2024metal,sun2018rapid}, and catalysis\cite{https://doi.org/10.1002/adma.201601133, C4CS00032C, JIAO201943}. With over 100,000 reported MOFs\cite{doi:10.1021/acs.chemmater.7b00441} and countless hypothetical structures\cite{wilmer2012large}, their vast chemical space offers tremendous opportunities for tailored material design but presents significant challenges in identifying optimal candidates for specific applications\cite{doi:10.1021/acs.chemrev.0c00004}.

Traditional computational screening approaches using molecular simulations and density functional theory (DFT)\cite{C6SC01477A, https://doi.org/10.1002/jcc.25787} face scalability limitations due to their high computational cost\cite{NANDY20231585}. Data-driven methods\cite{karamad2020orbital, doi:10.1021/jacs.0c09105, islamov2023high, ROSEN2022100760, doi:10.1021/jacsau.4c00618} and language modeling\cite{doi:10.1021/jacs.2c11420,kang2025harnessing,bai2024evaluation, radford2018improving, brown2020language}, have emerged as promising alternatives for large scale screening of MOFs\cite{magar2022crystal, DEMIR2023215112, doi:10.1021/acs.jcim.1c00191, nandy2022mofsimplify, lim2024accelerating}. The strong performance of deep learning models in property prediction has motivated efforts to explore such models as possible generative tools to enable the design of novel structures that satisfy specific functional requirements\cite{xie2021crystal,  doi:10.1021/acs.jcim.3c02070, ock2024multimodal, doi:10.1021/acs.jpclett.3c02398,  doi:10.1021/acscatal.3c04956}. However, MOFs present unique challenges for generative models due to their complex structures, diverse topologies, and heterogeneous chemical compositions\cite{doi:10.1021/acscentsci.7b00197}.  Generative modeling approaches for molecular design have evolved over time\cite{https://doi.org/10.1002/minf.201700111, 298725, mendez2020novo, doi:10.1021/acs.jcim.8b00263, reiser2022graph, park2024generative, inizan2025agenticaidiscoverymetalorganic, fu2024mofdiff, D3TA06274K}, yet it remains challenging for MOFs due to large number of atoms, diverse topologies and the multiple degrees of freedom involved in their design\cite{kotsias2020direct, doi:10.1021/acs.jcim.9b00943, yao2021inverse, Che2016ModeRG, muenkler2023vaesbadreconstructingmolecular, moosavi2020understanding, PARK20242355}. These characteristics make coordinate-based or graph-based generative models difficult to scale and limit their ability to generalize across different constraints. In contrast, sequence-based representations of MOFs open the door to alternative modeling strategies that can bypass these structural complexities.

Transformer-based language models\cite{vaswani2017attention, radford2019language} offer a compelling alternative for processing MOF representations as sequential data. Building on the effectiveness of GPT-like models for molecular design using SMILES-based representations\cite{doi:10.1021/acs.jcim.1c00600}, we adopt specialized approaches to capture the characteristic features of MOFs. For the same, we leverage MOFid\cite{doi:10.1021/acs.cgd.9b01050} - a string-based encoding that captures both chemical composition through SMILES\cite{doi:10.1021/ci00057a005} and topology through standardized RCSR labels\cite{doi:10.1021/ar800124u}. However, generative modeling alone cannot guarantee that MOF candidates possess desirable properties. Methods like supervised finetuning, though proven efficient at generating structures, are not guaranteed to be target-specific\cite{doi:10.1126/sciadv.aax9324}. Drawing inspiration from the success of reinforcement learning in optimizing large language models for specific tasks\cite{deepseekai2025deepseekr1incentivizingreasoningcapability}, we adapt these principles to MOF generation, enabling targeted design of structures with desired properties. Reinforcement learning (RL) has proven effective in molecular design\cite{doi:10.1021/acs.jcim.7b00690, doi:10.1126/sciadv.aap7885, mazuz2023molecule, pan2022deep}, and we seek to adapt and replicate this success for MOFs, where both local chemistry and global topology must be optimized simultaneously\cite{park2024inverse}. Building on this foundation, we introduce a comprehensive RL framework that enables multi-parameter optimization, allowing for the simultaneous control of key structural and functional properties during generation.

\begin{figure}[htbp]
    \centering
    \includegraphics[width=\textwidth]{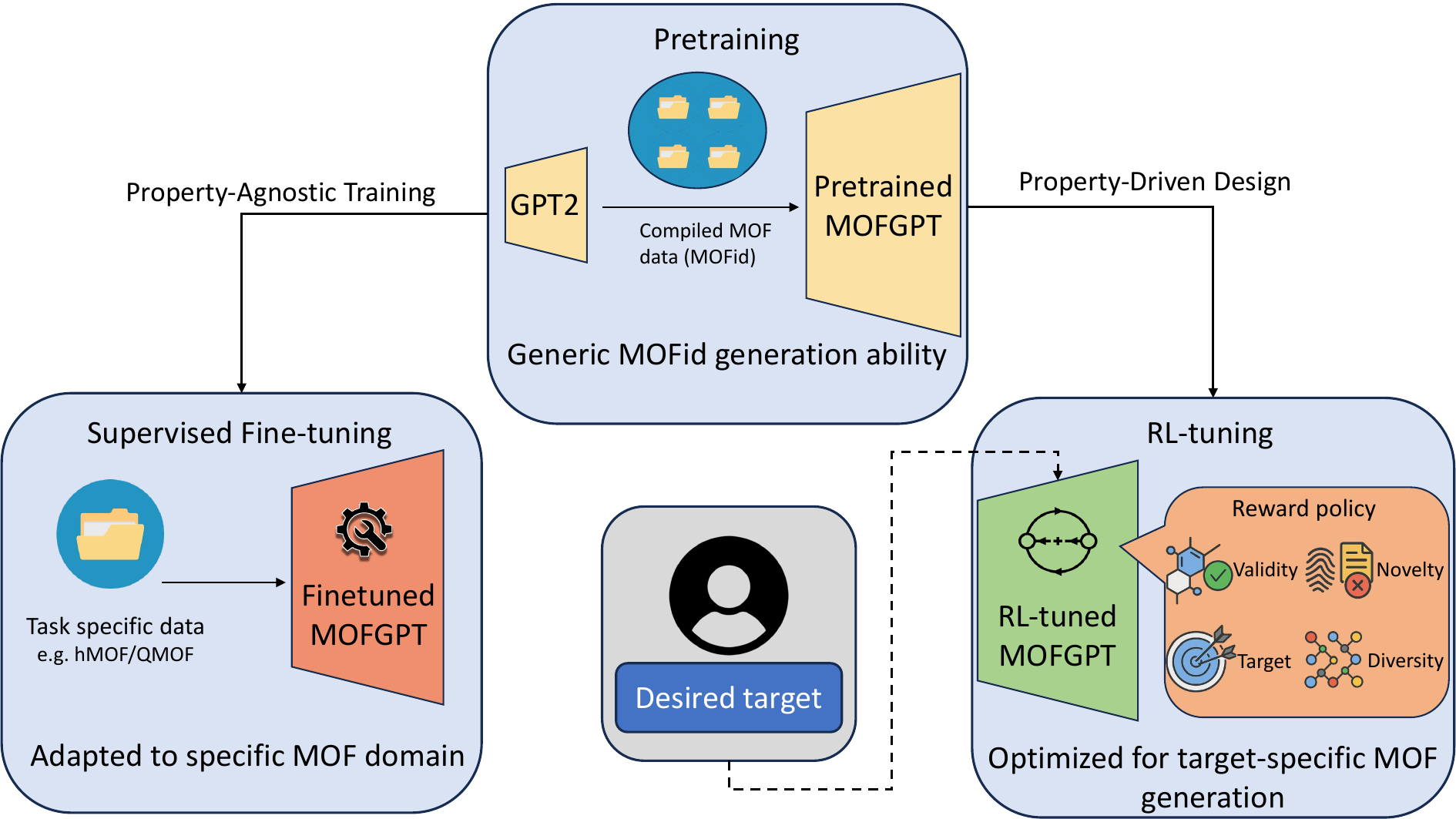}
     \caption{Overview of the MOFGPT framework. The pretrained MOFGPT model is first trained on a large, diverse corpus of MOF structures represented as MOFid strings to learn general MOF generation capabilities. This model is then finetuned using task-specific datasets (e.g., hMOF, QMOF) to adapt to particular domains or target properties. Alternatively, reinforcement learning (RL) is applied using a reward policy that optimizes for target-specific MOF generation with high validity, novelty, and diversity.}
    \label{fig:arch_1}
\end{figure}

In this work, we introduce a reinforcement learning-enhanced GPT-based generative framework for target-specific de novo MOF design as an alternative to existing supervised finetuning methods, as seen in Figure \ref{fig:arch_1}. Our RL approach integrates three key components: (1) a GPT-based MOF generator trained on MOFid sequences, (2) a transformer-based property predictor based on MOFormer\cite{doi:10.1021/jacs.2c11420}, and (3) a reward policy gradient-based RL mechanism that optimizes generated MOFs toward targeted properties. Our framework rewards novelty, validity, and structural diversity while guiding generation toward specific property targets, overcoming limitations of purely generative approaches not having property-based evaluation.

Unlike previous efforts focused on database searches, rule-based design, specific applications like CO$_2$ capture\cite{park2024inverse}, or even those that use a generic GPT framework for materials generation\cite{https://doi.org/10.1002/anie.202311983, kang2024chatmof}, our approach provides a target-specific framework for MOF discovery with domain-adapted architecture trained specifically on MOFs. By combining transformer-based generative modeling with RL and property-aware feedback, we present an advance in AI-driven MOF discovery, enabling scalable, targeted exploration of the vast MOF design space.

\section{Methods}

Our approach addresses the fundamental challenge of generating chemically valid MOFs with targeted properties through a three-stage computational pipeline that improves progressively: pretraining for general MOF generation capabilities, supervised fine-tuning (on property prediction) which guides MOF generation, and reinforcement learning for property-targeted optimization. This hierarchical approach ensures that the model first learns the basic "grammar" of the MOF structures through pretraining, before being guided toward specific functional objectives. A salient feature of our framework is the integration of reinforcement learning with transformer-based generative models for property-targeted MOF design. The novelty lies in our comprehensive RL setup that provides explicit control over functional objectives while maintaining chemical validity and structural diversity through a multi-component reward architecture specifically designed for MOF generation.

\subsection{Data Representation}
A critical component of our approach is the efficient representation of MOF structures in a format suitable for language models. We utilize the MOFid representation\cite{doi:10.1021/acs.cgd.9b01050}, which encodes both chemical composition and topology in a compact string format. MOFid consists of two primary components: the SMILES notation of secondary building units (SBUs)\cite{doi:10.1021/ci00057a005} and topology codes from the Reticular Chemistry Structure Resource (RCSR) database\cite{doi:10.1021/ar800124u}, connected by a separator token ("$\&\&$"):
\begin{equation}
\textrm{MOFid} = [organic\_components].[inorganic\_components] \&\& [topology].[catenation]
\end{equation}
This representation enables our model to generate complete MOF descriptions including both building blocks and topology in a single integrated framework. We implemented a specialized tokenizer that processes both chemical and topological components, with all sequences standardized to a maximum length of 512 tokens. Further details on tokenization and representation can be found in the Supporting Information.

Our study aggregates MOF structures from three comprehensive repositories: Boyd \& Woo\cite{boyd2019data}, quantum MOF (QMOF)\cite{ROSEN20211578, rosen2022high}, and hypothetical MOF (hMOF)\cite{wilmer2012large} datasets, creating a training set of 323,469 samples and a held-out test set of 81,260 samples. This combined dataset is used to pretrain our GPT framework. On top of this pretraining dataset, we have task-specific datasets for our finetuning and reinforcement learning approaches. For the same, we leverage two distinct property datasets: (1) the hMOF dataset for gas adsorption properties, which consists of gas adsorption values (in mol/kg) for CO$_2$ and CH$_4$ at various pressures, and (2) the QMOF dataset for electronic band gap values (in eV), thereby providing diverse objective functions for our optimization framework. We note that for an unbiased study, all the task-specific hMOF datasets have the same set of MOFids subject to adsorption of the two gases at different pressures. In our study specifically, we utilize 5 CH$_4$ adsorption datasets at pressures of 0.05, 0.5, 0.9, 1.0, and 5.0 bar, 5 CO$_2$ adsorption datasets at pressures of 0.01, 0.1, 0.5, 1.0, and 5.0 bar, and 1 electronic band gap dataset from QMOF. For our primary results demonstration, we selected representative conditions to showcase the framework's versatility across different physical regimes and property types. Additional datasets are utilized in supplementary analyses to validate the generalizability of our approach across the full range of adsorption conditions.

For gas adsorption properties, higher values indicate better performance; therefore the optimization objective is to predict materials with property values greater than or equal to the desired target, since increased uptake capacity is desirable for applications such as gas storage and separation. For electronic band gap optimization, lower values are typically targeted to improve electrical conductivity, which is particularly valuable for energy storage and electronic device applications - so the optimization objective is to generate materials with band gap property values lower than or equal to the desired target. 

\subsection{Model Architecture and Training Methodology}

Our MOFGPT framework utilizes a transformer-based architecture derived from GPT-2, adapted specifically for processing MOFid representations. The model consists of 12 transformer decoder layers with an embedding dimension of 768, 12 attention heads, and a feed-forward dimension of 3072. This configuration provides sufficient capacity to capture the complex relationships in MOF structures while remaining computationally tractable. The training methodology follows a three-stage approach progressively builds model capabilities:

\subsubsection{Stage 1: Pretraining for General MOF Generation}

We first pretrain the base language model using next-token prediction on our large corpus of MOF structures. The pretraining objective maximizes the log-likelihood of the next token given previous tokens:
\begin{equation}
\mathcal{L}_{\text{pretrain}} = -\sum_{t=1}^{T} \log P_\theta(x_t | x_{<t})
\end{equation}

where $x_t$ is the token at position $t$, $x_{<t}$ represents all tokens before position $t$, and $P_\theta$ is the probability distribution over the vocabulary given by the model with parameters $\theta$. This stage enables the model to learn the fundamental patterns and constraints of MOF structures, including chemical bonding rules, coordination environments, and topological relationships. The pretraining phase is crucial for establishing a strong foundation that ensures generated structures remain chemically sensible throughout subsequent optimization.

\subsubsection{Stage 2: Supervised Fine-tuning for Property Prediction}

After pretraining, we fine-tune the model for property prediction using supervised learning. We extend the base language model with a specialized regression head that processes the output embeddings from the transformer decoder through a series of feed-forward layers to extract property-relevant features. The fine-tuning objective minimizes the mean squared error between predicted and actual property values:
\begin{equation}
\mathcal{L}_{\text{finetune}} = \frac{1}{N} \sum_{i=1}^{N} (y_i - \hat{y}_i)^2
\end{equation}

where $y_i$ is the ground truth property value for the $i$-th MOF, $\hat{y}_i$ is the predicted value, and $N$ is the number of training examples. This stage is essential for two reasons: first, it allows the model to capture structure-property relationships, and second, it creates a property predictor that will serve as the "value function" in our reinforcement learning framework. The dual role of this fine-tuned model as both generator and property evaluator ensures consistency between generation and evaluation processes. Since fine-tuning does not guarantee target-specific generations, we explore RL as a strong and compelling alternative for our objective. 

\subsubsection{Stage 3: Reinforcement Learning Framework}

The reinforcement learning stage optimizes the pretrained model to generate MOFs with specific target properties. The RL framework consists of two main components: (1) Policy network, here the language model that generates possible outputs at each step, and (2) Value network, here the frozen property prediction model (MOFormer\cite{doi:10.1021/jacs.2c11420}) that is used to estimate the expected reward for a generated MOF sequence.

\begin{figure}[htbp]
\centering
\includegraphics[width=0.8\textwidth]{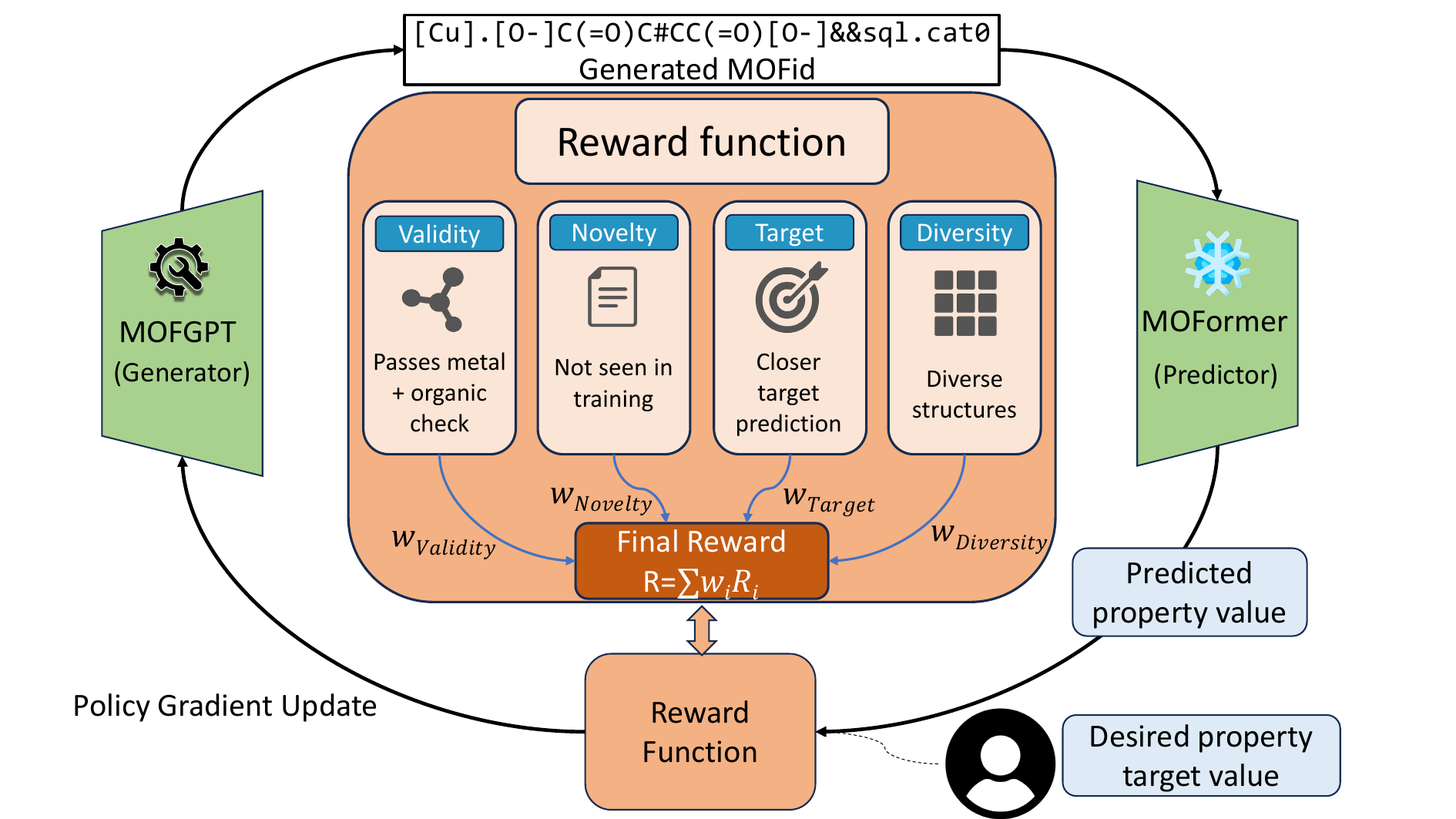}
\caption{Reinforcement learning framework for property-driven MOF generation with multi-objective reward function. The RL-tuned MOFGPT model generates MOFid candidates evaluated by MOFormer. The reward function integrates four components: (1) validity (chemical and MOF-specific constraints), (2) novelty (penalizing training duplicates), (3) target accuracy (rewarding desired property values), and (4) diversity (encouraging structural variety). The weighted final reward R = $\sum w_i R_i$ drives policy gradient updates to optimize the generation process.}
\label{fig:rl_framework}
\end{figure}

As illustrated in Figure \ref{fig:rl_framework}, the RL training loop consists of four main steps: (1) generating a batch of MOF structures with the generative model, (2) evaluating each structure using the frozen property predictor, (3) calculating policy gradients based on computed rewards, and (4) updating policy parameters accordingly. The RL objective is to maximize the expected reward:
\begin{equation}
J(\theta) = \mathbb{E}_{\tau \sim \pi_\theta} [R(\tau)]
\end{equation}
where $\pi_\theta$ is the policy defined by the model parameters $\theta$ and $R(\tau)$ is our multi-component reward function. The policy gradient update follows the REINFORCE with Baseline algorithm\cite{NIPS1999_464d828b}, which aims to maximize the expected reward by adjusting the policy parameters. The gradient of the expected reward $J(\theta)$ with respect to the policy parameters $\theta$ is approximated as:
\begin{equation}
\nabla_\theta J(\theta) \approx \frac{1}{N}\sum_{i=1}^{N}\sum_{t=1}^{T_i} \nabla_\theta \log \pi_\theta(a_t^i|s_t^i) \cdot (R_i - \mu) \cdot \gamma^{t-1}
\end{equation}
where $\pi_\theta$ is the policy, $a_t^i$ is the action (token) selected at time $t$ for sequence $i$, $s_t^i$ is the corresponding state (partial sequence), $R_i$ is the reward for the complete sequence, $\mu$ is the baseline (typically the mean reward), and $\gamma$ is a discount factor.

The complete set of model configurations (for all three stages) and training hyperparameters are detailed in the Supporting Information.

\subsection{Reward Function Design}

The effectiveness of our reinforcement learning approach depends critically on the reward function design. The design of our multi-component reward function addresses fundamental challenges unique to MOF generative modeling that distinguish it from small molecule generation. While small organic molecules can be validated through established chemical rules, MOFs require additional consideration of both local coordination chemistry and global topological consistency. This challenge necessitated our comprehensive reward architecture that balances multiple competing objectives while maintaining focus on property targeting. As illustrated in the center of Figure \ref{fig:rl_framework}, our multi-component reward function integrates four key objectives with carefully tuned weights:
\begin{equation}
R_{\text{total}}(m) = \beta_{\text{target}} \cdot R_{\text{target}}(m) + \alpha_n \cdot R_{\text{novelty}}(m) + \alpha_v \cdot R_{\text{validity}}(m) + \alpha_d \cdot R_{\text{diversity}}(m)
\end{equation}
where $\beta_{\text{target}}$, $\alpha_n$, $\alpha_v$, and $\alpha_d$ are experimentally-chosen weighting coefficients that balance property optimization, exploration, chemical feasibility, and structural variety, respectively (specific values provided in Supporting Information). The total reward calculation follows an adaptive two-tier architecture that differentiates between promising and poor-performing structures. High-performing structures (selected via top-K mechanism explained later in this section) receive full multi-component rewards including target proximity, validity bonuses, novelty incentives, and diversity encouragement. Lower-performing structures receive reduced target signals to focus learning on promising candidates. This hybrid structure ensures target property optimization dominates the reward signal while maintaining exploration across the full chemical space. 

\subsubsection{Core Reward Components}

\textbf{Target Property Reward:} The target property reward guides generation toward structures with desired functional properties, serving as the primary optimization objective:
\begin{equation}
R_{\text{target}}(m) = \sum_{i=1}^{k} w_i \cdot R_{\text{proximity}}(\hat{p}_i(m), T_i)
\end{equation}
where $\hat{p}_i(m)$ is the predicted property value, $T_i$ is the target value, and $w_i$ is the importance weight. The proximity reward implements a tiered structure with distinct achievement levels: excellent performance (within 5\% of target, reward = 15.0), very good (5-10\%, reward = 12.0), good (10-20\%, reward = 8.0), moderate (20-50\%, reward = 4.0), and poor but improving performance with linearly decreasing rewards. This tiered reward structure addresses the challenge that simple distance-based metrics fail to distinguish between meaningful progress toward targets versus random fluctuations around poor performance levels. Additionally, we also provide direction-specific bonuses to address the asymmetric nature of property optimization where the desired direction varies by application. For gas adsorption properties, values meeting or exceeding targets indicate better performance, while for electronic properties like band gaps, values at or below targets are preferred for better conductivity. 

\textbf{Novelty Reward:} The novelty reward encourages exploration of new chemical space by identifying structures not present in the training data:
\begin{equation}
R_{\text{novelty}}(m) =
\begin{cases}
1 & \text{if } m \notin \mathcal{D}_{\text{train}} \\
0 & \text{otherwise}
\end{cases}
\end{equation}
where $\mathcal{D}_{\text{train}}$ represents the set of MOF structures in the training dataset. This binary reward prevents overfitting to known MOF structures while maintaining computational efficiency through exact string matching of MOFid representations.

\textbf{Validity Reward:} The validity reward ensures chemical feasibility by verifying that generated structures satisfy established chemical and topological constraints:
\begin{equation}
R_{\text{validity}}(m) =
\begin{cases}
1 & \text{if valid according to validation procedure} \\
0 & \text{otherwise}
\end{cases}
\end{equation}
Our validation procedure encompasses multiple checks: SMILES syntax validation using RDKit with metal atom substitution, metal node presence verification, structural balance confirmation requiring both organic and inorganic components, topology validity against known RCSR database entries when topology tokens are present, and coordination number validation for metal centers.

\textbf{Diversity Reward:} The diversity reward prevents mode collapse by encouraging structural variety within generated batches through multiple complementary metrics:
\begin{equation}
R_{\text{diversity}}(m, \mathcal{B}, \mathcal{H}) = w_b \cdot S_{\text{batch}}(m, \mathcal{B}) + w_n \cdot S_{\text{ngram}}(m, \mathcal{B}) + w_h \cdot S_{\text{history}}(m, \mathcal{H}) + w_c \cdot S_{\text{composition}}(m)
\end{equation}
where $\mathcal{B}$ is the current batch, $\mathcal{H}$ is the generation history, and the weights are $w_b = 0.30$ (batch diversity), $w_n = 0.25$ (n-gram pattern diversity), $w_h = 0.35$ (historical uniqueness), and $w_c = 0.10$ (compositional diversity). This multi-faceted approach captures different aspects of structural variety to ensure broad exploration of the chemical space. As part of this reward, we maintain a rolling memory of 500 previously generated structures ($\mathcal{H}$) that prevents cycling between high-reward candidates and ensures continued exploration of new chemical space throughout the optimization process.

\subsubsection{Advanced Reward Mechanisms}

Beyond the core components, our reward system incorporates several mechanisms to improve training stability and exploration:

\textbf{Global Memory:} To prevent loss of high-performing structures discovered during training, we maintain a global memory of the best discoveries across all epochs:
\begin{equation}
\mathcal{M}_{\text{global}} = \{(m_j, p_j, s_j, r_j)\}_{j=1}^{200}
\end{equation}
where each entry contains a MOF structure $m_j$, predicted properties $p_j$, target progress score $s_j$, and total reward $r_j$. The target progress score provides unified ranking across all objectives:
\begin{equation}
s_j = \sum_{i=1}^{k} w_i \times \text{Progress}_i(p_{j,i}, T_i)
\end{equation}
The progress function rewards both achievement and over-achievement in the desired optimization direction, ensuring structures exceeding targets receive bonuses proportional to their performance while maintaining minimum scores for all candidates. This global memory ensures that promising structures discovered at any point during training are retained and can guide future generations.

\textbf{Top-K Selection:} Early in training, most generated structures are invalid or far from targets. Our top-K implementation focuses learning on promising structures by selecting only the highest-performing candidates. Our implementation also becomes increasingly selective as training progresses (50\% early training, decreasing to 30\% later). This progressive selectivity is established due to the model's improving capability - early training requires broader sampling to establish basic chemical competency, while later training benefits from concentrated optimization of promising candidates. 

\textbf{Reward normalization:} Reward normalization becomes critical, as unnormalized rewards can prevent effective policy updates. Our reward normalization implementation applies scaling only when batch statistics indicate extreme outliers or excessive variance, preserving natural reward relationships while preventing training instability.

These mechanisms work together to address key challenges in MOF generation: balancing exploration of new structures with focused optimization toward targets, maintaining consistent performance across different property ranges, and preventing mode collapse while still converging to high-quality solutions. The complete mathematical formulations, implementation details, and parameter values (chosen through experimentation) for all reward components are provided in the Supporting Information.

\section{Results and Discussion}

\subsection{Fine-tuned Models Fail to Generate Valid MOFs}

A critical finding that motivates our entire RL approach is that fine-tuned models without RL optimization fail to generate any chemically valid MOF structures across all property domains and targets. This failure to produce viable chemical structures represents a limitation of supervised fine-tuning approaches and establishes the necessity of our reinforcement learning framework.

The fine-tuned model, despite successful property prediction during training, cannot translate this predictive capability into generation of chemically feasible MOF structures. While the model produces numerical property values, these correspond to structurally invalid MOFs that violate basic chemical rules, coordination constraints, and topological requirements. This validity challenge underscores why property targeting in MOF generation requires sophisticated reward-guided optimization rather than simple fine-tuning approaches. This fundamental limitation motivated our reinforcement learning approach, which we demonstrate successfully addresses both validity and property targeting simultaneously.

\subsection{RL Framework: Simultaneous Validity and Property Control}

\begin{figure}[htbp!]
\centering
\includegraphics[width=\textwidth]{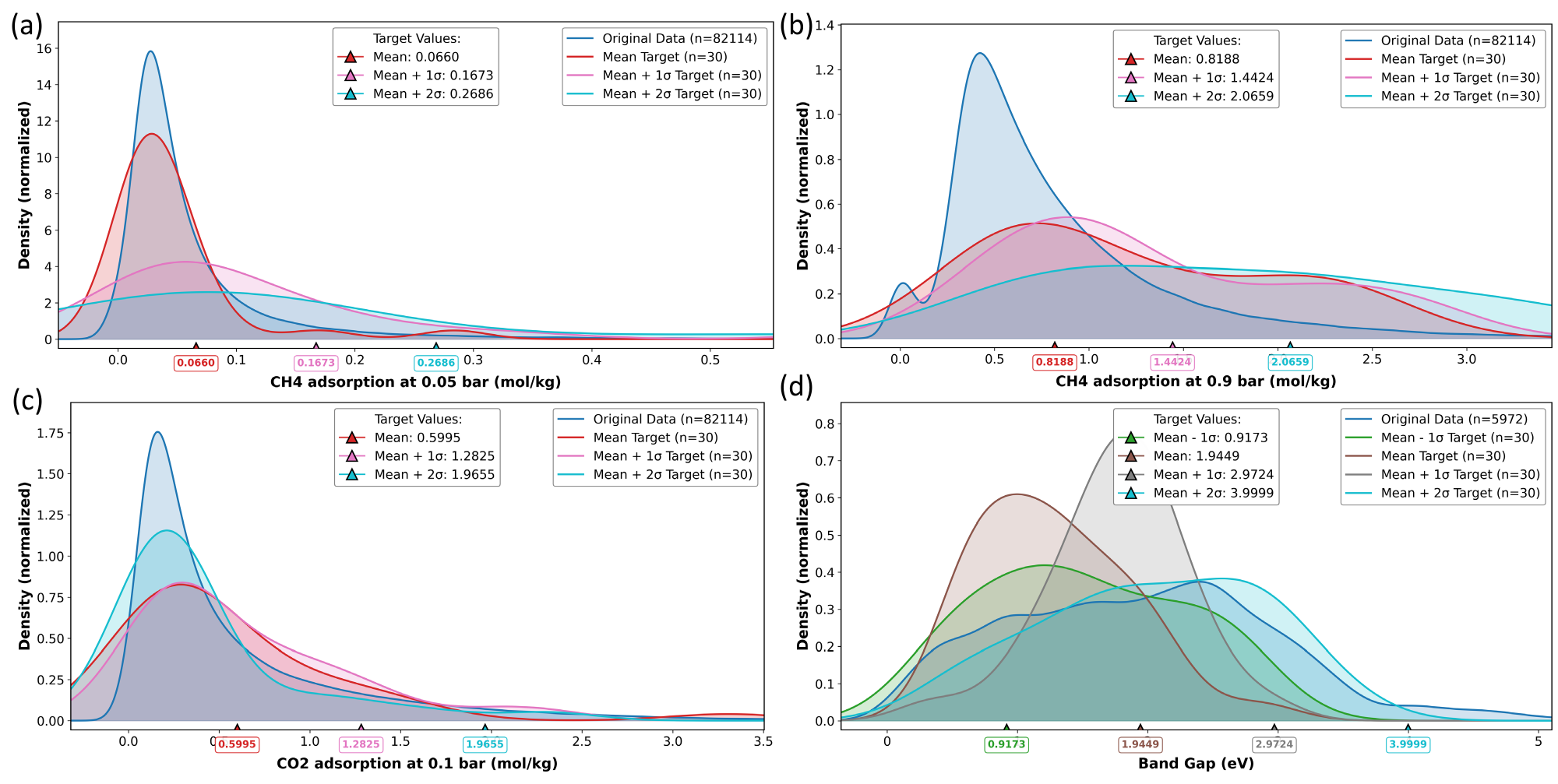}
\caption{Property-targeted generation results across four tasks. (a) CH$_4$ adsorption at 0.05 bar. (b) CH$_4$ adsorption at 0.9 bar. (c) CO$_2$ adsorption at 0.01 bar. (d) Electronic band gap optimization. RL-optimized distributions (colored curves) show focused targeting around desired property values, contrasting with the broad original dataset distributions (blue).}
\label{fig:property_distribution}
\end{figure}

Our results demonstrate three key achievements: resolution of the validity issues affecting generative MOF design, systematic property targeting across diverse domains, and achieving structural diversity during optimization. Here, we show results demonstrating that our reinforcement learning approach achieving chemical validity and property targeting across three distinct tasks: CH$_4$ adsorption at 0.05 bar, CH$_4$ adsorption at 0.9 bar, CO$_2$ adsorption at 0.01 bar, and electronic band gap values. For each task, we generated structures until obtaining 30 that satisfied all criteria (valid, novel, and diverse). The percentages reported in Table \ref{tab:multi_property_performance} represent the fraction of total generations meeting each criterion. We choose strategic targets for each particular dataset based on the corresponding original data distributions: we analyze the mean and standard deviation ($\sigma$) of the original distributions and choose three targets each corresponding to mean, mean+$\sigma$ and mean+2$\sigma$ of the original data distributions. Additionally we choose mean-$\sigma$ for band gap optimization, since a lower numeric value is a better target for that task. Figure \ref{fig:property_distribution} demonstrates our framework's ability to systematically reshape property distributions, inducing a distribution shift toward desired regions while maintaining chemical validity - a capability completely absent in fine-tuned approaches.

Table \ref{tab:multi_property_performance} presents validation of our RL approach: achieving substantial validity rates while simultaneously optimizing for specific property targets. The stark contrast between invalid generations from fine-tuned models and the validity percentages for RL approaches across all scenarios demonstrates an advantage of our framework. We observe that even the most challenging extreme targets (mean+2$\sigma$) at the right tail of the original data distribution, which have very few naturally existing points, maintain substantial validity rates. This proves that chemical feasibility and ambitious property targeting are not mutually exclusive when proper reward-guided optimization is employed.

\begin{table}[htbp!]
\centering
\caption{Performance Metrics for RL-Based MOF Generation Across Different Property Targets.}
\label{tab:multi_property_performance}
\begin{tabular}{p{5cm}lccc}
\toprule
\hline
\textbf{Property} & \textbf{Target} & \textbf{Validity (\%)} & \textbf{Novelty (\%)} & \textbf{Diversity (\%)} \\
\midrule
\hline
\multirow{3}{5cm}{CH$_4$ adsorption at 0.05 bar (mol/kg)}
& Mean & 49.2 & 90.9 & 97 \\
& Mean + 1$\sigma$ & 39.5 & 83.33 & 97 \\
& Mean + 2$\sigma$ & 58.3 & 85.71 & 97 \\
\midrule
\hline
\multirow{3}{5cm}{CH$_4$ adsorption at 0.9 bar (mol/kg)}
& Mean & 71.21 & 63.82 & 100 \\
& Mean + 1$\sigma$ & 52 & 76.92 & 99 \\
& Mean + 2$\sigma$ & 53.42 & 76.92 & 97 \\
\midrule
\hline
\multirow{3}{5cm}{CO$_2$ adsorption at 0.01 bar (mol/kg)}
& Mean & 42.85 & 83.33 & 99 \\
& Mean + 1$\sigma$ & 50 & 83.33 & 99 \\
& Mean + 2$\sigma$ & 100 & 100 & 100 \\
\midrule
\hline
\multirow{3}{5cm}{Band gap (eV)}
& Mean & 100 & 93.75 & 93.75 \\
& Mean + 1$\sigma$ & 64 & 93.75 & 100 \\
& Mean + 2$\sigma$ & 35.63 & 100 & 83.0 \\
& Mean - 1$\sigma$ & 39.75 & 90.90 & 100 \\
\hline
\bottomrule
\end{tabular}
\end{table}

Further, the validity metrics reveal trade-offs that reflect the realities of chemical space exploration. As optimization targets become more ambitious, validity rates show variation while novelty generally increases, indicating the framework is successfully exploring less-represented regions of chemical space where high-performance materials reside. For instance, for the CO$_2$ adsorption dataset, we attempt extreme targeting (mean+2$\sigma$), which achieves 100\% validity, representing a remarkable accomplishment given that these targets correspond to the top 4.5\% of property performance in the original dataset. Similarly, band gap optimization maintains validity rates above 39\% even for extreme targets (in this case the lower band gap values!), demonstrating the framework's ability to navigate challenging regions of electronic property space while preserving chemical feasibility. Novelty rates are consistently more than 63\% and diversity metrics consistently exceed 83\% across all scenarios, demonstrating effective prevention of mode collapse even when optimizing for specific property targets.

\begin{table}[htbp!]
\centering
\caption{Statistical Properties of Generated Structures Across Different Property Domains}
\label{tab:statistical_properties}
\begin{tabular}{llcc}
\toprule
\hline
\textbf{Property} & \textbf{Dataset (Target)} & \textbf{Mean} & \textbf{Std Dev} \\
\midrule
\hline
\multirow{3}{5cm}{CH$_4$ adsorption at 0.05 bar (mol/kg)} 
& Original Data & 0.066 & 0.101 \\
& RL (Mean) & 0.044 & 0.054 \\
& RL (Mean + 1$\sigma$) & 0.109 & 0.127 \\
& RL (Mean + 2$\sigma$) & 0.162 & 0.243 \\
\midrule
\hline
\multirow{3}{5cm}{CH$_4$ adsorption at 0.9 bar (mol/kg)} 
& Original Data & 0.818 & 0.623 \\
& RL (Mean) & 1.224 & 0.745 \\
& RL (Mean + 1$\sigma$) & 1.342 & 0.760 \\
& RL (Mean + 2$\sigma$) & 1.834 & 1.004 \\
\midrule
\hline
\multirow{3}{5cm}{CO$_2$ adsorption at 0.1 bar (mol/kg)} 
& Original Data & 0.599 & 0.682 \\
& RL (Mean) & 0.576 & 0.657 \\
& RL (Mean + 1$\sigma$) & 0.637 & 0.563 \\
& RL (Mean + 2$\sigma$) & 0.432 & 0.522 \\
\midrule
\hline
\multirow{3}{5cm}{Band gap (eV)}
& Original Data & 1.944 & 1.027 \\
& RL (Mean - 1$\sigma$) & 1.523 & 0.769 \\
& RL (Mean) & 1.289 & 0.577 \\
& RL (Mean + 1$\sigma$) & 1.767 & 0.520 \\
& RL (Mean + 2$\sigma$) & 2.044 & 0.83 \\
\hline
\bottomrule
\end{tabular}\\[0.2cm]
\footnotesize{*Fine-tuned model achieved 0\% validity; so the associated statistics are not shown in this table}
\end{table}

Table \ref{tab:statistical_properties} provides comprehensive statistical analysis of all chemically valid RL-generated structures. Beyond achieving specific means, our approach demonstrates potential control over entire distribution characteristics. For example, for CH$_4$ adsorption at 0.05 bar, extreme targeting (mean+2$\sigma$) achieved a broader distribution ($\sigma$ = 0.243 mol/kg) compared to the original dataset variance ($\sigma$ = 0.101 mol/kg), indicating successful exploration of high-performance regions. Band gap optimization shows distributional control across different targets, with standard deviations indicating the model centers reasonably around the desired targets. This tasks demonstrates versatility, spanning from better conductivity (lower band gaps) to improved insulating characteristics (higher band gaps).

To demonstrate the broad applicability of our framework, we conducted additional validation experiments across the complete range of pressure conditions in our dataset. While our primary results focus on select conditions, we also validated our approach across all available pressure conditions (detailed results provided in Supporting Information). This cross-pressure consistency demonstrates that our RL framework captures fundamental structure-property relationships rather than overfitting to specific experimental conditions.

\subsubsection{Sample Analysis of Generated MOF Structures}

To further our understanding of the type of outputs generated by our RL framework, we conduct a sample analysis of generated structures across various tasks. This analysis reveals consistent trends in how our RL framework explores chemical space across different operational regimes. Across property-specific targets, the model learns to assemble metal nodes and organic linkers into topologies consistent with known high-performing motifs, while also generating novel compositions. The generated MOFs across tasks, a sample of which are shown in Table \ref{tab:generated_examples} feature diverse topologies (e.g., \texttt{pcu}, \texttt{nbo}) and exhibit both common and unconventional linker-metal combinations. For instance, while Zn-based and Cu-based clusters are frequently selected, their pairing with alkynes, halogens, or amine-functionalized linkers shows some novelty beyond the training distribution.

\begin{table}[htbp!]
\centering
\caption{Representative Examples of RL-Generated MOF Structures with Target Properties}
\label{tab:generated_examples}
\begin{tabularx}{\textwidth}{p{4cm}p{9cm}p{2cm}}
\toprule
\hline
\textbf{Target} & \textbf{Generated MOFid} & \textbf{Predicted Property} \\
\midrule
\hline
High CH$_4$ adsorption at 0.5 bar (mol/kg) 
& \seqsplit{$CCCOc1cc(cc(c1C(=O)[O-])OCCC)c1ccc(cc1)c1cc(OCCC)c(c(c1)OCCC)C(=O)[O-].[O-]C(=O)c1ccc(cc1)C(=O)[O-].[Zn][Zn]\&\&nbo.cat0$}
& 2.414 mol/kg \\
\midrule
\hline
High CO$_2$ adsorption at 0.5 bar (mol/kg)
& \seqsplit{$CCCC(=CC(=O)[O-])C=CC(=O)[O-].CCCC1CC2(CCC1(CC2)C(=O)[O-])C(=O)[O-].CCC[N][CH][CH][NH].[Cu][Cu]\&\&pcu.cat0$} 
& 3.33 mol/kg \\
\midrule
\hline
Low Band Gap (eV)
& \seqsplit{$N\#CC\#N.[O-]C(=O)C\#CC\#CC(=O)[O-].[O-]C(=O)C12C3C4(C2(C2C1(C3(C42C(=O)[O-])Cl)Cl)Cl)Cl.[Zn][Zn]\&\&pcu.cat1$} 
& 0.435 eV \\
\bottomrule
\hline
\end{tabularx}
\end{table}

Our analysis reveals that the RL framework successfully identifies structure-property correlations that drive enhanced performance. For high CH$_4$ uptake targets, generated MOFs show preference for rigid aromatic linkers bearing carboxylates, combined with high-porosity, low-catenation topologies such as \texttt{nbo}, and Zn$^{2+}$ nodes known for forming stable, spacious cages\cite{wilmer2012large, ahmed2019exceptional}. High CO$_2$ uptake structures consistently feature open Cu$^{2+}$ paddlewheel units and nitrogen- or oxygen-rich linkers, enabling stronger electrostatic and Lewis acid-base interactions\cite{chung2014computation, mason2011evaluating}. In contrast, low band gap optimization yields MOFs with strong $\pi$-conjugation, donor-acceptor functionality (e.g., $-C\#N$, $-C\#C$ triple bonds, Cl halogens), and $\pi$-stacking potential enabled by interpenetrated \texttt{pcu.cat1} networks, together creating low-energy frontier orbitals and reduced band gaps\cite{sun2017iron, pham2014engineering, narayan2012high}.

\section{Conclusions}

In this work, we introduced a reinforcement learning-enhanced GPT-based generative framework for target-specific de novo MOF design. Our approach integrates a GPT-based MOF generator with property-aware feedback through a transformer-based property predictor, using a comprehensive reward function that balances validity, novelty, diversity, and property targeting. The framework addresses the challenge of generating chemically valid MOFs while controlling their functional properties.

The experimental results reveal an important limitation in current generative design approaches like supervised fine-tuning that our RL framework helps to address. While fine-tuned models consistently generate invalid structures across tasks, our RL-optimized approach guides MOF generation toward specific property targets while maintaining reasonable validity rates, with structural diversity exceeding 83\%. This difference between fine-tuned and RL approaches represents a meaningful improvement in computational materials design methodology. The observed shifts in property distributions toward challenging targets demonstrate the effectiveness of our approach in navigating the complex chemical space of MOFs.

The targeting performance analysis shows controlled optimization over MOF property landscapes across three distinct domains. Our framework demonstrates good targeting performance for moderate objectives, achieving improvements in CH$_4$ adsorption targets and systematic band gap reduction for enhanced conductivity applications. Even for extreme targets corresponding to the top 4.5\% of property performance in the original dataset, the framework maintains appreciable validity rates while generating more than 70\% novel structures. The ability to target properties in the tail of the distribution (mean+2$\sigma$) while maintaining reasonable validity suggests that our approach can explore underrepresented regions of chemical space, potentially leading to the discovery of MOFs with enhanced properties that would be difficult to identify through traditional screening approaches.

Cross-domain validation across gas adsorption and electronic properties establishes the framework's applicability without requiring domain-specific modifications. The consistent performance across different property types demonstrates that the underlying RL methodology can optimize diverse functional characteristics using the same computational infrastructure with minimal changes. This versatility is valuable for materials applications requiring multi-property optimization.

This work contributes to AI-driven materials discovery by combining generative transformers, reinforcement learning, and domain-specific molecular representations to address the challenge of simultaneous validity and property control. The proposed framework enables targeted design of MOFs with optimized functional properties, contributing to a shift from screening existing materials to designing new ones with predetermined characteristics. The reinforcement learning framework developed here provides a foundation that could be extended to other materials design tasks. Future work could focus on integrating feedback from physics-based simulations and multimodal approaches, to further enhance the design of high-performance MOFs.

\section{Data and Code Availability}

The necessary information containing the code and data used in this
study is available here: https://github.com/srivathsanb14/MOFGPT

\section{Supporting Information}

The Supporting Information includes detailed MOFid tokenization procedures, comprehensive dataset information, complete model architecture and three-stage training methodology, multi-component reward function design, generation and evaluation protocols, and additional property optimization results across all 11 conditions including CH$_4$ and CO$_2$ adsorption at multiple pressures and electronic band gap targeting.

\section{Author Contributions}

S.B., R.M., A.A. and A.B.F. conceived the project. S.B. developed the core architecture and algorithms. R.M. and A.A. contributed to the implementation of the reinforcement learning framework. R.S.M. assisted with data preprocessing. All authors contributed to the analysis of results and manuscript preparation. A.B.F. supervised the project.

\bibliography{main}

\providecommand{\latin}[1]{#1}
\makeatletter
\providecommand{\doi}
  {\begingroup\let\do\@makeother\dospecials
  \catcode`\{=1 \catcode`\}=2 \doi@aux}
\providecommand{\doi@aux}[1]{\endgroup\texttt{#1}}
\makeatother
\providecommand*\mcitethebibliography{\thebibliography}
\csname @ifundefined\endcsname{endmcitethebibliography}  {\let\endmcitethebibliography\endthebibliography}{}
\begin{mcitethebibliography}{76}
\providecommand*\natexlab[1]{#1}
\providecommand*\mciteSetBstSublistMode[1]{}
\providecommand*\mciteSetBstMaxWidthForm[2]{}
\providecommand*\mciteBstWouldAddEndPuncttrue
  {\def\EndOfBibitem{\unskip.}}
\providecommand*\mciteBstWouldAddEndPunctfalse
  {\let\EndOfBibitem\relax}
\providecommand*\mciteSetBstMidEndSepPunct[3]{}
\providecommand*\mciteSetBstSublistLabelBeginEnd[3]{}
\providecommand*\EndOfBibitem{}
\mciteSetBstSublistMode{f}
\mciteSetBstMaxWidthForm{subitem}{(\alph{mcitesubitemcount})}
\mciteSetBstSublistLabelBeginEnd
  {\mcitemaxwidthsubitemform\space}
  {\relax}
  {\relax}

\bibitem[Getman \latin{et~al.}(2012)Getman, Bae, Wilmer, and Snurr]{getman2012review}
Getman,~R.~B.; Bae,~Y.-S.; Wilmer,~C.~E.; Snurr,~R.~Q. Review and analysis of molecular simulations of methane, hydrogen, and acetylene storage in metal--organic frameworks. \emph{Chemical reviews} \textbf{2012}, \emph{112}, 703--723\relax
\mciteBstWouldAddEndPuncttrue
\mciteSetBstMidEndSepPunct{\mcitedefaultmidpunct}
{\mcitedefaultendpunct}{\mcitedefaultseppunct}\relax
\EndOfBibitem
\bibitem[Boyd \latin{et~al.}(2019)Boyd, Chidambaram, Garc{\'\i}a-D{\'\i}ez, Ireland, Daff, Bounds, G{\l}adysiak, Schouwink, Moosavi, Maroto-Valer, \latin{et~al.} others]{boyd2019data}
Boyd,~P.~G.; Chidambaram,~A.; Garc{\'\i}a-D{\'\i}ez,~E.; Ireland,~C.~P.; Daff,~T.~D.; Bounds,~R.; G{\l}adysiak,~A.; Schouwink,~P.; Moosavi,~S.~M.; Maroto-Valer,~M.~M.; others Data-driven design of metal--organic frameworks for wet flue gas CO2 capture. \emph{Nature} \textbf{2019}, \emph{576}, 253--256\relax
\mciteBstWouldAddEndPuncttrue
\mciteSetBstMidEndSepPunct{\mcitedefaultmidpunct}
{\mcitedefaultendpunct}{\mcitedefaultseppunct}\relax
\EndOfBibitem
\bibitem[Cao \latin{et~al.}(2019)Cao, Liu, and Barati~Farimani]{cao2019water}
Cao,~Z.; Liu,~V.; Barati~Farimani,~A. Water desalination with two-dimensional metal--organic framework membranes. \emph{Nano letters} \textbf{2019}, \emph{19}, 8638--8643\relax
\mciteBstWouldAddEndPuncttrue
\mciteSetBstMidEndSepPunct{\mcitedefaultmidpunct}
{\mcitedefaultendpunct}{\mcitedefaultseppunct}\relax
\EndOfBibitem
\bibitem[Dutta \latin{et~al.}(2024)Dutta, de~Luis, Goscianska, Demessence, Ettlinger, and Wuttke]{dutta2024metal}
Dutta,~S.; de~Luis,~R.~F.; Goscianska,~J.; Demessence,~A.; Ettlinger,~R.; Wuttke,~S. Metal--organic frameworks for water desalination. \emph{Advanced Functional Materials} \textbf{2024}, \emph{34}, 2304790\relax
\mciteBstWouldAddEndPuncttrue
\mciteSetBstMidEndSepPunct{\mcitedefaultmidpunct}
{\mcitedefaultendpunct}{\mcitedefaultseppunct}\relax
\EndOfBibitem
\bibitem[Sun \latin{et~al.}(2018)Sun, Peng, Reeder, Moosavi, Tiana, Britt, Oveisi, and Queen]{sun2018rapid}
Sun,~D.~T.; Peng,~L.; Reeder,~W.~S.; Moosavi,~S.~M.; Tiana,~D.; Britt,~D.~K.; Oveisi,~E.; Queen,~W.~L. Rapid, selective heavy metal removal from water by a metal--organic framework/polydopamine composite. \emph{ACS central science} \textbf{2018}, \emph{4}, 349--356\relax
\mciteBstWouldAddEndPuncttrue
\mciteSetBstMidEndSepPunct{\mcitedefaultmidpunct}
{\mcitedefaultendpunct}{\mcitedefaultseppunct}\relax
\EndOfBibitem
\bibitem[Li \latin{et~al.}(2016)Li, Wen, Cui, Zhou, Qian, and Chen]{https://doi.org/10.1002/adma.201601133}
Li,~B.; Wen,~H.-M.; Cui,~Y.; Zhou,~W.; Qian,~G.; Chen,~B. Emerging Multifunctional Metal–Organic Framework Materials. \emph{Advanced Materials} \textbf{2016}, \emph{28}, 8819--8860\relax
\mciteBstWouldAddEndPuncttrue
\mciteSetBstMidEndSepPunct{\mcitedefaultmidpunct}
{\mcitedefaultendpunct}{\mcitedefaultseppunct}\relax
\EndOfBibitem
\bibitem[He \latin{et~al.}(2014)He, Zhou, Qian, and Chen]{C4CS00032C}
He,~Y.; Zhou,~W.; Qian,~G.; Chen,~B. Methane storage in metal–organic frameworks. \emph{Chem. Soc. Rev.} \textbf{2014}, \emph{43}, 5657--5678\relax
\mciteBstWouldAddEndPuncttrue
\mciteSetBstMidEndSepPunct{\mcitedefaultmidpunct}
{\mcitedefaultendpunct}{\mcitedefaultseppunct}\relax
\EndOfBibitem
\bibitem[Jiao \latin{et~al.}(2019)Jiao, Seow, Skinner, Wang, and Jiang]{JIAO201943}
Jiao,~L.; Seow,~J. Y.~R.; Skinner,~W.~S.; Wang,~Z.~U.; Jiang,~H.-L. Metal–organic frameworks: Structures and functional applications. \emph{Materials Today} \textbf{2019}, \emph{27}, 43--68\relax
\mciteBstWouldAddEndPuncttrue
\mciteSetBstMidEndSepPunct{\mcitedefaultmidpunct}
{\mcitedefaultendpunct}{\mcitedefaultseppunct}\relax
\EndOfBibitem
\bibitem[Moghadam \latin{et~al.}(2017)Moghadam, Li, Wiggin, Tao, Maloney, Wood, Ward, and Fairen-Jimenez]{doi:10.1021/acs.chemmater.7b00441}
Moghadam,~P.~Z.; Li,~A.; Wiggin,~S.~B.; Tao,~A.; Maloney,~A. G.~P.; Wood,~P.~A.; Ward,~S.~C.; Fairen-Jimenez,~D. Development of a Cambridge Structural Database Subset: A Collection of Metal–Organic Frameworks for Past, Present, and Future. \emph{Chemistry of Materials} \textbf{2017}, \emph{29}, 2618--2625\relax
\mciteBstWouldAddEndPuncttrue
\mciteSetBstMidEndSepPunct{\mcitedefaultmidpunct}
{\mcitedefaultendpunct}{\mcitedefaultseppunct}\relax
\EndOfBibitem
\bibitem[Wilmer \latin{et~al.}(2012)Wilmer, Leaf, Lee, Farha, Hauser, Hupp, and Snurr]{wilmer2012large}
Wilmer,~C.~E.; Leaf,~M.; Lee,~C.~Y.; Farha,~O.~K.; Hauser,~B.~G.; Hupp,~J.~T.; Snurr,~R.~Q. Large-scale screening of hypothetical metal--organic frameworks. \emph{Nature chemistry} \textbf{2012}, \emph{4}, 83--89\relax
\mciteBstWouldAddEndPuncttrue
\mciteSetBstMidEndSepPunct{\mcitedefaultmidpunct}
{\mcitedefaultendpunct}{\mcitedefaultseppunct}\relax
\EndOfBibitem
\bibitem[Jablonka \latin{et~al.}(2020)Jablonka, Ongari, Moosavi, and Smit]{doi:10.1021/acs.chemrev.0c00004}
Jablonka,~K.~M.; Ongari,~D.; Moosavi,~S.~M.; Smit,~B. Big-Data Science in Porous Materials: Materials Genomics and Machine Learning. \emph{Chemical Reviews} \textbf{2020}, \emph{120}, 8066--8129, PMID: 32520531\relax
\mciteBstWouldAddEndPuncttrue
\mciteSetBstMidEndSepPunct{\mcitedefaultmidpunct}
{\mcitedefaultendpunct}{\mcitedefaultseppunct}\relax
\EndOfBibitem
\bibitem[Witman \latin{et~al.}(2016)Witman, Ling, Anderson, Tong, Stylianou, Slater, Smit, and Haranczyk]{C6SC01477A}
Witman,~M.; Ling,~S.; Anderson,~S.; Tong,~L.; Stylianou,~K.~C.; Slater,~B.; Smit,~B.; Haranczyk,~M. In silico design and screening of hypothetical MOF-74 analogs and their experimental synthesis. \emph{Chem. Sci.} \textbf{2016}, \emph{7}, 6263--6272\relax
\mciteBstWouldAddEndPuncttrue
\mciteSetBstMidEndSepPunct{\mcitedefaultmidpunct}
{\mcitedefaultendpunct}{\mcitedefaultseppunct}\relax
\EndOfBibitem
\bibitem[Rosen \latin{et~al.}(2019)Rosen, Notestein, and Snurr]{https://doi.org/10.1002/jcc.25787}
Rosen,~A.~S.; Notestein,~J.~M.; Snurr,~R.~Q. Identifying promising metal–organic frameworks for heterogeneous catalysis via high-throughput periodic density functional theory. \emph{Journal of Computational Chemistry} \textbf{2019}, \emph{40}, 1305--1318\relax
\mciteBstWouldAddEndPuncttrue
\mciteSetBstMidEndSepPunct{\mcitedefaultmidpunct}
{\mcitedefaultendpunct}{\mcitedefaultseppunct}\relax
\EndOfBibitem
\bibitem[Nandy \latin{et~al.}(2023)Nandy, Yue, Oh, Duan, Terrones, Chung, and Kulik]{NANDY20231585}
Nandy,~A.; Yue,~S.; Oh,~C.; Duan,~C.; Terrones,~G.~G.; Chung,~Y.~G.; Kulik,~H.~J. A database of ultrastable MOFs reassembled from stable fragments with machine learning models. \emph{Matter} \textbf{2023}, \emph{6}, 1585--1603\relax
\mciteBstWouldAddEndPuncttrue
\mciteSetBstMidEndSepPunct{\mcitedefaultmidpunct}
{\mcitedefaultendpunct}{\mcitedefaultseppunct}\relax
\EndOfBibitem
\bibitem[Karamad \latin{et~al.}(2020)Karamad, Magar, Shi, Siahrostami, Gates, and Barati~Farimani]{karamad2020orbital}
Karamad,~M.; Magar,~R.; Shi,~Y.; Siahrostami,~S.; Gates,~I.~D.; Barati~Farimani,~A. Orbital graph convolutional neural network for material property prediction. \emph{Physical Review Materials} \textbf{2020}, \emph{4}, 093801\relax
\mciteBstWouldAddEndPuncttrue
\mciteSetBstMidEndSepPunct{\mcitedefaultmidpunct}
{\mcitedefaultendpunct}{\mcitedefaultseppunct}\relax
\EndOfBibitem
\bibitem[Moosavi \latin{et~al.}(2020)Moosavi, Jablonka, and Smit]{doi:10.1021/jacs.0c09105}
Moosavi,~S.~M.; Jablonka,~K.~M.; Smit,~B. The Role of Machine Learning in the Understanding and Design of Materials. \emph{Journal of the American Chemical Society} \textbf{2020}, \emph{142}, 20273--20287, PMID: 33170678\relax
\mciteBstWouldAddEndPuncttrue
\mciteSetBstMidEndSepPunct{\mcitedefaultmidpunct}
{\mcitedefaultendpunct}{\mcitedefaultseppunct}\relax
\EndOfBibitem
\bibitem[Islamov \latin{et~al.}(2023)Islamov, Babaei, Anderson, Sezginel, Long, McGaughey, Gomez-Gualdron, and Wilmer]{islamov2023high}
Islamov,~M.; Babaei,~H.; Anderson,~R.; Sezginel,~K.~B.; Long,~J.~R.; McGaughey,~A.~J.; Gomez-Gualdron,~D.~A.; Wilmer,~C.~E. High-throughput screening of hypothetical metal-organic frameworks for thermal conductivity. \emph{npj Computational Materials} \textbf{2023}, \emph{9}, 11\relax
\mciteBstWouldAddEndPuncttrue
\mciteSetBstMidEndSepPunct{\mcitedefaultmidpunct}
{\mcitedefaultendpunct}{\mcitedefaultseppunct}\relax
\EndOfBibitem
\bibitem[Rosen \latin{et~al.}(2022)Rosen, Notestein, and Snurr]{ROSEN2022100760}
Rosen,~A.~S.; Notestein,~J.~M.; Snurr,~R.~Q. Realizing the data-driven, computational discovery of metal-organic framework catalysts. \emph{Current Opinion in Chemical Engineering} \textbf{2022}, \emph{35}, 100760\relax
\mciteBstWouldAddEndPuncttrue
\mciteSetBstMidEndSepPunct{\mcitedefaultmidpunct}
{\mcitedefaultendpunct}{\mcitedefaultseppunct}\relax
\EndOfBibitem
\bibitem[Park \latin{et~al.}(2024)Park, Kim, Kang, Lim, and Kim]{doi:10.1021/jacsau.4c00618}
Park,~J.; Kim,~H.; Kang,~Y.; Lim,~Y.; Kim,~J. From Data to Discovery: Recent Trends of Machine Learning in Metal–Organic Frameworks. \emph{JACS Au} \textbf{2024}, \emph{4}, 3727--3743\relax
\mciteBstWouldAddEndPuncttrue
\mciteSetBstMidEndSepPunct{\mcitedefaultmidpunct}
{\mcitedefaultendpunct}{\mcitedefaultseppunct}\relax
\EndOfBibitem
\bibitem[Cao \latin{et~al.}(2023)Cao, Magar, Wang, and Barati~Farimani]{doi:10.1021/jacs.2c11420}
Cao,~Z.; Magar,~R.; Wang,~Y.; Barati~Farimani,~A. MOFormer: Self-Supervised Transformer Model for Metal–Organic Framework Property Prediction. \emph{Journal of the American Chemical Society} \textbf{2023}, \emph{145}, 2958--2967, PMID: 36706365\relax
\mciteBstWouldAddEndPuncttrue
\mciteSetBstMidEndSepPunct{\mcitedefaultmidpunct}
{\mcitedefaultendpunct}{\mcitedefaultseppunct}\relax
\EndOfBibitem
\bibitem[Kang \latin{et~al.}(2025)Kang, Lee, Bae, Han, Jang, and Kim]{kang2025harnessing}
Kang,~Y.; Lee,~W.; Bae,~T.; Han,~S.; Jang,~H.; Kim,~J. Harnessing Large Language Models to Collect and Analyze Metal--Organic Framework Property Data Set. \emph{Journal of the American Chemical Society} \textbf{2025}, \relax
\mciteBstWouldAddEndPunctfalse
\mciteSetBstMidEndSepPunct{\mcitedefaultmidpunct}
{}{\mcitedefaultseppunct}\relax
\EndOfBibitem
\bibitem[Bai \latin{et~al.}(2024)Bai, Xie, Zhang, Han, and Li]{bai2024evaluation}
Bai,~X.; Xie,~Y.; Zhang,~X.; Han,~H.; Li,~J.-R. Evaluation of open-source large language models for metal--organic frameworks research. \emph{Journal of Chemical Information and Modeling} \textbf{2024}, \emph{64}, 4958--4965\relax
\mciteBstWouldAddEndPuncttrue
\mciteSetBstMidEndSepPunct{\mcitedefaultmidpunct}
{\mcitedefaultendpunct}{\mcitedefaultseppunct}\relax
\EndOfBibitem
\bibitem[Radford \latin{et~al.}(2018)Radford, Narasimhan, Salimans, Sutskever, \latin{et~al.} others]{radford2018improving}
Radford,~A.; Narasimhan,~K.; Salimans,~T.; Sutskever,~I.; others Improving language understanding by generative pre-training. \textbf{2018}, \relax
\mciteBstWouldAddEndPunctfalse
\mciteSetBstMidEndSepPunct{\mcitedefaultmidpunct}
{}{\mcitedefaultseppunct}\relax
\EndOfBibitem
\bibitem[Brown \latin{et~al.}(2020)Brown, Mann, Ryder, Subbiah, Kaplan, Dhariwal, Neelakantan, Shyam, Sastry, Askell, Agarwal, Herbert-Voss, Krueger, Henighan, Child, Ramesh, Ziegler, Wu, Winter, Hesse, Chen, Sigler, Litwin, Gray, Chess, Clark, Berner, McCandlish, Radford, Sutskever, and Amodei]{brown2020language}
Brown,~T.~B. \latin{et~al.}  Language Models are Few-Shot Learners. \textbf{2020}, \relax
\mciteBstWouldAddEndPunctfalse
\mciteSetBstMidEndSepPunct{\mcitedefaultmidpunct}
{}{\mcitedefaultseppunct}\relax
\EndOfBibitem
\bibitem[Magar \latin{et~al.}(2022)Magar, Wang, and Barati~Farimani]{magar2022crystal}
Magar,~R.; Wang,~Y.; Barati~Farimani,~A. Crystal twins: self-supervised learning for crystalline material property prediction. \emph{npj Computational Materials} \textbf{2022}, \emph{8}, 231\relax
\mciteBstWouldAddEndPuncttrue
\mciteSetBstMidEndSepPunct{\mcitedefaultmidpunct}
{\mcitedefaultendpunct}{\mcitedefaultseppunct}\relax
\EndOfBibitem
\bibitem[Demir \latin{et~al.}(2023)Demir, Daglar, Gulbalkan, Aksu, and Keskin]{DEMIR2023215112}
Demir,~H.; Daglar,~H.; Gulbalkan,~H.~C.; Aksu,~G.~O.; Keskin,~S. Recent advances in computational modeling of MOFs: From molecular simulations to machine learning. \emph{Coordination Chemistry Reviews} \textbf{2023}, \emph{484}, 215112\relax
\mciteBstWouldAddEndPuncttrue
\mciteSetBstMidEndSepPunct{\mcitedefaultmidpunct}
{\mcitedefaultendpunct}{\mcitedefaultseppunct}\relax
\EndOfBibitem
\bibitem[Altintas \latin{et~al.}(2021)Altintas, Altundal, Keskin, and Yildirim]{doi:10.1021/acs.jcim.1c00191}
Altintas,~C.; Altundal,~O.~F.; Keskin,~S.; Yildirim,~R. Machine Learning Meets with Metal Organic Frameworks for Gas Storage and Separation. \emph{Journal of Chemical Information and Modeling} \textbf{2021}, \emph{61}, 2131--2146, PMID: 33914526\relax
\mciteBstWouldAddEndPuncttrue
\mciteSetBstMidEndSepPunct{\mcitedefaultmidpunct}
{\mcitedefaultendpunct}{\mcitedefaultseppunct}\relax
\EndOfBibitem
\bibitem[Nandy \latin{et~al.}(2022)Nandy, Terrones, Arunachalam, Duan, Kastner, and Kulik]{nandy2022mofsimplify}
Nandy,~A.; Terrones,~G.; Arunachalam,~N.; Duan,~C.; Kastner,~D.~W.; Kulik,~H.~J. MOFSimplify, machine learning models with extracted stability data of three thousand metal--organic frameworks. \emph{Scientific Data} \textbf{2022}, \emph{9}, 74\relax
\mciteBstWouldAddEndPuncttrue
\mciteSetBstMidEndSepPunct{\mcitedefaultmidpunct}
{\mcitedefaultendpunct}{\mcitedefaultseppunct}\relax
\EndOfBibitem
\bibitem[Lim \latin{et~al.}(2024)Lim, Park, Walsh, and Kim]{lim2024accelerating}
Lim,~Y.; Park,~H.; Walsh,~A.; Kim,~J. Accelerating CO₂ Direct Air Capture Screening for Metal-Organic Frameworks with a Transferable Machine Learning Force Field. \textbf{2024}, \relax
\mciteBstWouldAddEndPunctfalse
\mciteSetBstMidEndSepPunct{\mcitedefaultmidpunct}
{}{\mcitedefaultseppunct}\relax
\EndOfBibitem
\bibitem[Xie \latin{et~al.}(2021)Xie, Fu, Ganea, Barzilay, and Jaakkola]{xie2021crystal}
Xie,~T.; Fu,~X.; Ganea,~O.-E.; Barzilay,~R.; Jaakkola,~T. Crystal Diffusion Variational Autoencoder for Periodic Material Generation. \emph{arXiv preprint arXiv:2110.06197} \textbf{2021}, \relax
\mciteBstWouldAddEndPunctfalse
\mciteSetBstMidEndSepPunct{\mcitedefaultmidpunct}
{}{\mcitedefaultseppunct}\relax
\EndOfBibitem
\bibitem[Luong and Singh(2024)Luong, and Singh]{doi:10.1021/acs.jcim.3c02070}
Luong,~K.-D.; Singh,~A. Application of Transformers in Cheminformatics. \emph{Journal of Chemical Information and Modeling} \textbf{2024}, \emph{64}, 4392--4409, PMID: 38815246\relax
\mciteBstWouldAddEndPuncttrue
\mciteSetBstMidEndSepPunct{\mcitedefaultmidpunct}
{\mcitedefaultendpunct}{\mcitedefaultseppunct}\relax
\EndOfBibitem
\bibitem[Ock \latin{et~al.}(2024)Ock, Badrinarayanan, Magar, Antony, and Barati~Farimani]{ock2024multimodal}
Ock,~J.; Badrinarayanan,~S.; Magar,~R.; Antony,~A.; Barati~Farimani,~A. Multimodal language and graph learning of adsorption configuration in catalysis. \emph{Nature Machine Intelligence} \textbf{2024}, 1--11\relax
\mciteBstWouldAddEndPuncttrue
\mciteSetBstMidEndSepPunct{\mcitedefaultmidpunct}
{\mcitedefaultendpunct}{\mcitedefaultseppunct}\relax
\EndOfBibitem
\bibitem[Guntuboina \latin{et~al.}(2023)Guntuboina, Das, Mollaei, Kim, and Barati~Farimani]{doi:10.1021/acs.jpclett.3c02398}
Guntuboina,~C.; Das,~A.; Mollaei,~P.; Kim,~S.; Barati~Farimani,~A. PeptideBERT: A Language Model Based on Transformers for Peptide Property Prediction. \emph{The Journal of Physical Chemistry Letters} \textbf{2023}, \emph{14}, 10427--10434, PMID: 37956397\relax
\mciteBstWouldAddEndPuncttrue
\mciteSetBstMidEndSepPunct{\mcitedefaultmidpunct}
{\mcitedefaultendpunct}{\mcitedefaultseppunct}\relax
\EndOfBibitem
\bibitem[Ock \latin{et~al.}(2023)Ock, Guntuboina, and Barati~Farimani]{doi:10.1021/acscatal.3c04956}
Ock,~J.; Guntuboina,~C.; Barati~Farimani,~A. Catalyst Energy Prediction with CatBERTa: Unveiling Feature Exploration Strategies through Large Language Models. \emph{ACS Catalysis} \textbf{2023}, \emph{13}, 16032--16044\relax
\mciteBstWouldAddEndPuncttrue
\mciteSetBstMidEndSepPunct{\mcitedefaultmidpunct}
{\mcitedefaultendpunct}{\mcitedefaultseppunct}\relax
\EndOfBibitem
\bibitem[Hendon \latin{et~al.}(2017)Hendon, Rieth, Korzyński, and Dincă]{doi:10.1021/acscentsci.7b00197}
Hendon,~C.~H.; Rieth,~A.~J.; Korzyński,~M.~D.; Dincă,~M. Grand Challenges and Future Opportunities for Metal–Organic Frameworks. \emph{ACS Central Science} \textbf{2017}, \emph{3}, 554--563, PMID: 28691066\relax
\mciteBstWouldAddEndPuncttrue
\mciteSetBstMidEndSepPunct{\mcitedefaultmidpunct}
{\mcitedefaultendpunct}{\mcitedefaultseppunct}\relax
\EndOfBibitem
\bibitem[Gupta \latin{et~al.}(2018)Gupta, Müller, Huisman, Fuchs, Schneider, and Schneider]{https://doi.org/10.1002/minf.201700111}
Gupta,~A.; Müller,~A.~T.; Huisman,~B. J.~H.; Fuchs,~J.~A.; Schneider,~P.; Schneider,~G. Generative Recurrent Networks for De Novo Drug Design. \emph{Molecular Informatics} \textbf{2018}, \emph{37}, 1700111\relax
\mciteBstWouldAddEndPuncttrue
\mciteSetBstMidEndSepPunct{\mcitedefaultmidpunct}
{\mcitedefaultendpunct}{\mcitedefaultseppunct}\relax
\EndOfBibitem
\bibitem[Bengio \latin{et~al.}(1993)Bengio, Frasconi, and Simard]{298725}
Bengio,~Y.; Frasconi,~P.; Simard,~P. The problem of learning long-term dependencies in recurrent networks. IEEE International Conference on Neural Networks. 1993; pp 1183--1188 vol.3\relax
\mciteBstWouldAddEndPuncttrue
\mciteSetBstMidEndSepPunct{\mcitedefaultmidpunct}
{\mcitedefaultendpunct}{\mcitedefaultseppunct}\relax
\EndOfBibitem
\bibitem[M{\'e}ndez-Lucio \latin{et~al.}(2020)M{\'e}ndez-Lucio, Baillif, Clevert, Rouqui{\'e}, and Wichard]{mendez2020novo}
M{\'e}ndez-Lucio,~O.; Baillif,~B.; Clevert,~D.-A.; Rouqui{\'e},~D.; Wichard,~J. De novo generation of hit-like molecules from gene expression signatures using artificial intelligence. \emph{Nature communications} \textbf{2020}, \emph{11}, 10\relax
\mciteBstWouldAddEndPuncttrue
\mciteSetBstMidEndSepPunct{\mcitedefaultmidpunct}
{\mcitedefaultendpunct}{\mcitedefaultseppunct}\relax
\EndOfBibitem
\bibitem[Kang and Cho(2019)Kang, and Cho]{doi:10.1021/acs.jcim.8b00263}
Kang,~S.; Cho,~K. Conditional Molecular Design with Deep Generative Models. \emph{Journal of Chemical Information and Modeling} \textbf{2019}, \emph{59}, 43--52, PMID: 30016587\relax
\mciteBstWouldAddEndPuncttrue
\mciteSetBstMidEndSepPunct{\mcitedefaultmidpunct}
{\mcitedefaultendpunct}{\mcitedefaultseppunct}\relax
\EndOfBibitem
\bibitem[Reiser \latin{et~al.}(2022)Reiser, Neubert, Eberhard, Torresi, Zhou, Shao, Metni, van Hoesel, Schopmans, Sommer, \latin{et~al.} others]{reiser2022graph}
Reiser,~P.; Neubert,~M.; Eberhard,~A.; Torresi,~L.; Zhou,~C.; Shao,~C.; Metni,~H.; van Hoesel,~C.; Schopmans,~H.; Sommer,~T.; others Graph neural networks for materials science and chemistry. \emph{Communications Materials} \textbf{2022}, \emph{3}, 93\relax
\mciteBstWouldAddEndPuncttrue
\mciteSetBstMidEndSepPunct{\mcitedefaultmidpunct}
{\mcitedefaultendpunct}{\mcitedefaultseppunct}\relax
\EndOfBibitem
\bibitem[Park \latin{et~al.}(2024)Park, Yan, Zhu, Huerta, Chaudhuri, Cooper, Foster, and Tajkhorshid]{park2024generative}
Park,~H.; Yan,~X.; Zhu,~R.; Huerta,~E.~A.; Chaudhuri,~S.; Cooper,~D.; Foster,~I.; Tajkhorshid,~E. A generative artificial intelligence framework based on a molecular diffusion model for the design of metal-organic frameworks for carbon capture. \emph{Communications Chemistry} \textbf{2024}, \emph{7}, 21\relax
\mciteBstWouldAddEndPuncttrue
\mciteSetBstMidEndSepPunct{\mcitedefaultmidpunct}
{\mcitedefaultendpunct}{\mcitedefaultseppunct}\relax
\EndOfBibitem
\bibitem[Inizan \latin{et~al.}(2025)Inizan, Yang, Kaplan, hsu Lin, Yin, Mirzaei, Abdelgaid, Alawadhi, Cho, Zheng, Cubuk, Borgs, Chayes, Persson, and Yaghi]{inizan2025agenticaidiscoverymetalorganic}
Inizan,~T.~J.; Yang,~S.; Kaplan,~A.; hsu Lin,~Y.; Yin,~J.; Mirzaei,~S.; Abdelgaid,~M.; Alawadhi,~A.~H.; Cho,~K.; Zheng,~Z.; Cubuk,~E.~D.; Borgs,~C.; Chayes,~J.~T.; Persson,~K.~A.; Yaghi,~O.~M. System of Agentic AI for the Discovery of Metal-Organic Frameworks. 2025; \url{https://arxiv.org/abs/2504.14110}\relax
\mciteBstWouldAddEndPuncttrue
\mciteSetBstMidEndSepPunct{\mcitedefaultmidpunct}
{\mcitedefaultendpunct}{\mcitedefaultseppunct}\relax
\EndOfBibitem
\bibitem[Fu \latin{et~al.}(2024)Fu, Xie, Rosen, Jaakkola, and Smith]{fu2024mofdiff}
Fu,~X.; Xie,~T.; Rosen,~A.~S.; Jaakkola,~T.~S.; Smith,~J.~A. {MOFD}iff: Coarse-grained Diffusion for Metal-Organic Framework Design. The Twelfth International Conference on Learning Representations. 2024\relax
\mciteBstWouldAddEndPuncttrue
\mciteSetBstMidEndSepPunct{\mcitedefaultmidpunct}
{\mcitedefaultendpunct}{\mcitedefaultseppunct}\relax
\EndOfBibitem
\bibitem[Park \latin{et~al.}(2024)Park, Gill, Moosavi, and Kim]{D3TA06274K}
Park,~J.; Gill,~A. P.~S.; Moosavi,~S.~M.; Kim,~J. Inverse design of porous materials: a diffusion model approach. \emph{J. Mater. Chem. A} \textbf{2024}, \emph{12}, 6507--6514\relax
\mciteBstWouldAddEndPuncttrue
\mciteSetBstMidEndSepPunct{\mcitedefaultmidpunct}
{\mcitedefaultendpunct}{\mcitedefaultseppunct}\relax
\EndOfBibitem
\bibitem[Kotsias \latin{et~al.}(2020)Kotsias, Ar{\'u}s-Pous, Chen, Engkvist, Tyrchan, and Bjerrum]{kotsias2020direct}
Kotsias,~P.-C.; Ar{\'u}s-Pous,~J.; Chen,~H.; Engkvist,~O.; Tyrchan,~C.; Bjerrum,~E.~J. Direct steering of de novo molecular generation with descriptor conditional recurrent neural networks. \emph{Nature Machine Intelligence} \textbf{2020}, \emph{2}, 254--265\relax
\mciteBstWouldAddEndPuncttrue
\mciteSetBstMidEndSepPunct{\mcitedefaultmidpunct}
{\mcitedefaultendpunct}{\mcitedefaultseppunct}\relax
\EndOfBibitem
\bibitem[Grisoni \latin{et~al.}(2020)Grisoni, Moret, Lingwood, and Schneider]{doi:10.1021/acs.jcim.9b00943}
Grisoni,~F.; Moret,~M.; Lingwood,~R.; Schneider,~G. Bidirectional Molecule Generation with Recurrent Neural Networks. \emph{Journal of Chemical Information and Modeling} \textbf{2020}, \emph{60}, 1175--1183, PMID: 31904964\relax
\mciteBstWouldAddEndPuncttrue
\mciteSetBstMidEndSepPunct{\mcitedefaultmidpunct}
{\mcitedefaultendpunct}{\mcitedefaultseppunct}\relax
\EndOfBibitem
\bibitem[Yao \latin{et~al.}(2021)Yao, S{\'a}nchez-Lengeling, Bobbitt, Bucior, Kumar, Collins, Burns, Woo, Farha, Snurr, \latin{et~al.} others]{yao2021inverse}
Yao,~Z.; S{\'a}nchez-Lengeling,~B.; Bobbitt,~N.~S.; Bucior,~B.~J.; Kumar,~S. G.~H.; Collins,~S.~P.; Burns,~T.; Woo,~T.~K.; Farha,~O.~K.; Snurr,~R.~Q.; others Inverse design of nanoporous crystalline reticular materials with deep generative models. \emph{Nature Machine Intelligence} \textbf{2021}, \emph{3}, 76--86\relax
\mciteBstWouldAddEndPuncttrue
\mciteSetBstMidEndSepPunct{\mcitedefaultmidpunct}
{\mcitedefaultendpunct}{\mcitedefaultseppunct}\relax
\EndOfBibitem
\bibitem[Che \latin{et~al.}(2016)Che, Li, Jacob, Bengio, and Li]{Che2016ModeRG}
Che,~T.; Li,~Y.; Jacob,~A.~P.; Bengio,~Y.; Li,~W. Mode Regularized Generative Adversarial Networks. \emph{ArXiv} \textbf{2016}, \emph{abs/1612.02136}\relax
\mciteBstWouldAddEndPuncttrue
\mciteSetBstMidEndSepPunct{\mcitedefaultmidpunct}
{\mcitedefaultendpunct}{\mcitedefaultseppunct}\relax
\EndOfBibitem
\bibitem[Muenkler \latin{et~al.}(2023)Muenkler, Misztela, Pikusa, Segler, Schneider, and Maziarz]{muenkler2023vaesbadreconstructingmolecular}
Muenkler,~H.; Misztela,~H.; Pikusa,~M.; Segler,~M.; Schneider,~N.; Maziarz,~K. Are VAEs Bad at Reconstructing Molecular Graphs? 2023; \url{https://arxiv.org/abs/2305.03041}\relax
\mciteBstWouldAddEndPuncttrue
\mciteSetBstMidEndSepPunct{\mcitedefaultmidpunct}
{\mcitedefaultendpunct}{\mcitedefaultseppunct}\relax
\EndOfBibitem
\bibitem[Moosavi \latin{et~al.}(2020)Moosavi, Nandy, Jablonka, Ongari, Janet, Boyd, Lee, Smit, and Kulik]{moosavi2020understanding}
Moosavi,~S.~M.; Nandy,~A.; Jablonka,~K.~M.; Ongari,~D.; Janet,~J.~P.; Boyd,~P.~G.; Lee,~Y.; Smit,~B.; Kulik,~H.~J. Understanding the diversity of the metal-organic framework ecosystem. \emph{Nature communications} \textbf{2020}, \emph{11}, 4068\relax
\mciteBstWouldAddEndPuncttrue
\mciteSetBstMidEndSepPunct{\mcitedefaultmidpunct}
{\mcitedefaultendpunct}{\mcitedefaultseppunct}\relax
\EndOfBibitem
\bibitem[Park \latin{et~al.}(2024)Park, Li, and Walsh]{PARK20242355}
Park,~H.; Li,~Z.; Walsh,~A. Has generative artificial intelligence solved inverse materials design? \emph{Matter} \textbf{2024}, \emph{7}, 2355--2367\relax
\mciteBstWouldAddEndPuncttrue
\mciteSetBstMidEndSepPunct{\mcitedefaultmidpunct}
{\mcitedefaultendpunct}{\mcitedefaultseppunct}\relax
\EndOfBibitem
\bibitem[Vaswani \latin{et~al.}(2017)Vaswani, Shazeer, Parmar, Uszkoreit, Jones, Gomez, Kaiser, and Polosukhin]{vaswani2017attention}
Vaswani,~A.; Shazeer,~N.; Parmar,~N.; Uszkoreit,~J.; Jones,~L.; Gomez,~A.~N.; Kaiser,~{\L}.; Polosukhin,~I. Attention is all you need. \emph{Advances in neural information processing systems} \textbf{2017}, \emph{30}\relax
\mciteBstWouldAddEndPuncttrue
\mciteSetBstMidEndSepPunct{\mcitedefaultmidpunct}
{\mcitedefaultendpunct}{\mcitedefaultseppunct}\relax
\EndOfBibitem
\bibitem[Radford \latin{et~al.}(2019)Radford, Wu, Child, Luan, Amodei, and Sutskever]{radford2019language}
Radford,~A.; Wu,~J.; Child,~R.; Luan,~D.; Amodei,~D.; Sutskever,~I. Language Models are Unsupervised Multitask Learners. \textbf{2019}, \relax
\mciteBstWouldAddEndPunctfalse
\mciteSetBstMidEndSepPunct{\mcitedefaultmidpunct}
{}{\mcitedefaultseppunct}\relax
\EndOfBibitem
\bibitem[Bagal \latin{et~al.}(2022)Bagal, Aggarwal, Vinod, and Priyakumar]{doi:10.1021/acs.jcim.1c00600}
Bagal,~V.; Aggarwal,~R.; Vinod,~P.~K.; Priyakumar,~U.~D. MolGPT: Molecular Generation Using a Transformer-Decoder Model. \emph{Journal of Chemical Information and Modeling} \textbf{2022}, \emph{62}, 2064--2076, PMID: 34694798\relax
\mciteBstWouldAddEndPuncttrue
\mciteSetBstMidEndSepPunct{\mcitedefaultmidpunct}
{\mcitedefaultendpunct}{\mcitedefaultseppunct}\relax
\EndOfBibitem
\bibitem[Bucior \latin{et~al.}(2019)Bucior, Rosen, Haranczyk, Yao, Ziebel, Farha, Hupp, Siepmann, Aspuru-Guzik, and Snurr]{doi:10.1021/acs.cgd.9b01050}
Bucior,~B.~J.; Rosen,~A.~S.; Haranczyk,~M.; Yao,~Z.; Ziebel,~M.~E.; Farha,~O.~K.; Hupp,~J.~T.; Siepmann,~J.~I.; Aspuru-Guzik,~A.; Snurr,~R.~Q. Identification Schemes for Metal–Organic Frameworks To Enable Rapid Search and Cheminformatics Analysis. \emph{Crystal Growth \& Design} \textbf{2019}, \emph{19}, 6682--6697\relax
\mciteBstWouldAddEndPuncttrue
\mciteSetBstMidEndSepPunct{\mcitedefaultmidpunct}
{\mcitedefaultendpunct}{\mcitedefaultseppunct}\relax
\EndOfBibitem
\bibitem[Weininger(1988)]{doi:10.1021/ci00057a005}
Weininger,~D. SMILES, a chemical language and information system. 1. Introduction to methodology and encoding rules. \emph{Journal of Chemical Information and Computer Sciences} \textbf{1988}, \emph{28}, 31--36\relax
\mciteBstWouldAddEndPuncttrue
\mciteSetBstMidEndSepPunct{\mcitedefaultmidpunct}
{\mcitedefaultendpunct}{\mcitedefaultseppunct}\relax
\EndOfBibitem
\bibitem[O’Keeffe \latin{et~al.}(2008)O’Keeffe, Peskov, Ramsden, and Yaghi]{doi:10.1021/ar800124u}
O’Keeffe,~M.; Peskov,~M.~A.; Ramsden,~S.~J.; Yaghi,~O.~M. The Reticular Chemistry Structure Resource (RCSR) Database of, and Symbols for, Crystal Nets. \emph{Accounts of Chemical Research} \textbf{2008}, \emph{41}, 1782--1789, PMID: 18834152\relax
\mciteBstWouldAddEndPuncttrue
\mciteSetBstMidEndSepPunct{\mcitedefaultmidpunct}
{\mcitedefaultendpunct}{\mcitedefaultseppunct}\relax
\EndOfBibitem
\bibitem[Kim \latin{et~al.}(2020)Kim, Lee, and Kim]{doi:10.1126/sciadv.aax9324}
Kim,~B.; Lee,~S.; Kim,~J. Inverse design of porous materials using artificial neural networks. \emph{Science Advances} \textbf{2020}, \emph{6}, eaax9324\relax
\mciteBstWouldAddEndPuncttrue
\mciteSetBstMidEndSepPunct{\mcitedefaultmidpunct}
{\mcitedefaultendpunct}{\mcitedefaultseppunct}\relax
\EndOfBibitem
\bibitem[DeepSeek-AI \latin{et~al.}(2025)DeepSeek-AI, Guo, Yang, Zhang, Song, Zhang, Xu, Zhu, Ma, Wang, Bi, Zhang, Yu, Wu, Wu, Gou, Shao, Li, Gao, Liu, Xue, Wang, Wu, Feng, Lu, Zhao, Deng, Zhang, Ruan, Dai, Chen, Ji, Li, Lin, Dai, Luo, Hao, Chen, Li, Zhang, Bao, Xu, Wang, Ding, Xin, Gao, Qu, Li, Guo, Li, Wang, Chen, Yuan, Qiu, Li, Cai, Ni, Liang, Chen, Dong, Hu, Gao, Guan, Huang, Yu, Wang, Zhang, Zhao, Wang, Zhang, Xu, Xia, Zhang, Zhang, Tang, Li, Wang, Li, Tian, Huang, Zhang, Wang, Chen, Du, Ge, Zhang, Pan, Wang, Chen, Jin, Chen, Lu, Zhou, Chen, Ye, Wang, Yu, Zhou, Pan, Li, Zhou, Wu, Ye, Yun, Pei, Sun, Wang, Zeng, Zhao, Liu, Liang, Gao, Yu, Zhang, Xiao, An, Liu, Wang, Chen, Nie, Cheng, Liu, Xie, Liu, Yang, Li, Su, Lin, Li, Jin, Shen, Chen, Sun, Wang, Song, Zhou, Wang, Shan, Li, Wang, Wei, Zhang, Xu, Li, Zhao, Sun, Wang, Yu, Zhang, Shi, Xiong, He, Piao, Wang, Tan, Ma, Liu, Guo, Ou, Wang, Gong, Zou, He, Xiong, Luo, You, Liu, Zhou, Zhu, Xu, Huang, Li, Zheng, Zhu, Ma, Tang, Zha, Yan, Ren, Ren, Sha, Fu, Xu, Xie,
  Zhang, Hao, Ma, Yan, Wu, Gu, Zhu, Liu, Li, Xie, Song, Pan, Huang, Xu, Zhang, and Zhang]{deepseekai2025deepseekr1incentivizingreasoningcapability}
DeepSeek-AI \latin{et~al.}  DeepSeek-R1: Incentivizing Reasoning Capability in LLMs via Reinforcement Learning. 2025; \url{https://arxiv.org/abs/2501.12948}\relax
\mciteBstWouldAddEndPuncttrue
\mciteSetBstMidEndSepPunct{\mcitedefaultmidpunct}
{\mcitedefaultendpunct}{\mcitedefaultseppunct}\relax
\EndOfBibitem
\bibitem[Putin \latin{et~al.}(2018)Putin, Asadulaev, Ivanenkov, Aladinskiy, Sanchez-Lengeling, Aspuru-Guzik, and Zhavoronkov]{doi:10.1021/acs.jcim.7b00690}
Putin,~E.; Asadulaev,~A.; Ivanenkov,~Y.; Aladinskiy,~V.; Sanchez-Lengeling,~B.; Aspuru-Guzik,~A.; Zhavoronkov,~A. Reinforced Adversarial Neural Computer for de Novo Molecular Design. \emph{Journal of Chemical Information and Modeling} \textbf{2018}, \emph{58}, 1194--1204, PMID: 29762023\relax
\mciteBstWouldAddEndPuncttrue
\mciteSetBstMidEndSepPunct{\mcitedefaultmidpunct}
{\mcitedefaultendpunct}{\mcitedefaultseppunct}\relax
\EndOfBibitem
\bibitem[Popova \latin{et~al.}(2018)Popova, Isayev, and Tropsha]{doi:10.1126/sciadv.aap7885}
Popova,~M.; Isayev,~O.; Tropsha,~A. Deep reinforcement learning for de novo drug design. \emph{Science Advances} \textbf{2018}, \emph{4}, eaap7885\relax
\mciteBstWouldAddEndPuncttrue
\mciteSetBstMidEndSepPunct{\mcitedefaultmidpunct}
{\mcitedefaultendpunct}{\mcitedefaultseppunct}\relax
\EndOfBibitem
\bibitem[Mazuz \latin{et~al.}(2023)Mazuz, Shtar, Shapira, and Rokach]{mazuz2023molecule}
Mazuz,~E.; Shtar,~G.; Shapira,~B.; Rokach,~L. Molecule generation using transformers and policy gradient reinforcement learning. \emph{Scientific Reports} \textbf{2023}, \emph{13}, 8799\relax
\mciteBstWouldAddEndPuncttrue
\mciteSetBstMidEndSepPunct{\mcitedefaultmidpunct}
{\mcitedefaultendpunct}{\mcitedefaultseppunct}\relax
\EndOfBibitem
\bibitem[Pan \latin{et~al.}(2022)Pan, Karpovich, and Olivetti]{pan2022deep}
Pan,~E.; Karpovich,~C.; Olivetti,~E. Deep reinforcement learning for inverse inorganic materials design. \emph{arXiv preprint arXiv:2210.11931} \textbf{2022}, \relax
\mciteBstWouldAddEndPunctfalse
\mciteSetBstMidEndSepPunct{\mcitedefaultmidpunct}
{}{\mcitedefaultseppunct}\relax
\EndOfBibitem
\bibitem[Park \latin{et~al.}(2024)Park, Majumdar, Zhang, Kim, and Smit]{park2024inverse}
Park,~H.; Majumdar,~S.; Zhang,~X.; Kim,~J.; Smit,~B. Inverse design of metal--organic frameworks for direct air capture of CO 2 via deep reinforcement learning. \emph{Digital Discovery} \textbf{2024}, \emph{3}, 728--741\relax
\mciteBstWouldAddEndPuncttrue
\mciteSetBstMidEndSepPunct{\mcitedefaultmidpunct}
{\mcitedefaultendpunct}{\mcitedefaultseppunct}\relax
\EndOfBibitem
\bibitem[Zheng \latin{et~al.}(2023)Zheng, Rong, Rampal, Borgs, Chayes, and Yaghi]{https://doi.org/10.1002/anie.202311983}
Zheng,~Z.; Rong,~Z.; Rampal,~N.; Borgs,~C.; Chayes,~J.~T.; Yaghi,~O.~M. A GPT-4 Reticular Chemist for Guiding MOF Discovery. \emph{Angewandte Chemie International Edition} \textbf{2023}, \emph{62}, e202311983\relax
\mciteBstWouldAddEndPuncttrue
\mciteSetBstMidEndSepPunct{\mcitedefaultmidpunct}
{\mcitedefaultendpunct}{\mcitedefaultseppunct}\relax
\EndOfBibitem
\bibitem[Kang and Kim(2024)Kang, and Kim]{kang2024chatmof}
Kang,~Y.; Kim,~J. ChatMOF: an artificial intelligence system for predicting and generating metal-organic frameworks using large language models. \emph{Nature Communications} \textbf{2024}, \emph{15}, 4705\relax
\mciteBstWouldAddEndPuncttrue
\mciteSetBstMidEndSepPunct{\mcitedefaultmidpunct}
{\mcitedefaultendpunct}{\mcitedefaultseppunct}\relax
\EndOfBibitem
\bibitem[Rosen \latin{et~al.}(2021)Rosen, Iyer, Ray, Yao, Aspuru-Guzik, Gagliardi, Notestein, and Snurr]{ROSEN20211578}
Rosen,~A.~S.; Iyer,~S.~M.; Ray,~D.; Yao,~Z.; Aspuru-Guzik,~A.; Gagliardi,~L.; Notestein,~J.~M.; Snurr,~R.~Q. Machine learning the quantum-chemical properties of metal–organic frameworks for accelerated materials discovery. \emph{Matter} \textbf{2021}, \emph{4}, 1578--1597\relax
\mciteBstWouldAddEndPuncttrue
\mciteSetBstMidEndSepPunct{\mcitedefaultmidpunct}
{\mcitedefaultendpunct}{\mcitedefaultseppunct}\relax
\EndOfBibitem
\bibitem[Rosen \latin{et~al.}(2022)Rosen, Fung, Huck, O’Donnell, Horton, Truhlar, Persson, Notestein, and Snurr]{rosen2022high}
Rosen,~A.~S.; Fung,~V.; Huck,~P.; O’Donnell,~C.~T.; Horton,~M.~K.; Truhlar,~D.~G.; Persson,~K.~A.; Notestein,~J.~M.; Snurr,~R.~Q. High-throughput predictions of metal--organic framework electronic properties: theoretical challenges, graph neural networks, and data exploration. \emph{npj Computational Materials} \textbf{2022}, \emph{8}, 1--10\relax
\mciteBstWouldAddEndPuncttrue
\mciteSetBstMidEndSepPunct{\mcitedefaultmidpunct}
{\mcitedefaultendpunct}{\mcitedefaultseppunct}\relax
\EndOfBibitem
\bibitem[Sutton \latin{et~al.}(1999)Sutton, McAllester, Singh, and Mansour]{NIPS1999_464d828b}
Sutton,~R.~S.; McAllester,~D.; Singh,~S.; Mansour,~Y. Policy Gradient Methods for Reinforcement Learning with Function Approximation. Advances in Neural Information Processing Systems. 1999\relax
\mciteBstWouldAddEndPuncttrue
\mciteSetBstMidEndSepPunct{\mcitedefaultmidpunct}
{\mcitedefaultendpunct}{\mcitedefaultseppunct}\relax
\EndOfBibitem
\bibitem[Ahmed \latin{et~al.}(2019)Ahmed, Seth, Purewal, Wong-Foy, Veenstra, Matzger, and Siegel]{ahmed2019exceptional}
Ahmed,~A.; Seth,~S.; Purewal,~J.; Wong-Foy,~A.~G.; Veenstra,~M.; Matzger,~A.~J.; Siegel,~D.~J. Exceptional hydrogen storage achieved by screening nearly half a million metal-organic frameworks. \emph{Nature communications} \textbf{2019}, \emph{10}, 1568\relax
\mciteBstWouldAddEndPuncttrue
\mciteSetBstMidEndSepPunct{\mcitedefaultmidpunct}
{\mcitedefaultendpunct}{\mcitedefaultseppunct}\relax
\EndOfBibitem
\bibitem[Chung \latin{et~al.}(2014)Chung, Camp, Haranczyk, Sikora, Bury, Krungleviciute, Yildirim, Farha, Sholl, and Snurr]{chung2014computation}
Chung,~Y.~G.; Camp,~J.; Haranczyk,~M.; Sikora,~B.~J.; Bury,~W.; Krungleviciute,~V.; Yildirim,~T.; Farha,~O.~K.; Sholl,~D.~S.; Snurr,~R.~Q. Computation-ready, experimental metal--organic frameworks: A tool to enable high-throughput screening of nanoporous crystals. \emph{Chemistry of Materials} \textbf{2014}, \emph{26}, 6185--6192\relax
\mciteBstWouldAddEndPuncttrue
\mciteSetBstMidEndSepPunct{\mcitedefaultmidpunct}
{\mcitedefaultendpunct}{\mcitedefaultseppunct}\relax
\EndOfBibitem
\bibitem[Mason \latin{et~al.}(2011)Mason, Sumida, Herm, Krishna, and Long]{mason2011evaluating}
Mason,~J.~A.; Sumida,~K.; Herm,~Z.~R.; Krishna,~R.; Long,~J.~R. Evaluating metal--organic frameworks for post-combustion carbon dioxide capture via temperature swing adsorption. \emph{Energy \& Environmental Science} \textbf{2011}, \emph{4}, 3030--3040\relax
\mciteBstWouldAddEndPuncttrue
\mciteSetBstMidEndSepPunct{\mcitedefaultmidpunct}
{\mcitedefaultendpunct}{\mcitedefaultseppunct}\relax
\EndOfBibitem
\bibitem[Sun \latin{et~al.}(2017)Sun, Hendon, Park, Tulchinsky, Wan, Wang, Walsh, and Dinc{\u{a}}]{sun2017iron}
Sun,~L.; Hendon,~C.~H.; Park,~S.~S.; Tulchinsky,~Y.; Wan,~R.; Wang,~F.; Walsh,~A.; Dinc{\u{a}},~M. Is iron unique in promoting electrical conductivity in MOFs? \emph{Chemical science} \textbf{2017}, \emph{8}, 4450--4457\relax
\mciteBstWouldAddEndPuncttrue
\mciteSetBstMidEndSepPunct{\mcitedefaultmidpunct}
{\mcitedefaultendpunct}{\mcitedefaultseppunct}\relax
\EndOfBibitem
\bibitem[Pham \latin{et~al.}(2014)Pham, Mai, Pham-Tran, Kawazoe, Mizuseki, and Nguyen-Manh]{pham2014engineering}
Pham,~H.~Q.; Mai,~T.; Pham-Tran,~N.-N.; Kawazoe,~Y.; Mizuseki,~H.; Nguyen-Manh,~D. Engineering of band gap in metal--organic frameworks by functionalizing organic linker: A systematic density functional theory investigation. \emph{The Journal of Physical Chemistry C} \textbf{2014}, \emph{118}, 4567--4577\relax
\mciteBstWouldAddEndPuncttrue
\mciteSetBstMidEndSepPunct{\mcitedefaultmidpunct}
{\mcitedefaultendpunct}{\mcitedefaultseppunct}\relax
\EndOfBibitem
\bibitem[Narayan \latin{et~al.}(2012)Narayan, Miyakai, Seki, and Dincă]{narayan2012high}
Narayan,~T.~C.; Miyakai,~T.; Seki,~S.; Dincă,~M. High charge mobility in a tetrathiafulvalene-based microporous metal--organic framework. \emph{Journal of the American Chemical Society} \textbf{2012}, \emph{134}, 12932--12935\relax
\mciteBstWouldAddEndPuncttrue
\mciteSetBstMidEndSepPunct{\mcitedefaultmidpunct}
{\mcitedefaultendpunct}{\mcitedefaultseppunct}\relax
\EndOfBibitem
\end{mcitethebibliography}


\providecommand{\latin}[1]{#1}
\makeatletter
\providecommand{\doi}
  {\begingroup\let\do\@makeother\dospecials
  \catcode`\{=1 \catcode`\}=2 \doi@aux}
\providecommand{\doi@aux}[1]{\endgroup\texttt{#1}}
\makeatother
\providecommand*\mcitethebibliography{\thebibliography}
\csname @ifundefined\endcsname{endmcitethebibliography}  {\let\endmcitethebibliography\endthebibliography}{}
\begin{mcitethebibliography}{36}
\providecommand*\natexlab[1]{#1}
\providecommand*\mciteSetBstSublistMode[1]{}
\providecommand*\mciteSetBstMaxWidthForm[2]{}
\providecommand*\mciteBstWouldAddEndPuncttrue
  {\def\EndOfBibitem{\unskip.}}
\providecommand*\mciteBstWouldAddEndPunctfalse
  {\let\EndOfBibitem\relax}
\providecommand*\mciteSetBstMidEndSepPunct[3]{}
\providecommand*\mciteSetBstSublistLabelBeginEnd[3]{}
\providecommand*\EndOfBibitem{}
\mciteSetBstSublistMode{f}
\mciteSetBstMaxWidthForm{subitem}{(\alph{mcitesubitemcount})}
\mciteSetBstSublistLabelBeginEnd
  {\mcitemaxwidthsubitemform\space}
  {\relax}
  {\relax}

\bibitem[Bucior \latin{et~al.}(2019)Bucior, Rosen, Haranczyk, Yao, Ziebel, Farha, Hupp, Siepmann, Aspuru-Guzik, and Snurr]{doi:10.1021/acs.cgd.9b01050}
Bucior,~B.~J.; Rosen,~A.~S.; Haranczyk,~M.; Yao,~Z.; Ziebel,~M.~E.; Farha,~O.~K.; Hupp,~J.~T.; Siepmann,~J.~I.; Aspuru-Guzik,~A.; Snurr,~R.~Q. Identification Schemes for Metal–Organic Frameworks To Enable Rapid Search and Cheminformatics Analysis. \emph{Crystal Growth \& Design} \textbf{2019}, \emph{19}, 6682--6697\relax
\mciteBstWouldAddEndPuncttrue
\mciteSetBstMidEndSepPunct{\mcitedefaultmidpunct}
{\mcitedefaultendpunct}{\mcitedefaultseppunct}\relax
\EndOfBibitem
\bibitem[Weininger(1988)]{doi:10.1021/ci00057a005}
Weininger,~D. SMILES, a chemical language and information system. 1. Introduction to methodology and encoding rules. \emph{Journal of Chemical Information and Computer Sciences} \textbf{1988}, \emph{28}, 31--36\relax
\mciteBstWouldAddEndPuncttrue
\mciteSetBstMidEndSepPunct{\mcitedefaultmidpunct}
{\mcitedefaultendpunct}{\mcitedefaultseppunct}\relax
\EndOfBibitem
\bibitem[O’Keeffe \latin{et~al.}(2008)O’Keeffe, Peskov, Ramsden, and Yaghi]{doi:10.1021/ar800124u}
O’Keeffe,~M.; Peskov,~M.~A.; Ramsden,~S.~J.; Yaghi,~O.~M. The Reticular Chemistry Structure Resource (RCSR) Database of, and Symbols for, Crystal Nets. \emph{Accounts of Chemical Research} \textbf{2008}, \emph{41}, 1782--1789, PMID: 18834152\relax
\mciteBstWouldAddEndPuncttrue
\mciteSetBstMidEndSepPunct{\mcitedefaultmidpunct}
{\mcitedefaultendpunct}{\mcitedefaultseppunct}\relax
\EndOfBibitem
\bibitem[Schwaller \latin{et~al.}(2019)Schwaller, Laino, Gaudin, Bolgar, Hunter, Bekas, and Lee]{doi:10.1021/acscentsci.9b00576}
Schwaller,~P.; Laino,~T.; Gaudin,~T.; Bolgar,~P.; Hunter,~C.~A.; Bekas,~C.; Lee,~A.~A. Molecular Transformer: A Model for Uncertainty-Calibrated Chemical Reaction Prediction. \emph{ACS Central Science} \textbf{2019}, \emph{5}, 1572--1583, PMID: 31572784\relax
\mciteBstWouldAddEndPuncttrue
\mciteSetBstMidEndSepPunct{\mcitedefaultmidpunct}
{\mcitedefaultendpunct}{\mcitedefaultseppunct}\relax
\EndOfBibitem
\bibitem[Schwaller \latin{et~al.}(2018)Schwaller, Gaudin, Lányi, Bekas, and Laino]{C8SC02339E}
Schwaller,~P.; Gaudin,~T.; Lányi,~D.; Bekas,~C.; Laino,~T. “Found in Translation”: predicting outcomes of complex organic chemistry reactions using neural sequence-to-sequence models. \emph{Chem. Sci.} \textbf{2018}, \emph{9}, 6091--6098\relax
\mciteBstWouldAddEndPuncttrue
\mciteSetBstMidEndSepPunct{\mcitedefaultmidpunct}
{\mcitedefaultendpunct}{\mcitedefaultseppunct}\relax
\EndOfBibitem
\bibitem[Schwaller \latin{et~al.}(2021)Schwaller, Probst, Vaucher, Nair, Kreutter, Laino, and Reymond]{schwaller2021mapping}
Schwaller,~P.; Probst,~D.; Vaucher,~A.~C.; Nair,~V.~H.; Kreutter,~D.; Laino,~T.; Reymond,~J.-L. Mapping the space of chemical reactions using attention-based neural networks. \emph{Nature Machine Intelligence} \textbf{2021}, \emph{3}, 144--152\relax
\mciteBstWouldAddEndPuncttrue
\mciteSetBstMidEndSepPunct{\mcitedefaultmidpunct}
{\mcitedefaultendpunct}{\mcitedefaultseppunct}\relax
\EndOfBibitem
\bibitem[Devlin \latin{et~al.}(2019)Devlin, Chang, Lee, and Toutanova]{devlin2019bertpretrainingdeepbidirectional}
Devlin,~J.; Chang,~M.-W.; Lee,~K.; Toutanova,~K. BERT: Pre-training of Deep Bidirectional Transformers for Language Understanding. 2019; \url{https://arxiv.org/abs/1810.04805}\relax
\mciteBstWouldAddEndPuncttrue
\mciteSetBstMidEndSepPunct{\mcitedefaultmidpunct}
{\mcitedefaultendpunct}{\mcitedefaultseppunct}\relax
\EndOfBibitem
\bibitem[Boyd \latin{et~al.}(2019)Boyd, Chidambaram, Garc{\'\i}a-D{\'\i}ez, Ireland, Daff, Bounds, G{\l}adysiak, Schouwink, Moosavi, Maroto-Valer, \latin{et~al.} others]{boyd2019data}
Boyd,~P.~G.; Chidambaram,~A.; Garc{\'\i}a-D{\'\i}ez,~E.; Ireland,~C.~P.; Daff,~T.~D.; Bounds,~R.; G{\l}adysiak,~A.; Schouwink,~P.; Moosavi,~S.~M.; Maroto-Valer,~M.~M.; others Data-driven design of metal--organic frameworks for wet flue gas CO2 capture. \emph{Nature} \textbf{2019}, \emph{576}, 253--256\relax
\mciteBstWouldAddEndPuncttrue
\mciteSetBstMidEndSepPunct{\mcitedefaultmidpunct}
{\mcitedefaultendpunct}{\mcitedefaultseppunct}\relax
\EndOfBibitem
\bibitem[Rosen \latin{et~al.}(2021)Rosen, Iyer, Ray, Yao, Aspuru-Guzik, Gagliardi, Notestein, and Snurr]{ROSEN20211578}
Rosen,~A.~S.; Iyer,~S.~M.; Ray,~D.; Yao,~Z.; Aspuru-Guzik,~A.; Gagliardi,~L.; Notestein,~J.~M.; Snurr,~R.~Q. Machine learning the quantum-chemical properties of metal–organic frameworks for accelerated materials discovery. \emph{Matter} \textbf{2021}, \emph{4}, 1578--1597\relax
\mciteBstWouldAddEndPuncttrue
\mciteSetBstMidEndSepPunct{\mcitedefaultmidpunct}
{\mcitedefaultendpunct}{\mcitedefaultseppunct}\relax
\EndOfBibitem
\bibitem[Rosen \latin{et~al.}(2022)Rosen, Fung, Huck, O’Donnell, Horton, Truhlar, Persson, Notestein, and Snurr]{rosen2022high}
Rosen,~A.~S.; Fung,~V.; Huck,~P.; O’Donnell,~C.~T.; Horton,~M.~K.; Truhlar,~D.~G.; Persson,~K.~A.; Notestein,~J.~M.; Snurr,~R.~Q. High-throughput predictions of metal--organic framework electronic properties: theoretical challenges, graph neural networks, and data exploration. \emph{npj Computational Materials} \textbf{2022}, \emph{8}, 1--10\relax
\mciteBstWouldAddEndPuncttrue
\mciteSetBstMidEndSepPunct{\mcitedefaultmidpunct}
{\mcitedefaultendpunct}{\mcitedefaultseppunct}\relax
\EndOfBibitem
\bibitem[Wilmer \latin{et~al.}(2012)Wilmer, Leaf, Lee, Farha, Hauser, Hupp, and Snurr]{wilmer2012large}
Wilmer,~C.~E.; Leaf,~M.; Lee,~C.~Y.; Farha,~O.~K.; Hauser,~B.~G.; Hupp,~J.~T.; Snurr,~R.~Q. Large-scale screening of hypothetical metal--organic frameworks. \emph{Nature chemistry} \textbf{2012}, \emph{4}, 83--89\relax
\mciteBstWouldAddEndPuncttrue
\mciteSetBstMidEndSepPunct{\mcitedefaultmidpunct}
{\mcitedefaultendpunct}{\mcitedefaultseppunct}\relax
\EndOfBibitem
\bibitem[Xie \latin{et~al.}(2020)Xie, Skorupskii, and Dincă]{doi:10.1021/acs.chemrev.9b00766}
Xie,~L.~S.; Skorupskii,~G.; Dincă,~M. Electrically Conductive Metal–Organic Frameworks. \emph{Chemical Reviews} \textbf{2020}, \emph{120}, 8536--8580, PMID: 32275412\relax
\mciteBstWouldAddEndPuncttrue
\mciteSetBstMidEndSepPunct{\mcitedefaultmidpunct}
{\mcitedefaultendpunct}{\mcitedefaultseppunct}\relax
\EndOfBibitem
\bibitem[Sheberla \latin{et~al.}(2017)Sheberla, Bachman, Elias, Sun, Shao-Horn, and Dinc{\u{a}}]{sheberla2017conductive}
Sheberla,~D.; Bachman,~J.~C.; Elias,~J.~S.; Sun,~C.-J.; Shao-Horn,~Y.; Dinc{\u{a}},~M. Conductive MOF electrodes for stable supercapacitors with high areal capacitance. \emph{Nature materials} \textbf{2017}, \emph{16}, 220--224\relax
\mciteBstWouldAddEndPuncttrue
\mciteSetBstMidEndSepPunct{\mcitedefaultmidpunct}
{\mcitedefaultendpunct}{\mcitedefaultseppunct}\relax
\EndOfBibitem
\bibitem[Wang \latin{et~al.}(2022)Wang, Zhou, Zhou, and Sundmacher]{WANG2022107739}
Wang,~Z.; Zhou,~Y.; Zhou,~T.; Sundmacher,~K. Identification of optimal metal-organic frameworks by machine learning: Structure decomposition, feature integration, and predictive modeling. \emph{Computers \& Chemical Engineering} \textbf{2022}, \emph{160}, 107739\relax
\mciteBstWouldAddEndPuncttrue
\mciteSetBstMidEndSepPunct{\mcitedefaultmidpunct}
{\mcitedefaultendpunct}{\mcitedefaultseppunct}\relax
\EndOfBibitem
\bibitem[Altintas \latin{et~al.}(2021)Altintas, Altundal, Keskin, and Yildirim]{doi:10.1021/acs.jcim.1c00191}
Altintas,~C.; Altundal,~O.~F.; Keskin,~S.; Yildirim,~R. Machine Learning Meets with Metal Organic Frameworks for Gas Storage and Separation. \emph{Journal of Chemical Information and Modeling} \textbf{2021}, \emph{61}, 2131--2146, PMID: 33914526\relax
\mciteBstWouldAddEndPuncttrue
\mciteSetBstMidEndSepPunct{\mcitedefaultmidpunct}
{\mcitedefaultendpunct}{\mcitedefaultseppunct}\relax
\EndOfBibitem
\bibitem[Cao \latin{et~al.}(2023)Cao, Magar, Wang, and Barati~Farimani]{doi:10.1021/jacs.2c11420}
Cao,~Z.; Magar,~R.; Wang,~Y.; Barati~Farimani,~A. MOFormer: Self-Supervised Transformer Model for Metal–Organic Framework Property Prediction. \emph{Journal of the American Chemical Society} \textbf{2023}, \emph{145}, 2958--2967, PMID: 36706365\relax
\mciteBstWouldAddEndPuncttrue
\mciteSetBstMidEndSepPunct{\mcitedefaultmidpunct}
{\mcitedefaultendpunct}{\mcitedefaultseppunct}\relax
\EndOfBibitem
\bibitem[Nandy \latin{et~al.}(2023)Nandy, Yue, Oh, Duan, Terrones, Chung, and Kulik]{NANDY20231585}
Nandy,~A.; Yue,~S.; Oh,~C.; Duan,~C.; Terrones,~G.~G.; Chung,~Y.~G.; Kulik,~H.~J. A database of ultrastable MOFs reassembled from stable fragments with machine learning models. \emph{Matter} \textbf{2023}, \emph{6}, 1585--1603\relax
\mciteBstWouldAddEndPuncttrue
\mciteSetBstMidEndSepPunct{\mcitedefaultmidpunct}
{\mcitedefaultendpunct}{\mcitedefaultseppunct}\relax
\EndOfBibitem
\bibitem[Butova \latin{et~al.}(2016)Butova, Soldatov, Guda, Lomachenko, and Lamberti]{Butova_2016}
Butova,~V.~V.; Soldatov,~M.~A.; Guda,~A.~A.; Lomachenko,~K.~A.; Lamberti,~C. Metal-organic frameworks: structure, properties, methods of synthesis and characterization. \emph{Russian Chemical Reviews} \textbf{2016}, \emph{85}, 280\relax
\mciteBstWouldAddEndPuncttrue
\mciteSetBstMidEndSepPunct{\mcitedefaultmidpunct}
{\mcitedefaultendpunct}{\mcitedefaultseppunct}\relax
\EndOfBibitem
\bibitem[Putin \latin{et~al.}(2018)Putin, Asadulaev, Ivanenkov, Aladinskiy, Sanchez-Lengeling, Aspuru-Guzik, and Zhavoronkov]{doi:10.1021/acs.jcim.7b00690}
Putin,~E.; Asadulaev,~A.; Ivanenkov,~Y.; Aladinskiy,~V.; Sanchez-Lengeling,~B.; Aspuru-Guzik,~A.; Zhavoronkov,~A. Reinforced Adversarial Neural Computer for de Novo Molecular Design. \emph{Journal of Chemical Information and Modeling} \textbf{2018}, \emph{58}, 1194--1204, PMID: 29762023\relax
\mciteBstWouldAddEndPuncttrue
\mciteSetBstMidEndSepPunct{\mcitedefaultmidpunct}
{\mcitedefaultendpunct}{\mcitedefaultseppunct}\relax
\EndOfBibitem
\bibitem[Popova \latin{et~al.}(2018)Popova, Isayev, and Tropsha]{doi:10.1126/sciadv.aap7885}
Popova,~M.; Isayev,~O.; Tropsha,~A. Deep reinforcement learning for de novo drug design. \emph{Science Advances} \textbf{2018}, \emph{4}, eaap7885\relax
\mciteBstWouldAddEndPuncttrue
\mciteSetBstMidEndSepPunct{\mcitedefaultmidpunct}
{\mcitedefaultendpunct}{\mcitedefaultseppunct}\relax
\EndOfBibitem
\bibitem[Luong and Singh(2024)Luong, and Singh]{doi:10.1021/acs.jcim.3c02070}
Luong,~K.-D.; Singh,~A. Application of Transformers in Cheminformatics. \emph{Journal of Chemical Information and Modeling} \textbf{2024}, \emph{64}, 4392--4409, PMID: 38815246\relax
\mciteBstWouldAddEndPuncttrue
\mciteSetBstMidEndSepPunct{\mcitedefaultmidpunct}
{\mcitedefaultendpunct}{\mcitedefaultseppunct}\relax
\EndOfBibitem
\bibitem[Park \latin{et~al.}(2024)Park, Li, and Walsh]{PARK20242355}
Park,~H.; Li,~Z.; Walsh,~A. Has generative artificial intelligence solved inverse materials design? \emph{Matter} \textbf{2024}, \emph{7}, 2355--2367\relax
\mciteBstWouldAddEndPuncttrue
\mciteSetBstMidEndSepPunct{\mcitedefaultmidpunct}
{\mcitedefaultendpunct}{\mcitedefaultseppunct}\relax
\EndOfBibitem
\bibitem[Fu \latin{et~al.}(2024)Fu, Xie, Rosen, Jaakkola, and Smith]{fu2024mofdiff}
Fu,~X.; Xie,~T.; Rosen,~A.~S.; Jaakkola,~T.~S.; Smith,~J.~A. {MOFD}iff: Coarse-grained Diffusion for Metal-Organic Framework Design. The Twelfth International Conference on Learning Representations. 2024\relax
\mciteBstWouldAddEndPuncttrue
\mciteSetBstMidEndSepPunct{\mcitedefaultmidpunct}
{\mcitedefaultendpunct}{\mcitedefaultseppunct}\relax
\EndOfBibitem
\bibitem[Schulman \latin{et~al.}(2017)Schulman, Wolski, Dhariwal, Radford, and Klimov]{Schulman2017}
Schulman,~J.; Wolski,~F.; Dhariwal,~P.; Radford,~A.; Klimov,~O. Proximal Policy Optimization Algorithms. \emph{arXiv preprint arXiv:1707.06347} \textbf{2017}, \relax
\mciteBstWouldAddEndPunctfalse
\mciteSetBstMidEndSepPunct{\mcitedefaultmidpunct}
{}{\mcitedefaultseppunct}\relax
\EndOfBibitem
\bibitem[Haarnoja \latin{et~al.}(2018)Haarnoja, Zhou, Abbeel, and Levine]{Haarnoja2018}
Haarnoja,~T.; Zhou,~A.; Abbeel,~P.; Levine,~S. Soft Actor-Critic: Off-Policy Maximum Entropy Deep Reinforcement Learning with a Stochastic Actor. \emph{arXiv preprint arXiv:1801.01290} \textbf{2018}, \relax
\mciteBstWouldAddEndPunctfalse
\mciteSetBstMidEndSepPunct{\mcitedefaultmidpunct}
{}{\mcitedefaultseppunct}\relax
\EndOfBibitem
\bibitem[Park \latin{et~al.}(2024)Park, Majumdar, Zhang, Kim, and Smit]{park2024inverse}
Park,~H.; Majumdar,~S.; Zhang,~X.; Kim,~J.; Smit,~B. Inverse design of metal--organic frameworks for direct air capture of CO 2 via deep reinforcement learning. \emph{Digital Discovery} \textbf{2024}, \emph{3}, 728--741\relax
\mciteBstWouldAddEndPuncttrue
\mciteSetBstMidEndSepPunct{\mcitedefaultmidpunct}
{\mcitedefaultendpunct}{\mcitedefaultseppunct}\relax
\EndOfBibitem
\bibitem[Majumdar \latin{et~al.}(2021)Majumdar, Moosavi, Jablonka, Ongari, and Smit]{doi:10.1021/acsami.1c16220}
Majumdar,~S.; Moosavi,~S.~M.; Jablonka,~K.~M.; Ongari,~D.; Smit,~B. Diversifying Databases of Metal Organic Frameworks for High-Throughput Computational Screening. \emph{ACS Applied Materials \& Interfaces} \textbf{2021}, \emph{13}, 61004--61014, PMID: 34910455\relax
\mciteBstWouldAddEndPuncttrue
\mciteSetBstMidEndSepPunct{\mcitedefaultmidpunct}
{\mcitedefaultendpunct}{\mcitedefaultseppunct}\relax
\EndOfBibitem
\bibitem[Li \latin{et~al.}(2016)Li, Wen, Cui, Zhou, Qian, and Chen]{https://doi.org/10.1002/adma.201601133}
Li,~B.; Wen,~H.-M.; Cui,~Y.; Zhou,~W.; Qian,~G.; Chen,~B. Emerging Multifunctional Metal–Organic Framework Materials. \emph{Advanced Materials} \textbf{2016}, \emph{28}, 8819--8860\relax
\mciteBstWouldAddEndPuncttrue
\mciteSetBstMidEndSepPunct{\mcitedefaultmidpunct}
{\mcitedefaultendpunct}{\mcitedefaultseppunct}\relax
\EndOfBibitem
\bibitem[Zhang and Lin(2014)Zhang, and Lin]{C4CS00103F}
Zhang,~T.; Lin,~W. Metal–organic frameworks for artificial photosynthesis and photocatalysis. \emph{Chem. Soc. Rev.} \textbf{2014}, \emph{43}, 5982--5993\relax
\mciteBstWouldAddEndPuncttrue
\mciteSetBstMidEndSepPunct{\mcitedefaultmidpunct}
{\mcitedefaultendpunct}{\mcitedefaultseppunct}\relax
\EndOfBibitem
\bibitem[Furukawa \latin{et~al.}(2013)Furukawa, Cordova, O’Keeffe, and Yaghi]{doi:10.1126/science.1230444}
Furukawa,~H.; Cordova,~K.~E.; O’Keeffe,~M.; Yaghi,~O.~M. The Chemistry and Applications of Metal-Organic Frameworks. \emph{Science} \textbf{2013}, \emph{341}, 1230444\relax
\mciteBstWouldAddEndPuncttrue
\mciteSetBstMidEndSepPunct{\mcitedefaultmidpunct}
{\mcitedefaultendpunct}{\mcitedefaultseppunct}\relax
\EndOfBibitem
\bibitem[Jiao \latin{et~al.}(2019)Jiao, Seow, Skinner, Wang, and Jiang]{JIAO201943}
Jiao,~L.; Seow,~J. Y.~R.; Skinner,~W.~S.; Wang,~Z.~U.; Jiang,~H.-L. Metal–organic frameworks: Structures and functional applications. \emph{Materials Today} \textbf{2019}, \emph{27}, 43--68\relax
\mciteBstWouldAddEndPuncttrue
\mciteSetBstMidEndSepPunct{\mcitedefaultmidpunct}
{\mcitedefaultendpunct}{\mcitedefaultseppunct}\relax
\EndOfBibitem
\bibitem[Chen and Xu(2019)Chen, and Xu]{CHEN201957}
Chen,~L.; Xu,~Q. Metal-Organic Framework Composites for Catalysis. \emph{Matter} \textbf{2019}, \emph{1}, 57--89\relax
\mciteBstWouldAddEndPuncttrue
\mciteSetBstMidEndSepPunct{\mcitedefaultmidpunct}
{\mcitedefaultendpunct}{\mcitedefaultseppunct}\relax
\EndOfBibitem
\bibitem[He \latin{et~al.}(2014)He, Zhou, Qian, and Chen]{C4CS00032C}
He,~Y.; Zhou,~W.; Qian,~G.; Chen,~B. Methane storage in metal–organic frameworks. \emph{Chem. Soc. Rev.} \textbf{2014}, \emph{43}, 5657--5678\relax
\mciteBstWouldAddEndPuncttrue
\mciteSetBstMidEndSepPunct{\mcitedefaultmidpunct}
{\mcitedefaultendpunct}{\mcitedefaultseppunct}\relax
\EndOfBibitem
\bibitem[Allendorf and Stavila(2015)Allendorf, and Stavila]{Allendorf2015229}
Allendorf,~M.~D.; Stavila,~V. Crystal engineering, structure-function relationships, and the future of metal-organic frameworks. \emph{CrystEngComm} \textbf{2015}, \emph{17}, 229 – 246, Cited by: 233; All Open Access, Green Open Access\relax
\mciteBstWouldAddEndPuncttrue
\mciteSetBstMidEndSepPunct{\mcitedefaultmidpunct}
{\mcitedefaultendpunct}{\mcitedefaultseppunct}\relax
\EndOfBibitem
\bibitem[Safaei \latin{et~al.}(2019)Safaei, Foroughi, Ebrahimpoor, Jahani, Omidi, and Khatami]{SAFAEI2019401}
Safaei,~M.; Foroughi,~M.~M.; Ebrahimpoor,~N.; Jahani,~S.; Omidi,~A.; Khatami,~M. A review on metal-organic frameworks: Synthesis and applications. \emph{TrAC Trends in Analytical Chemistry} \textbf{2019}, \emph{118}, 401--425\relax
\mciteBstWouldAddEndPuncttrue
\mciteSetBstMidEndSepPunct{\mcitedefaultmidpunct}
{\mcitedefaultendpunct}{\mcitedefaultseppunct}\relax
\EndOfBibitem
\end{mcitethebibliography}

\end{document}

\section{Data Representation and Preprocessing}

\subsection{MOFid Representation and Tokenization}

Our work utilizes a specialized dual-component tokenizer designed to interpret Metal-Organic Framework identifiers (MOFids), which encapsulates both chemical and structural information in a concise format. This representation follows the approach introduced by Bucior et al.\cite{doi:10.1021/acs.cgd.9b01050}, where MOFid consists of two primary components: the SMILES notation\cite{doi:10.1021/ci00057a005} of secondary building units (SBUs) and the topology codes from the Reticular Chemistry Structure Resource (RCSR) database\cite{doi:10.1021/ar800124u}.

The tokenization process employs a two-stage pipeline to account for the distinct characteristics of chemical and topological information. For the SMILES component, we implemented a regex-based tokenizer derived from the molecular transformer work of Schwaller et al.\cite{doi:10.1021/acscentsci.9b00576, C8SC02339E}, which effectively captures atomic symbols, bonds, branches, and other chemical notation elements. This approach ensures precise tokenization of complex chemical structures within MOFs. The topology component undergoes separate processing through a specialized tokenizer that parses comma-separated topology codes and optional categorical information denoted after period delimiters.

Our vocabulary is derived from a comprehensive source originally developed for the USPTO dataset\cite{schwaller2021mapping}, which provides extensive coverage of chemical tokens while maintaining a manageable vocabulary size for efficient model training. The tokenizer manages special tokens including beginning-of-sequence, end-of-sequence, padding, masking, and unknown tokens to facilitate transformer-based sequence processing. Chemical and topological information are clearly delineated using a separator token "$\&\&$" between the components, preserving the logical structure of the MOFid representation. Table \ref{tab:mofid_examples} presents representative examples of MOFid representations for various types of MOFs, illustrating how the format captures both chemical composition through SMILES strings and topological information. Following the BERT approach\cite{devlin2019bertpretrainingdeepbidirectional}, our tokenizer adds a [CLS] token and a [SEP] token (equivalent to [BOS] and [EOS] in our implementation) at the beginning and end of the sequence, respectively, to symbolize the start and end of the MOF representation.

\begin{table}[h!]
\centering
\caption{Examples of MOFid Representations for Various Metal-Organic Frameworks}
\label{tab:mofid_examples}
\begin{tabular}{p{12cm}}
\hline
\textbf{Example 1:} Copper node with benzenedicarboxylate linker in primitive cubic topology \\
\textsf{C1=CC(=CC=C1C(=O)O)C(=O)O.Cu \&\& pcu} \\
\hline
\textbf{Example 2:} Copper-based MOF with mixed linkers (nitro-functionalized and unfunctionalized benzoates) \\
\textsf{c1cc(cc(c1)C(=O)O)N(=O)=O.c1cc(cc(c1)C(=O)O)C(=O)O.Cu \&\& nbo-d} \\
\hline
\textbf{Example 3:} Zinc-based MOF with benzonitrile linkers in diamondoid topology with catenation \\
\textsf{C1=CC=C(C=C1)C\#N.Zn \&\& dia-c} \\
\hline
\end{tabular}
\end{table}

All MOF representations are standardized to a maximum sequence length of 512 tokens through padding or truncation as needed. Our analysis confirmed that this approach effectively preserves structural information, with only 0.37\% of structures in the hMOF dataset exceeding the token limit. For these rare cases, the truncation preserves the most chemically relevant information while maintaining computational efficiency. During model training, appropriate attention masking techniques ensure that padding tokens do not contribute to the learned representations.

The tokenization pipeline supports batch processing with dynamic padding based on the longest sequence in each batch, optimizing computational efficiency during training. This approach transforms the MOF structural and chemical information into a format suitable for autoregressive sequence modeling while maintaining the critical relationships between building units and topological arrangements that define MOF functionality.

\subsection{Datasets and Property Representations}
\subsubsection{Dataset Composition and Training Split}
Our study aggregates MOF structures from three comprehensive, publicly available repositories for GPT pretraining purpose: Boyd \& Woo\cite{boyd2019data}, quantum MOF (QMOF)\cite{ROSEN20211578, rosen2022high}, and hypothetical MOF (hMOF)\cite{wilmer2012large} datasets. Following deduplication and shuffling procedures, we created a training set of 323,469 samples ($\sim$ 80\%) and a held-out test set of 81,260 samples ($\sim$ 20\%). Table \ref{tab:data_composition} provides the detailed composition of our dataset split across the three sources. The Boyd \& Woo dataset constitutes the majority of our corpus, while the hMOF dataset provides about a quarter of the samples, and the QMOF dataset contributes a smaller but significant portion. This diverse combination ensures our model learns from a broad spectrum of MOF architectures and chemical compositions.

\begin{table}[ht]
  \centering
  \caption{Composition of the pretraining (train) and test datasets by source.}
  \label{tab:data_composition}
  \begin{tabular}{lrrrr}
    \toprule
    \textbf{Dataset}    & \textbf{Train} & \textbf{Test} & \%Train & \%Test \\
    \midrule
    Boyd \& Woo         & 235,841 &  58,950 & 73.0\% & 72.5\% \\
    hMOF                &  82,114 &  20,744 & 25.4\% & 25.5\% \\
    QMOF                &   5,514 &   1,566 &  1.7\% &  1.9\% \\
    \midrule
    \textbf{Total}      & 323,469 &  81,260 & 100\%  & 100\%  \\
    \bottomrule
  \end{tabular}
\end{table}

\subsubsection{Property Data for Reinforcement Learning}

Our reinforcement learning framework leverages two complementary datasets with distinct property domains to enable comprehensive MOF design capabilities: the hMOF dataset for gas adsorption properties and the QMOF dataset for electronic properties.

The hMOF dataset contains gas adsorption property measurements across various conditions. Our dataset systematically captures both the target gas (CO$_2$ or CH$_4$) and the pressure conditions (ranging from 0.01 to 4.5 bar) at which adsorption properties were simulated. A key strength of this dataset lies in the fact that all adsorption properties were calculated using the same set of MOF structures but under different simulation conditions. This approach yielded 10 distinct property datasets (5 pressure points for each of the two gases) while maintaining structural consistency across all 82,114 training and 20,744 test MOF structures.

The pressure points were strategically selected to span the entire application spectrum for MOF-based gas systems: ultra-low pressures (0.01-0.05 bar) relevant for trace gas detection and sensing applications; medium pressures (0.1-0.9 bar) applicable to post-combustion carbon capture and environmental remediation; and higher pressures (2.5-4.5 bar) pertinent to gas storage and separation technologies. All adsorption values are reported in $mol.kg^{-1}$, with higher values indicating superior adsorption performance which is a critical consideration for applications such as carbon capture and gas storage where maximizing capacity is often a primary objective.

On the other hand, the QMOF dataset provides DFT-calculated band gap values measured in electron volts (eV) for each MOF structure\cite{doi:10.1021/acs.chemrev.9b00766, sheberla2017conductive}. Band gap is a critical electronic property in materials science, with lower values indicating better conductivity, which is a characteristic particularly valuable for energy storage applications and electronic devices. This dataset offers an opportunity to evaluate and optimize for electronic properties that involve complex quantum mechanical phenomena, broadening the application scope of our reinforcement learning framework beyond traditional gas adsorption targets.

Our comprehensive evaluation encompasses multiple property prediction tasks across different pressure conditions. Specifically, we utilize 5 CH$_4$ adsorption datasets at pressures of 0.05, 0.5, 0.9, 2.5, and 4.5 bar, 5 CO$_2$ adsorption datasets at pressures of 0.01, 0.05, 0.1, 0.5, and 2.5 bar, and 1 electronic band gap dataset from QMOF. For our primary results demonstration in the main text, we selected representative conditions that span low-pressure (0.05 bar CH$_4$, 0.01 bar CO$_2$), high-pressure (0.9 bar CH$_4$), and electronic properties (band gap) to showcase the framework's versatility across different physical regimes and property types. The selection of these specific conditions was strategically made to represent practically relevant pressure regimes: 0.05 bar for CH$_4$ represents low-pressure storage applications relevant to natural gas vehicles, 0.9 bar captures near-atmospheric conditions important for industrial separation processes, and 0.01 bar for CO$_2$ represents trace CO$_2$ capture scenarios critical for atmospheric remediation applications. Complete results for all 11 property prediction tasks are presented in Appendix A, demonstrating consistent performance trends with validity rates maintained above 35\% and novelty rates exceeding 60\% across all conditions.

Together, these datasets provide diverse objective functions for our reinforcement learning framework, enabling the generation of MOF structures with enhanced performance across multiple application domains. For all benchmark datasets used in targeted property generation, we followed established protocols by splitting them into training, validation, and test sets with a ratio of 0.8:0.05:0.15.

\section{Model Architecture and Training Methodology}

\subsection{Base Language Model}
Our MOFGPT framework utilizes a transformer-based architecture derived from the GPT-2 model, adapted specifically for processing MOFid representations. The model consists of 12 transformer decoder layers with an embedding dimension of 768, 12 attention heads, and a feed-forward dimension of 3072. The architecture follows the standard transformer decoder design with self-attention mechanisms, feed-forward neural networks, and residual connections. The transformer decoder employs a self-attention mechanism defined by:

\begin{equation}
\text{Attention}(Q, K, V) = \text{softmax}\left(\frac{QK^T}{\sqrt{d_k}} + M\right)V
\end{equation}

where $Q$, $K$, and $V$ are the query, key, and value matrices derived from the input embeddings, $d_k$ is the dimension of the key vectors, and $M$ is the attention mask that prevents attending to future tokens during training (causal attention). Each transformer layer incorporates a multi-head attention mechanism and position-wise feed-forward network with residual connections and layer normalization. The complete model configuration parameters are detailed in Table \ref{tab:model_config}.

\begin{table}[h!]
\centering
\caption{Base Model Configuration Parameters}
\label{tab:model_config}
\begin{tabular}{lc}
\hline
\textbf{Parameter} & \textbf{Value} \\
\hline
Number of layers & 12 \\
Hidden dimension & 768 \\
Feed-forward dimension & 3072 \\
Number of attention heads & 12 \\
Maximum sequence length & 512 \\
Vocabulary size & 4023 \\
Dropout rate & 0.1 \\
Activation function & SiLU (Swish) \\
Layer normalization epsilon & 1e-5 \\
\hline
\end{tabular}
\end{table}

\subsection{Model Training Pipeline}
Our implementation follows a three-stage training approach:

\subsubsection{Pretraining}
We first pretrain the base language model using a next-token prediction objective on a large corpus of MOF structures. The pretraining objective is to maximize the log-likelihood of the next token given the previous tokens:

\begin{equation}
\mathcal{L}_{\text{pretrain}} = -\sum_{t=1}^{T} \log P_\theta(x_t | x_{<t})
\end{equation}

where $x_t$ is the token at position $t$, $x_{<t}$ represents all tokens before position $t$, and $P_\theta$ is the probability distribution over the vocabulary given by the model with parameters $\theta$.

\subsubsection{Fine-tuning}
After pretraining, we fine-tune the model for property prediction using supervised learning. We extend the base language model with a specialized regression head that processes the output embeddings from the transformer decoder through a series of feed-forward layers with nonlinear activations to extract property-relevant features. The fine-tuning objective is to minimize the mean squared error between the predicted and actual property values:

\begin{equation}
\mathcal{L}_{\text{finetune}} = \frac{1}{N} \sum_{i=1}^{N} (y_i - \hat{y}_i)^2
\end{equation}

where $y_i$ is the ground truth property value for the $i$-th MOF, $\hat{y}_i$ is the predicted value, and $N$ is the number of training examples. During fine-tuning, we employ several techniques to improve model performance, including learning rate scheduling with warm-up, gradient clipping, and early stopping based on validation loss.

\subsubsection{Reinforcement Learning}
The reinforcement learning phase optimizes the model to generate MOFs with specific target properties. We employ the REINFORCE algorithm (policy gradient method) with several enhancements to improve training stability and efficiency. The RL model consists of two main components:
(1) Policy network, which is the pretrained language model that defines the probability distribution over possible outputs at each step (2) Value network, which is the fine-tuned property prediction model that estimates the expected reward for a given MOF sequence. The RL objective is to maximize the expected reward:

\begin{equation}
J(\theta) = \mathbb{E}_{\tau \sim \pi_\theta} [R(\tau)]
\end{equation}

where $\pi_\theta$ is the policy defined by the model parameters $\theta$ and $R(\tau)$ is our multi-component reward function. The gradient of this objective can be estimated using the policy gradient theorem:

\begin{equation}
\nabla_\theta J(\theta) \approx \frac{1}{B} \sum_{i=1}^{B} \sum_{t=1}^{T_i} \nabla_\theta \log \pi_\theta(a_t^i|s_t^i) \cdot R(\tau^i) \cdot \gamma^{T_i-t}
\end{equation}

where $B$ is the batch size, $T_i$ is the length of the $i$-th sequence, $a_t^i$ is the token selected at step $t$ for sequence $i$, $s_t^i$ is the corresponding state, $R(\tau^i)$ is the reward for the complete sequence, and $\gamma$ is a discount factor that gives greater weight to rewards associated with earlier actions.

We implement several enhancements to the basic REINFORCE algorithm. Our training employs a global memory system that tracks the highest-performing MOF structures across all epochs, enabling the model to maintain awareness of promising chemical space regions without requiring explicit experience replay mechanisms. We add a KL divergence term to prevent the policy from deviating too far from the pretrained model, ensuring that generated structures remain chemically sensible:

\begin{equation}
\mathcal{L}_{\text{prox}} = \beta \cdot D_{\text{KL}}(\pi_\theta || \pi_{\theta_0})
\end{equation}

where $\beta$ is a weighting factor and $\pi_{\theta_0}$ is the initial policy from the pretrained model. We subtract a baseline from the rewards to reduce variance in gradient estimates:

\begin{equation}
\nabla_\theta J(\theta) \approx \frac{1}{B} \sum_{i=1}^{B} \sum_{t=1}^{T_i} \nabla_\theta \log \pi_\theta(a_t^i|s_t^i) \cdot (R(\tau^i) - b)
\end{equation}

where $b$ is a baseline value, often chosen as the mean reward of the batch. We employ temperature scheduling and top-k/top-p sampling to balance exploration and exploitation during generation. We gradually increase the influence of property-specific rewards as training progresses to prevent premature convergence to suboptimal solutions. The complete RL training algorithm is provided in Algorithm \ref{alg:rl_training}. The hyperparameters for each training stage are detailed in Table \ref{tab:detailed_hyperparams}. As shown in Table \ref{tab:detailed_hyperparams}, we used a smaller learning rate for the RL phase (1e-5) compared to pretraining (1e-3) to ensure stable policy updates while maintaining the model's pretrained knowledge.

\begin{table}[h!]
\centering
\caption{Detailed Model Hyperparameters for Different Training Phases}
\label{tab:detailed_hyperparams}
\begin{tabular}{lccc}
\hline
\textbf{Hyperparameter} & \textbf{Pretraining} & \textbf{Fine-tuning} & \textbf{RL Training} \\
\hline
Learning rate & 1e-3 & 1e-4 & 1e-5 \\
Batch size & 128 & 8 & 32 \\
Warmup ratio & 0.03 & 0.03 & 0.03 \\
Scheduler & Cosine & None & Cosine \\
Weight decay & 0.001 & 0.001 & 0.0001 \\
Gradient accumulation steps & 1 & 1 & 1 \\
Maximum sequence length & 512 & 512 & 512 \\
Dropout rate & 0.1 & 0.1 & 0.1 \\
Epochs & 30 & 30 & 60 \\
\hline
\textbf{RL-Specific Parameters} & & & \\
\hline
Discount factor ($\gamma$) & - & - & 0.99 \\
Novelty factor ($\alpha_{\text{nov}}$) & - & - & 1.5 \\
Validity factor ($\alpha_{\text{val}}$) & - & - & 2.5 \\
Diversity factor ($\alpha_{\text{div}}$) & - & - & 2.0 \\
Target weights ($\alpha_{\text{tgt}}$) & - & - & 3.0 \\
Evaluation interval & - & - & 5 epochs \\
\hline
\end{tabular}
\end{table}

\begin{algorithm}
\caption{Reinforcement Learning Training Procedure}
\label{alg:rl_training}
\begin{algorithmic}[1]
\STATE \textbf{Input:} Pretrained model $\pi_{\theta_0}$, property predictor $f$, target properties $p^*$
\STATE \textbf{Output:} Optimized model $\pi_\theta$
\STATE Initialize model parameters: $\theta \gets \theta_0$
\STATE Initialize global memory: $\mathcal{M}_{\text{global}} \gets \emptyset$
\FOR{epoch $e = 1$ to $E$}
    \STATE Generate batch of sequences $\{\tau_1, \tau_2, ..., \tau_B\} \sim \pi_\theta$
    \STATE Convert sequences to MOF representations $\{m_1, m_2, ..., m_B\}$
    \STATE Predict properties $\{\hat{p}_1, \hat{p}_2, ..., \hat{p}_B\}$ using predictor $f$
    \STATE Calculate rewards $\{R(\tau_1), R(\tau_2), ..., R(\tau_B)\}$ using multi-component reward function
    \STATE Update global memory $\mathcal{M}_{\text{global}}$ with top-performing structures
    \STATE Select top-K sequences based on target rewards for focused learning
    \STATE Compute policy gradients using REINFORCE algorithm
    \STATE Update model parameters $\theta$ using computed gradients
    \IF{evaluation interval reached}
        \STATE Evaluate model on validation set
        \STATE Save checkpoint if performance improved
    \ENDIF
\ENDFOR
\RETURN Optimized model $\pi_\theta$
\end{algorithmic}
\end{algorithm}

\section{Reward Function Design}

This section provides comprehensive details of our multi-component reward function that guides MOF generation toward structures with desired properties while maintaining chemical validity, novelty, and structural diversity. Our implementation incorporates global memory management and adaptive mechanisms to address key challenges in molecular generation such as mode collapse and maintaining exploration throughout training.

\subsection{Design Philosophy and Challenges}

The design of our multi-component reward function addresses fundamental challenges unique to MOF generative modeling that distinguish it from small molecule generation. While small organic molecules can usually be validated through established chemical rules, MOFs require additional consideration of both local coordination chemistry and global topological consistency. This challenge necessitated our comprehensive reward architecture that balances multiple competing objectives while maintaining focus on property targeting.

Traditional reward functions in molecular generation often suffer from several critical issues. First, focusing solely on target properties can cause mode collapse, where the model generates only a few high-reward structures repeatedly. Second, balancing exploration (novelty and diversity) with exploitation (target optimization) requires careful design. Third, maintaining learning signals for all generated structures while focusing on the most promising candidates is essential for stable training.

Our reward function addresses these challenges through three key innovations: (1) global memory that preserves high-performing structures across epochs, (2) multi-component diversity rewards that prevent mode collapse while maintaining focus on target properties, and (3) adaptive top-K selection that focuses learning on promising structures while maintaining gradient flow.

\subsection{Reward Architecture and Implementation}

Our reward function integrates four fundamental components, each addressing a specific aspect of successful molecular generation. The implementation uses a straightforward architecture that balances target property achievement with exploration and validity constraints. The total reward calculation follows a simple additive approach with multiplicative bonuses:

For structures selected in the top-K performers:
\begin{equation}
R_{\text{total}}(m) = R_{\text{target}}(m) \cdot \beta_{\text{base}} \cdot M_{\text{valid}}(m) \cdot M_{\text{novel}}(m) + R_{\text{novelty}}(m) + R_{\text{validity}}(m) + R_{\text{diversity}}(m)
\end{equation}

For structures not in the top-K:
\begin{equation}
R_{\text{total}}(m) = R_{\text{target}}(m) \cdot \alpha_{\text{reduced}}
\end{equation}

where $\beta_{\text{base}} = 3.0$ is a fixed base multiplier, $\alpha_{\text{reduced}} = 0.3$ is the reduced reward factor, and the multiplicative components are applied individually:

\begin{equation}
M_{\text{valid}}(m) = 
\begin{cases}
1.1 & \text{if structure is valid} \\
1.0 & \text{otherwise}
\end{cases}
\end{equation}

\begin{equation}
M_{\text{novel}}(m) = 
\begin{cases}
1.1 & \text{if structure is novel} \\
1.0 & \text{otherwise}
\end{cases}
\end{equation}

The additive bonus components provide small independent rewards for exploration:
\begin{align}
R_{\text{novelty}}(m) &= \text{novelty\_factor} \times \mathbb{1}_{\text{novel}}(m) \times 0.1 \\
R_{\text{validity}}(m) &= \text{validity\_factor} \times \mathbb{1}_{\text{valid}}(m) \times 0.1 \\
R_{\text{diversity}}(m) &= \text{diversity\_factor} \times S_{\text{diversity}}(m) \times 0.1
\end{align}

The top-K selection mechanism focuses learning on the most promising structures while maintaining gradient flow for all generated molecules. The selection ratio decreases from 50\% early in training to 30\% in later stages to become increasingly selective as the model improves.

\subsection{Core Reward Components}

\subsubsection{Target Property Reward}

The target property reward is the core component that drives optimization toward desired property values. This component receives the highest weighting because property optimization is the primary objective of the generation process:

\begin{equation}
R_{\text{target}}(m) = \sum_{i=1}^{k} w_i \cdot R_{\text{proximity}}(\hat{p}_i(m), T_i)
\end{equation}

where $\hat{p}_i(m)$ is the predicted value of property $i$ for structure $m$, $T_i$ is the target value for property $i$, and $w_i$ is the importance weight for property $i$.

The proximity reward calculation provides rewards based on relative distance to target:

\begin{equation}
R_{\text{proximity}}(\hat{p}, T) = R_{\text{base}}(\delta_{\text{rel}}) \cdot f_{\text{direction}}(\hat{p}, T) \cdot w
\end{equation}

where $\delta_{\text{rel}} = \frac{|\hat{p} - T|}{|T| + \epsilon}$ is the relative distance to target, with $\epsilon = 1 \times 10^{-6}$ for numerical stability. The base reward function provides a tiered reward structure that matches the implementation:

\begin{equation}
R_{\text{base}}(\delta_{\text{rel}}) = 
\begin{cases}
15.0 & \text{if } \delta_{\text{rel}} \leq 0.05 \\
12.0 & \text{if } 0.05 < \delta_{\text{rel}} \leq 0.1 \\
8.0 & \text{if } 0.1 < \delta_{\text{rel}} \leq 0.2 \\
4.0 & \text{if } 0.2 < \delta_{\text{rel}} \leq 0.5 \\
\max(1.0, 4.0(1 - \delta_{\text{rel}})) & \text{if } \delta_{\text{rel}} > 0.5
\end{cases}
\end{equation}

The direction factor provides incentives for optimization in the desired direction:

\begin{equation}
f_{\text{direction}}(\hat{p}, T) = 
\begin{cases}
1.3 & \text{if optimization achieved in correct direction} \\
1.1 & \text{if close to target (within 20\% buffer)} \\
0.95 & \text{otherwise}
\end{cases}
\end{equation}

For "higher" optimization mode, the achievement condition is $\hat{p} \geq T$, while for "lower" mode it is $\hat{p} \leq T$. The close-to-target condition applies when the prediction is within 80-120\% of the target value depending on the optimization direction.

\subsubsection{Validity Reward}

The validity reward ensures that generated structures adhere to chemical and structural constraints:

\begin{equation}
R_{\text{validity}}(m) = \mathbb{1}_{\text{valid}}(m) = 
\begin{cases} 
1 & \text{if valid according to validation procedure} \\
0 & \text{otherwise}
\end{cases}
\end{equation}

Our validation procedure consists of multiple checks designed specifically for MOF structures: SMILES syntax validation using RDKit with metal atom substitution for compatibility, metal node presence verification, structural balance confirmation requiring both organic and inorganic components, topology validity against known RCSR database entries when topology tokens are present, and coordination number validation for metal centers. The validity component appears both as a small additive bonus (scaled by 0.1) and as a multiplicative enhancement (factor of 1.1) when applied to top-K structures.

\subsubsection{Novelty Reward}

The novelty reward encourages exploration of new regions of chemical space rather than reproducing training data structures:

\begin{equation}
R_{\text{novelty}}(m) = \mathbb{1}_{\text{novel}}(m) = 
\begin{cases} 
1 & \text{if } m \notin \mathcal{D}_{\text{train}} \\
0 & \text{otherwise}
\end{cases}
\end{equation}

where $\mathcal{D}_{\text{train}}$ represents the set of MOF structures in the training dataset. Novelty checking involves comparing the generated MOFid representation with all training examples using exact string matching. Like validity, novelty also appears as both a small additive bonus (scaled by 0.1) and a multiplicative enhancement (factor of 1.1) for top-K structures.

\subsubsection{Diversity Reward}

The diversity reward prevents mode collapse by encouraging variety in generated structures. This component incorporates multiple diversity metrics to capture different aspects of structural variety:

\begin{equation}
R_{\text{diversity}}(m, \mathcal{B}, \mathcal{H}) = w_b \cdot S_{\text{batch}}(m, \mathcal{B}) + w_n \cdot S_{\text{ngram}}(m, \mathcal{B}) + w_h \cdot S_{\text{history}}(m, \mathcal{H}) + w_c \cdot S_{\text{composition}}(m)
\end{equation}

where $\mathcal{B}$ is the current batch, $\mathcal{H}$ is the generation history, and the weights are $w_b = 0.30$, $w_n = 0.25$, $w_h = 0.35$, and $w_c = 0.10$.

\textbf{Batch Diversity Component:} Measures structural differences between MOFs in the same generation batch using approximate distance metrics:

\begin{equation}
S_{\text{batch}}(m, \mathcal{B}) = \frac{2}{|\mathcal{B}| - 1} \sum_{m' \in \mathcal{B}, m' \neq m} d_{\text{approx}}(m, m')
\end{equation}

where $d_{\text{approx}}(m, m')$ combines string length differences and character-wise differences for computational efficiency.

\textbf{N-gram Diversity Component:} Analyzes character n-grams (length 4) to identify overused structural patterns:

\begin{equation}
S_{\text{ngram}}(m, \mathcal{B}) = \min\left(1.0, \frac{1.0}{\text{AvgFreq}(m, \mathcal{B}) / |\mathcal{B}| + \epsilon}\right)
\end{equation}

\textbf{Historical Uniqueness Component:} Prevents cycling by maintaining memory of previously generated structures:

\begin{equation}
S_{\text{history}}(m, \mathcal{H}) = 
\begin{cases}
0.1 & \text{if } m \in \mathcal{H} \text{ (exact duplicate)} \\
\frac{1}{\max(1, C_{\text{sig}}(m, \mathcal{H}))} & \text{otherwise}
\end{cases}
\end{equation}

where $C_{\text{sig}}(m, \mathcal{H})$ counts structures with similar signatures in the generation history.

\textbf{Compositional Diversity Component:} Encourages variety in chemical building blocks:

\begin{equation}
S_{\text{composition}}(m) = \min\left(1.0, 0.5 \cdot \frac{|\text{Elements}(m)|}{10} + 0.5 \cdot \frac{|\text{FunctionalGroups}(m)|}{5}\right)
\end{equation}

\subsection{Advanced Reward Mechanisms}

\subsubsection{Global Memory and Cross-Epoch Learning}

To prevent loss of high-performing structures and enable learning from past discoveries, we maintain a global memory of the best structures across all training epochs:

\begin{equation}
\mathcal{M}_{\text{global}} = \{(m_j, p_j, s_j, r_j)\}_{j=1}^{N_{\text{mem}}}
\end{equation}

where each entry contains a MOF structure $m_j$, its predicted properties $p_j$, target progress score $s_j$, and total reward $r_j$, with $N_{\text{mem}} = 200$. The target score provides a unified ranking metric that considers progress toward all objectives:

\begin{equation}
s_j = \sum_{i=1}^{k} w_i \cdot \text{Progress}_i(p_{j,i}, T_i)
\end{equation}

The progress function rewards both achievement and over-achievement in the desired direction:

\begin{equation}
\text{Progress}_i(p, T) = 
\begin{cases}
1.0 + \frac{p - T}{|T|} & \text{if mode="higher" and } p \geq T \\
\max(0.1, \frac{p}{T}) & \text{if mode="higher" and } p < T \\
1.0 + \frac{T - p}{|T|} & \text{if mode="lower" and } p \leq T \\
\max(0.1, \frac{T}{p}) & \text{if mode="lower" and } p > T
\end{cases}
\end{equation}

This formulation ensures that structures exceeding targets receive bonuses proportional to their over-achievement, while structures making progress toward targets receive credit proportional to their advancement.

\subsection{Training Optimization Mechanisms}

\subsubsection{Top-K Selection and Reward Focusing}

To focus learning on promising structures, we implement an adaptive top-K selection mechanism:

\begin{equation}
N_{\text{top-k}}(e) = \max\left(3, \lfloor N_{\text{batch}} \cdot r_{\text{top-k}}(e) \rfloor\right)
\end{equation}

The top-K ratio becomes more selective as training progresses:

\begin{equation}
r_{\text{top-k}}(e) = 
\begin{cases}
0.5 & \text{if } e < 100 \\
0.4 & \text{if } 100 \leq e < 200 \\
0.3 & \text{if } e \geq 200
\end{cases}
\end{equation}

For structures not selected in the top-K, we provide a reduced reward signal:

\begin{equation}
R_{\text{reduced}}(m) = \max(0.1, 0.3 \cdot R_{\text{target}}(m))
\end{equation}

\subsubsection{Reward Normalization and Stability}

To prevent gradient explosion while preserving signal relationships, we apply conditional normalization only when rewards exceed reasonable bounds:

\begin{equation}
R_{\text{normalized}}(r) = 
\begin{cases}
\mu_{\text{target}} + \frac{r - \mu_R}{\sigma_R} \cdot \sigma_{\text{target}} & \text{if } \mu_R > 100 \text{ or } \sigma_R > 50 \\
r & \text{otherwise}
\end{cases}
\end{equation}

where $\mu_{\text{target}} = 20.0$ and $\sigma_{\text{target}} = 10.0$ are conservative normalization parameters.

A minimum reward guarantee ensures gradient flow:

\begin{equation}
R_{\text{final}}(m) = \max(0.1, R_{\text{normalized}}(m))
\end{equation}

\subsection{Implementation Parameters}

\begin{table}[h!]
\centering
\caption{Implemented Reward Function Parameters}
\label{tab:reward_parameters_implemented}
\begin{tabular}{lcc}
\hline
\textbf{Parameter} & \textbf{Value} & \textbf{Description} \\
\hline
\multicolumn{3}{l}{\textbf{Core Component Weights}} \\
Base target multiplier ($\beta_{\text{base}}$) & 3.0 & Fixed multiplier for target rewards \\
Individual bonus scaling factors & 0.1 & Applied separately to novelty, validity, diversity \\
Reduced reward factor ($\alpha_{\text{reduced}}$) & 0.3 & For non-top-K structures \\
\hline
\multicolumn{3}{l}{\textbf{Validity/Novelty Multipliers}} \\
Valid structure multiplier & 1.1 & Applied when structure is valid \\
Novel structure multiplier & 1.1 & Applied when structure is novel \\
\hline
\multicolumn{3}{l}{\textbf{Target Proximity Rewards}} \\
Excellent ($\delta_{\text{rel}} \leq 0.05$) & 15.0 & Highest base reward tier \\
Very good ($0.05 < \delta_{\text{rel}} \leq 0.1$) & 12.0 & Second reward tier \\
Good ($0.1 < \delta_{\text{rel}} \leq 0.2$) & 8.0 & Third reward tier \\
Moderate ($0.2 < \delta_{\text{rel}} \leq 0.5$) & 4.0 & Fourth reward tier \\
Achievement direction bonus & 1.3 & Target achieved in correct direction \\
Close to target bonus & 1.1 & Within 20\% of target \\
Wrong direction penalty & 0.95 & Mild penalty for opposite direction \\
\hline
\multicolumn{3}{l}{\textbf{Diversity Component Weights}} \\
Batch diversity ($w_b$) & 0.30 & Immediate batch-level variety \\
N-gram diversity ($w_n$) & 0.25 & Structural pattern diversity \\
Historical diversity ($w_h$) & 0.35 & Prevention of long-term cycling \\
Compositional diversity ($w_c$) & 0.10 & Chemical building block variety \\
Exact duplicate penalty & 0.1 & Heavy penalty for exact matches \\
\hline
\multicolumn{3}{l}{\textbf{Global Memory and Selection}} \\
Global memory size ($N_{\text{mem}}$) & 200 & Maximum stored structures \\
Early top-K ratio (epochs $< 100$) & 0.5 & Top 50\% selection \\
Mid top-K ratio (100-200) & 0.4 & Top 40\% selection \\
Late top-K ratio ($\geq 200$) & 0.3 & Top 30\% selection \\
Minimum top-K count & 3 & Ensures gradient flow \\
\hline
\multicolumn{3}{l}{\textbf{Stability and Normalization}} \\
Generation history size & 500 & For diversity calculation \\
N-gram size & 4 & Character subsequence length \\
Normalization mean threshold & 100.0 & Triggers normalization \\
Normalization std threshold & 50.0 & Triggers normalization \\
Target normalization mean & 20.0 & Conservative rescaling target \\
Target normalization std & 10.0 & Conservative rescaling spread \\
Minimum reward guarantee & 0.1 & Prevents zero rewards \\
\hline
\end{tabular}
\end{table}

The parameter values as shown in Table \ref{tab:reward_parameters_implemented} represent a balance between target achievement, chemical validity, structural diversity, and training stability, enabling successful optimization across MOF design challenges while maintaining computational efficiency and training stability.

\section{Generation and Evaluation}

\subsection{Model Inference Pipelines}

Our framework provides two distinct inference pathways tailored to the specific characteristics of fine-tuned and reinforcement learning-optimized models. Each pipeline incorporates specialized generation strategies, validation protocols, and evaluation metrics designed to maximize the effectiveness of the respective model types.

\subsubsection{Fine-tuned Model Inference}

The fine-tuned model inference pipeline focuses on generating diverse MOF structures while maintaining chemical validity and providing accurate property predictions. The generation process begins with the [BOS] token and employs autoregressive sampling to produce complete MOF sequences:

\begin{equation}
\text{Sequence} = \{[BOS], t_1, t_2, \ldots, t_T, [EOS]\}
\end{equation}

where each token $t_i$ is sampled from the probability distribution:

\begin{equation}
P(t_i | t_{<i}) = \text{softmax}\left(\frac{f_\theta(t_{<i})}{\tau}\right)
\end{equation}

where $f_\theta$ represents the fine-tuned model's output logits, $\tau$ is the temperature parameter controlling sampling diversity, and $t_{<i}$ denotes all previous tokens in the sequence.

The fine-tuned inference pipeline implements several key components:

\textbf{Sequence Generation}: We employ nucleus sampling (top-p) combined with temperature scaling to balance exploration and coherence. The generation continues until either the [EOS] token is produced or the maximum sequence length (512 tokens) is reached. Multiple sequences are generated in parallel to improve efficiency and provide diverse candidates.

\textbf{Property Prediction}: For each generated MOF sequence, we leverage the same fine-tuned model to predict target properties. The sequence is re-tokenized and passed through the model's regression head to obtain property estimates:

\begin{equation}
\hat{p} = f_{\text{pred}}(\text{Embed}(\text{MOF}_{\text{tokens}}))
\end{equation}

where $f_{\text{pred}}$ is the property prediction head and $\text{Embed}$ represents the token embedding and transformer encoding process.

\textbf{Validation and Filtering}: Generated MOFs undergo multi-stage validation including SMILES syntax verification using RDKit (with metal substitution for compatibility), structural completeness checks ensuring both organic and inorganic components are present, and novelty assessment against the training dataset to identify genuinely new structures.

\subsubsection{Reinforcement Learning Model Inference}

The RL model inference pipeline is specifically designed to generate MOFs that achieve target property values while maintaining diversity and chemical validity. This pipeline incorporates the sophisticated reward mechanisms developed during training to guide generation toward high-performing structures.

\textbf{Policy-Guided Generation}: The RL model operates as a learned policy $\pi_\theta$ that has been optimized to maximize expected rewards for target properties. During inference, we sample from this policy using controlled stochasticity:

\begin{equation}
a_t \sim \pi_\theta(\cdot | s_t)
\end{equation}

where $a_t$ is the selected token at step $t$ and $s_t$ is the current sequence state.

\textbf{Multi-Component Evaluation}: Each generated MOF is evaluated using the same multi-component reward function employed during training, providing comprehensive assessment across validity, novelty, diversity, and target property achievement. This evaluation helps rank generated structures and identify the most promising candidates.

\textbf{Curriculum-Informed Sampling}: The RL inference process can optionally incorporate examples from the global memory accumulated during training, using high-performing historical structures to guide generation toward promising regions of chemical space.

\textbf{Advanced Filtering Mechanisms}: Our RL inference pipeline includes optional relaxed filtering for structures with high predicted property values. This mechanism applies more lenient validity and novelty criteria to promising MOFs that might otherwise be discarded due to minor structural irregularities:

\begin{equation}
\text{Accept}(\text{MOF}) = \begin{cases}
\text{Standard\_Filter}(\text{MOF}) & \text{if } \hat{p} < \theta \cdot p_{\text{target}} \\
\text{Relaxed\_Filter}(\text{MOF}) & \text{if } \hat{p} \geq \theta \cdot p_{\text{target}}
\end{cases}
\end{equation}

where $\theta$ is a threshold factor (typically 0.8) and $p_{\text{target}}$ is the target property value.

\subsection{Generation Process Parameters}

The generation process employs carefully tuned parameters to optimize the balance between exploration and exploitation. Table \ref{tab:generation_params_detailed} provides comprehensive parameter settings for both model types.

\begin{table}[h!]
\centering
\caption{Detailed Generation Parameters for Model Inference}
\label{tab:generation_params_detailed}
\begin{tabular}{lcc}
\hline
\textbf{Parameter} & \textbf{Fine-tuned Model} & \textbf{RL-optimized Model} \\
\hline
Temperature & 0.7 & 0.7 \\
Top-k sampling & 400 & 100 \\
Top-p (nucleus) sampling & 0.9 & 0.9 \\
Beam size & 1 & 1 \\
Early stopping & True & True \\
Max sequence length & 512 & 512 \\
Batch size & 20 & 50 \\
Number of return sequences & Variable & 32 \\
\hline
\end{tabular}
\end{table}

\textbf{Temperature Control}: Both models use a temperature of 0.7, providing a balance between deterministic generation (which might lead to repetitive structures) and highly stochastic sampling (which could produce invalid sequences). This temperature was selected through empirical evaluation to maximize both diversity and validity.

\textbf{Sampling Strategy Differences}: The fine-tuned model uses more permissive top-k sampling (400 tokens) to encourage exploration of diverse chemical space, while the RL model uses more restrictive top-k sampling (100 tokens) to focus on high-reward regions identified during training.

\textbf{Batch Processing}: Generation is performed in batches to improve computational efficiency while maintaining memory constraints. The RL model uses larger batch sizes to better leverage the reward-based ranking mechanisms.

\subsection{Evaluation Metrics}

Our evaluation framework employs multiple complementary metrics to assess generation quality across different dimensions of MOF design success.

\textbf{Structural Validity Rate}: The percentage of generated MOFs that pass comprehensive chemical validation:

\begin{equation}
R_{\text{validity}} = \frac{\sum_{i=1}^{N} \mathbb{1}_{\text{valid}}(m_i)}{N} \times 100\%
\end{equation}

where $\mathbb{1}_{\text{valid}}(m_i)$ indicates whether MOF $m_i$ passes all validation checks including SMILES syntax correctness, presence of both organic and inorganic components, metal coordination feasibility, and topology consistency (when applicable). We verify that generated MOFs contain appropriate ratios of organic linkers and metal nodes, ensuring chemical plausibility of the proposed structures.

\textbf{Novelty Rate}: The fraction of valid MOFs that are absent from the training dataset:

\begin{equation}
R_{\text{novelty}} = \frac{\sum_{i=1}^{N_{\text{valid}}} \mathbb{1}_{\text{novel}}(m_i)}{N_{\text{valid}}} \times 100\%
\end{equation}

\textbf{Structural Diversity}: We quantify diversity through multiple complementary measures including unique structure count, average pairwise distance within generated sets, and n-gram pattern analysis to detect repetitive motifs:

\begin{equation}
D_{\text{diversity}} = \frac{|\text{Unique}(\mathcal{M}_{\text{generated}})|}{|\mathcal{M}_{\text{generated}}|}
\end{equation}

\textbf{Target Proximity}: For structures with predicted properties, we measure proximity to target values:

\begin{equation}
S_{\text{proximity}} = \frac{1}{N} \sum_{i=1}^{N} \exp\left(-\frac{|\hat{p}_i - p_{\text{target}}|}{p_{\text{target}}}\right)
\end{equation}

\textbf{Property Distribution Analysis}: Comprehensive statistical analysis of generated property distributions including mean, median, standard deviation, and comparison with training data distributions.

\subsubsection{Generation Efficiency Metrics}

\textbf{Overall Efficiency}: The fraction of generation attempts that result in valid, novel MOFs meeting target criteria:

\begin{equation}
E_{\text{overall}} = \frac{N_{\text{valid}} \cap N_{\text{novel}} \cap N_{\text{target}}}{N_{\text{attempted}}} \times 100\%
\end{equation}

\textbf{Computational Efficiency}: We track generation time per MOF, memory usage, and convergence characteristics to assess practical deployment feasibility.

\subsection{Inference-Time Enhancements}

During inference, we implement an optional relaxed filtering mechanism that applies more lenient validity and novelty criteria to MOF candidates with high predicted property values. This inference-time enhancement helps recover potentially valuable structures that might be discarded due to minor technical violations while maintaining strict standards for typical candidates.

The relaxed filtering is applied when:
\begin{equation}
\hat{p}_i \geq \theta \cdot p_{\text{target}} \quad \text{where } \theta = 0.8
\end{equation}

For such high-property candidates, the system applies:
\begin{itemize}
\item \textbf{Relaxed Validity}: Uses RDKit parsing without full sanitization, allowing recovery of chemically reasonable but technically non-standard structures
\item \textbf{Relaxed Novelty}: Permits structures with high similarity (\textgreater85\%) to training data if they represent meaningful chemical variations
\end{itemize}

This approach balances the need for chemical validity with the exploration of high-performance chemical space regions that might be missed by overly strict filtering criteria.

\section{Limitations, Future Work, and Applications}
\subsection{Limitations} 
While our approach demonstrates strong performance in generating MOFs with targeted properties, several limitations should be noted. The reinforcement learning optimization is only as good as the underlying property prediction model; inaccuracies in property prediction can lead to suboptimal targeting\cite{WANG2022107739, doi:10.1021/acs.jcim.1c00191}. Our approach generates MOFs in the form of MOFid representations, which capture chemical composition and topology but not detailed 3D coordinates; a separate step is required to generate the full 3D structure, which may introduce additional uncertainty\cite{doi:10.1021/acs.cgd.9b01050, doi:10.1021/jacs.2c11420}. While our model generates chemically valid structures, it does not explicitly consider synthetic accessibility or stability under realistic conditions\cite{NANDY20231585, Butova_2016}. The model's generations are influenced by the distribution of structures in the training data, which may limit exploration of novel regions of chemical space\cite{doi:10.1021/acs.jcim.7b00690}. The reinforcement learning training process is computationally intensive, requiring significant GPU resources and time\cite{doi:10.1126/sciadv.aap7885}.

\subsection{Future Work} 
Based on the limitations identified, several directions for future work emerge. Integrating more sophisticated property prediction models, such as graph neural networks or equivariant neural networks, could improve targeting accuracy\cite{doi:10.1021/acs.jcim.3c02070, PARK20242355}. Extending the framework to directly generate 3D structures would eliminate the need for a separate structure generation step and potentially improve property prediction accuracy\cite{fu2024mofdiff}. Adding synthetic accessibility and stability criteria to the reward function could lead to more practically useful MOF candidates\cite{doi:10.1021/jacs.2c11420, NANDY20231585}. Implementing an active learning loop that iteratively refines the property prediction model based on feedback from high-fidelity simulations or experiments would improve accuracy over time\cite{doi:10.1021/acs.jcim.3c02070}. Exploring more sample-efficient reinforcement learning algorithms, such as proximal policy optimization (PPO) or soft actor-critic (SAC), could reduce computational requirements\cite{Schulman2017, Haarnoja2018}. Extending the framework to simultaneously optimize multiple properties with potentially conflicting objectives would address more complex design challenges\cite{park2024inverse, doi:10.1021/acsami.1c16220}.

\subsection{Potential Applications} 
The MOFGPT framework enables the targeted design of MOFs for a wide range of applications. For gas storage and separation, it allows designing MOFs with optimized capacity and selectivity for CH$_4$, CO$_2$, H$_2$, and other gases, with applications in natural gas storage, carbon capture, and hydrogen economy\cite{https://doi.org/10.1002/adma.201601133, C4CS00103F, doi:10.1126/science.1230444}. In catalysis, the framework can generate MOFs with specific band gaps and active sites for heterogeneous catalysis, photocatalysis, and electrocatalysis\cite{JIAO201943, CHEN201957}. For sensing applications, it can create MOFs with tailored guest-host interactions for chemical sensing and detection of specific analytes\cite{C4CS00032C}. In drug delivery, the system can develop biocompatible MOFs with controlled pore sizes and functionalities for medical applications\cite{Allendorf2015229}. For electronic and optical materials, our approach can design MOFs with specific electronic and optical properties for applications in optoelectronics, semiconductors, and energy conversion\cite{SAFAEI2019401}.

\begin{appendices}

\section{Appendix: Additional Property Optimization Results}

This appendix presents comprehensive results for reinforcement learning-based MOF generation across eight additional property domains: four CH$_4$ adsorption conditions at different pressures and four CO$_2$ adsorption conditions at different pressures. The results follow the same evaluation framework as the main text, demonstrating the robustness and versatility of our approach across diverse operating conditions. It should be noteworthy again that the finetuned distributions correspond to completely invalid generations.

\subsection{Methane Adsorption Optimization at Multiple Pressures}

We evaluated our RL approach for CH$_4$ adsorption optimization at five pressure conditions: 0.05, 0.5, 0.9, 2.5, and 4.5 bar.

\subsubsection{CH$_4$ Adsorption at 0.05 bar}

\begin{table}[H]
\centering
\caption{Performance Metrics for RL-Based MOF Generation - CH$_4$ Adsorption at 0.05 bar}
\label{tab:ch4_p1_performance}
\begin{tabular}{lccc}
\toprule
\textbf{Target} & \textbf{Validity (\%)} & \textbf{Novelty (\%)} & \textbf{Diversity (\%)} \\
\midrule
Mean & 49.2 & 90.9 & 97 \\
Mean + 1$\sigma$ & 39.5 & 83.33 & 97 \\
Mean + 2$\sigma$ & 58.3 & 85.71 & 97 \\
\bottomrule
\end{tabular}
\end{table}

\begin{table}[H]
\centering
\caption{Statistical Properties of Generated Structures - CH$_4$ Adsorption at 0.05 bar}
\label{tab:ch4_p1_stats}
\begin{tabular}{lcccc}
\toprule
\textbf{Dataset} & \textbf{Mean} & \textbf{Std Dev} \\
\midrule
Original Data & 0.066 & 0.101  \\
Fine-tuned & 1.131 & 0.730 \\
RL (Mean) & 0.044 & 0.054  \\
RL (Mean + 1$\sigma$) & 0.109 & 0.127  \\
RL (Mean + 2$\sigma$) & 0.162 & 0.243 \\
\bottomrule
\end{tabular}
\end{table}

\begin{figure}[H]
    \centering
    \includegraphics[width=0.8\textwidth]{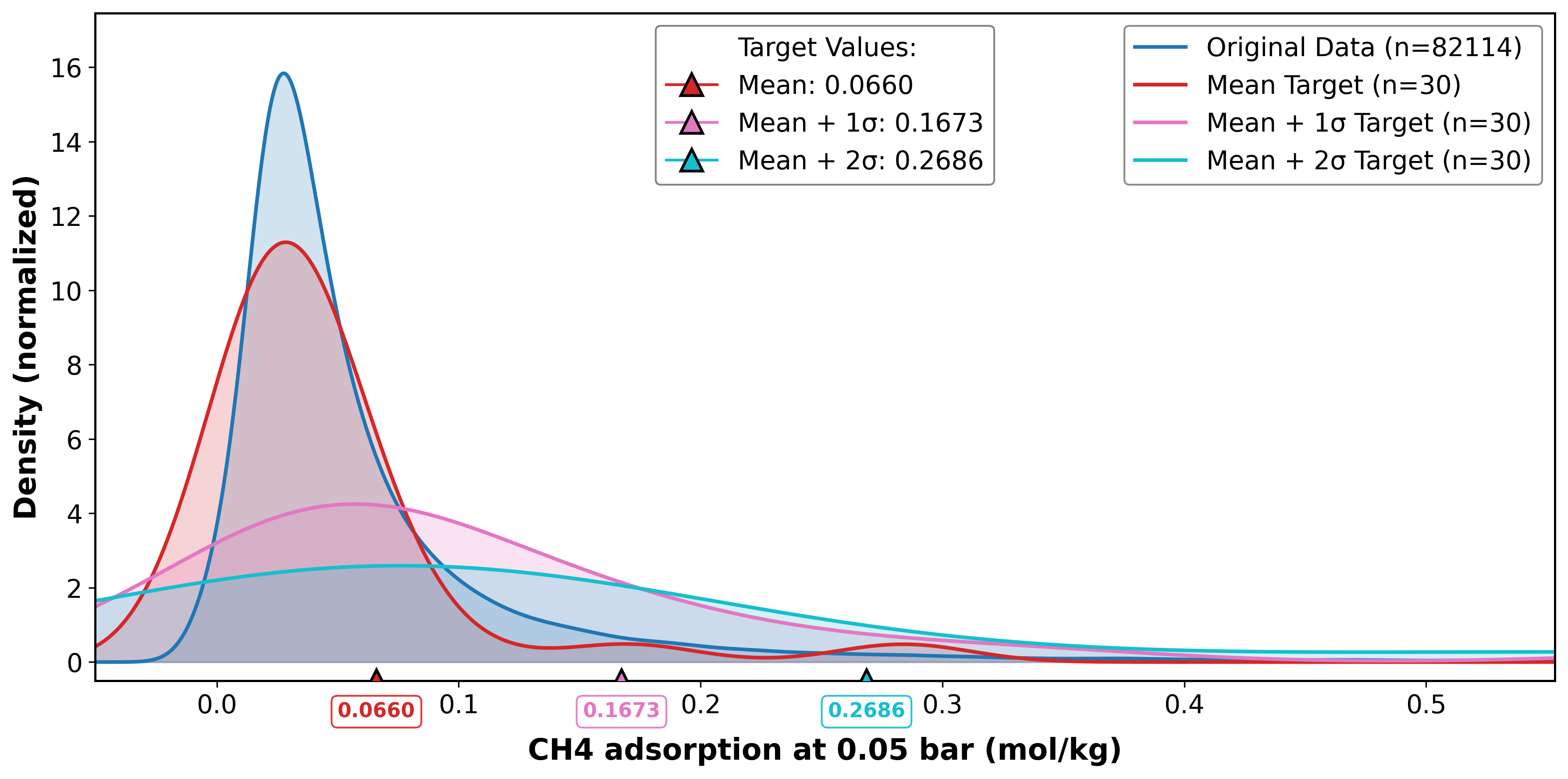}
    \caption{Normalized density distributions of CH$_4$ adsorption at 0.05 bar showing systematic targeting across mean, mean + 1$\sigma$, and mean + 2$\sigma$ scenarios.}
    \label{fig:ch4_p1_distribution}
\end{figure}

\subsubsection{CH$_4$ Adsorption at 0.5 bar}

\begin{table}[H]
\centering
\caption{Performance Metrics for RL-Based MOF Generation - CH$_4$ Adsorption at 0.5 bar}
\label{tab:ch4_p5_performance}
\begin{tabular}{lccc}
\toprule
\textbf{Target} & \textbf{Validity (\%)} & \textbf{Novelty (\%)} & \textbf{Diversity (\%)} \\
\midrule
Mean & 89.13 & 73.17 & 100 \\
Mean + 1$\sigma$ & 55.2 & 71.4 & 89.0 \\
Mean + 2$\sigma$ & 20.35 & 88.32 & 99.5 \\
\bottomrule
\end{tabular}
\end{table}

\begin{table}[H]
\centering
\caption{Statistical Properties of Generated Structures - CH$_4$ Adsorption at 0.5 bar}
\label{tab:ch4_p5_stats}
\begin{tabular}{lcccc}
\toprule
\textbf{Dataset} & \textbf{Mean} & \textbf{Std Dev} \\
\midrule
Original Data & 0.507 & 0.449 \\
Fine-tuned & 0.177 & 0.439 \\
RL (Mean) & 0.824 & 0.570 \\
RL (Mean + 1$\sigma$) & 0.732 & 0.581 \\
RL (Mean + 2$\sigma$) & 0.441 & 0.256 \\
\bottomrule
\end{tabular}
\end{table}

\begin{figure}[H]
    \centering
    \includegraphics[width=0.8\textwidth]{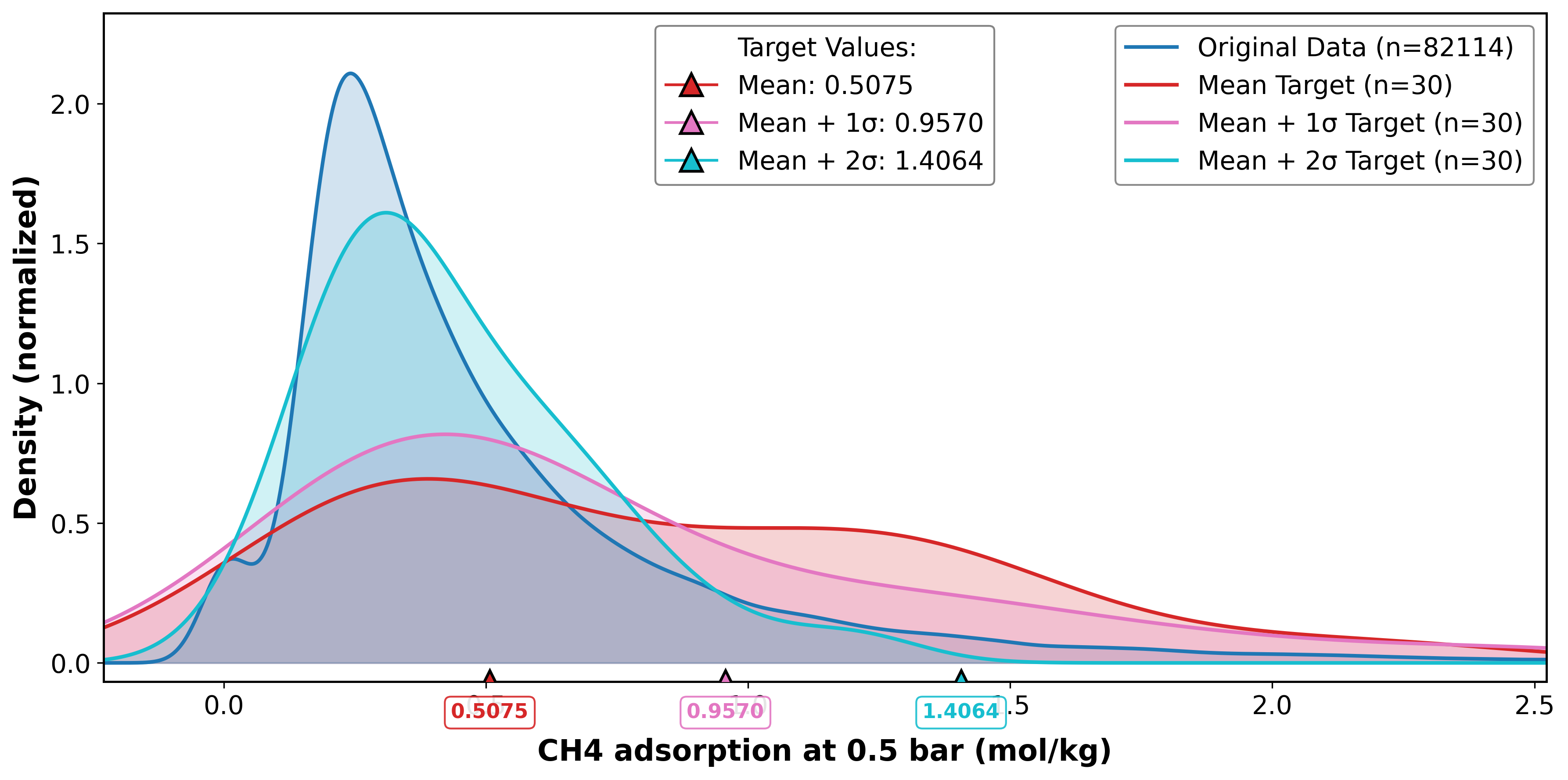}
    \caption{Normalized density distributions of CH$_4$ adsorption at 0.5 bar.}
    \label{fig:ch4_p5_distribution}
\end{figure}

\subsubsection{CH$_4$ Adsorption at 0.9 bar}

\begin{table}[H]
\centering
\caption{Performance Metrics for RL-Based MOF Generation - CH$_4$ Adsorption at 0.9 bar}
\label{tab:ch4_p2_performance}
\begin{tabular}{lccc}
\toprule
\textbf{Target} & \textbf{Validity (\%)} & \textbf{Novelty (\%)} & \textbf{Diversity (\%)} \\
\midrule
Mean & 71.21 & 63.82 & 100 \\
Mean + 1$\sigma$ & 52 & 76.92 & 99 \\
Mean + 2$\sigma$ & 53.42 & 76.92 & 97 \\
\bottomrule
\end{tabular}
\end{table}

\begin{table}[H]
\centering
\caption{Statistical Properties of Generated Structures - CH$_4$ Adsorption at 0.9 bar}
\label{tab:ch4_p2_stats}
\begin{tabular}{lcccc}
\toprule
\textbf{Dataset} & \textbf{Mean} & \textbf{Std Dev} \\
\midrule
Original Data & 0.818 & 0.623 \\
Fine-tuned & 0.608 & 0.769 \\
RL (Mean) & 1.224 & 0.745 \\
RL (Mean + 1$\sigma$) & 1.342 & 0.760  \\
RL (Mean + 2$\sigma$) & 1.834 & 1.004 \\
\bottomrule
\end{tabular}
\end{table}

\begin{figure}[H]
    \centering
    \includegraphics[width=0.8\textwidth]{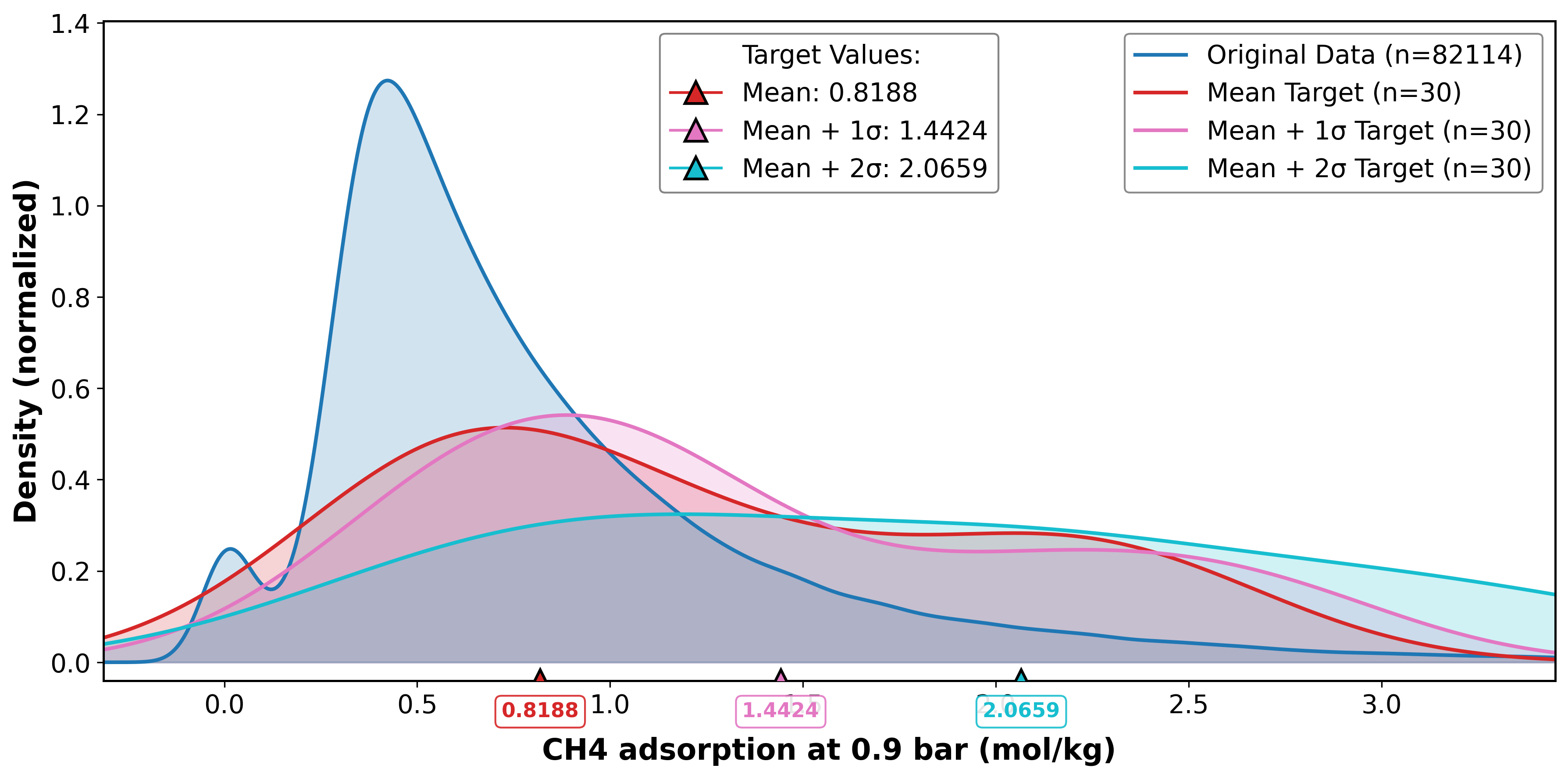}
    \caption{Normalized density distributions of CH$_4$ adsorption at 0.9 bar.}
    \label{fig:ch4_p2_distribution}
\end{figure}

\subsubsection{CH$_4$ Adsorption at 2.5 bar}

\begin{table}[H]
\centering
\caption{Performance Metrics for RL-Based MOF Generation - CH$_4$ Adsorption at 2.5 bar}
\label{tab:ch4_p3_performance}
\begin{tabular}{lccc}
\toprule
\textbf{Target} & \textbf{Validity (\%)} & \textbf{Novelty (\%)} & \textbf{Diversity (\%)} \\
\midrule
Mean & 82.35 & 53.57 & 100 \\
Mean + 1$\sigma$ & 56 & 71.42 & 100 \\
Mean + 2$\sigma$ & 64.28 & 86.11 & 99 \\
\bottomrule
\end{tabular}
\end{table}

\begin{table}[H]
\centering
\caption{Statistical Properties of Generated Structures - CH$_4$ Adsorption at 2.5 bar}
\label{tab:ch4_p3_stats}
\begin{tabular}{lcccc}
\toprule
\textbf{Dataset} & \textbf{Mean} & \textbf{Std Dev} \\
\midrule
Original Data & 1.788 & 1.044  \\
Fine-tuned &  1.421 & 1.391 \\
RL (Mean) & 2.447 & 1.089  \\
RL (Mean + 1$\sigma$) & 2.149 & 1.142  \\
RL (Mean + 2$\sigma$) & 2.568 & 1.152 \\
\bottomrule
\end{tabular}
\end{table}

\begin{figure}[H]
    \centering
    \includegraphics[width=0.8\textwidth]{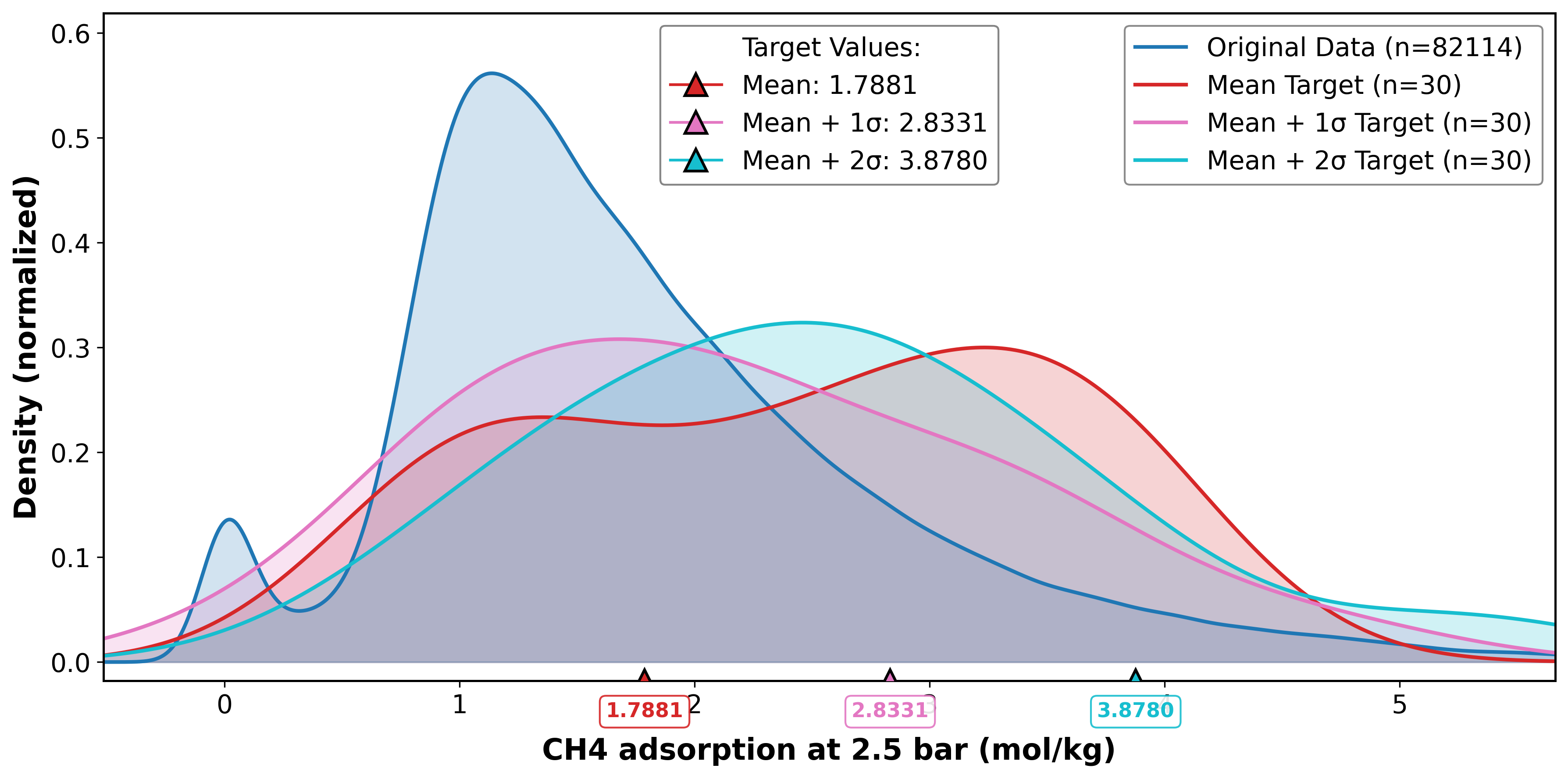}
    \caption{Normalized density distributions of CH$_4$ adsorption at 2.5 bar.}
    \label{fig:ch4_p3_distribution}
\end{figure}

\subsubsection{CH$_4$ Adsorption at 4.5 bar}

\begin{table}[H]
\centering
\caption{Performance Metrics for RL-Based MOF Generation - CH$_4$ Adsorption at 4.5 bar}
\label{tab:ch4_p4_performance}
\begin{tabular}{lccc}
\toprule
\textbf{Target} & \textbf{Validity (\%)} & \textbf{Novelty (\%)} & \textbf{Diversity (\%)} \\
\midrule
Mean & 23.94 & 58.82 & 100 \\
Mean + 1$\sigma$ & 62.33 & 62.5 & 99 \\
Mean + 2$\sigma$ & 60.52 & 65.21 & 98 \\
\bottomrule
\end{tabular}
\end{table}

\begin{table}[H]
\centering
\caption{Statistical Properties of Generated Structures - CH$_4$ Adsorption at 4.5 bar}
\label{tab:ch4_p4_stats}
\begin{tabular}{lcccc}
\toprule
\textbf{Dataset} & \textbf{Mean} & \textbf{Std Dev} \\
\midrule
Original Data & 2.708 & 1.377  \\
Fine-tuned & 0.530 & 0.882 \\
RL (Mean) & 2.485 & 1.236  \\
RL (Mean + 1$\sigma$) & 3.836 & 1.620  \\
RL (Mean + 2$\sigma$) & 3.863 & 1.225 \\
\bottomrule
\end{tabular}
\end{table}

\begin{figure}[H]
    \centering
    \includegraphics[width=0.8\textwidth]{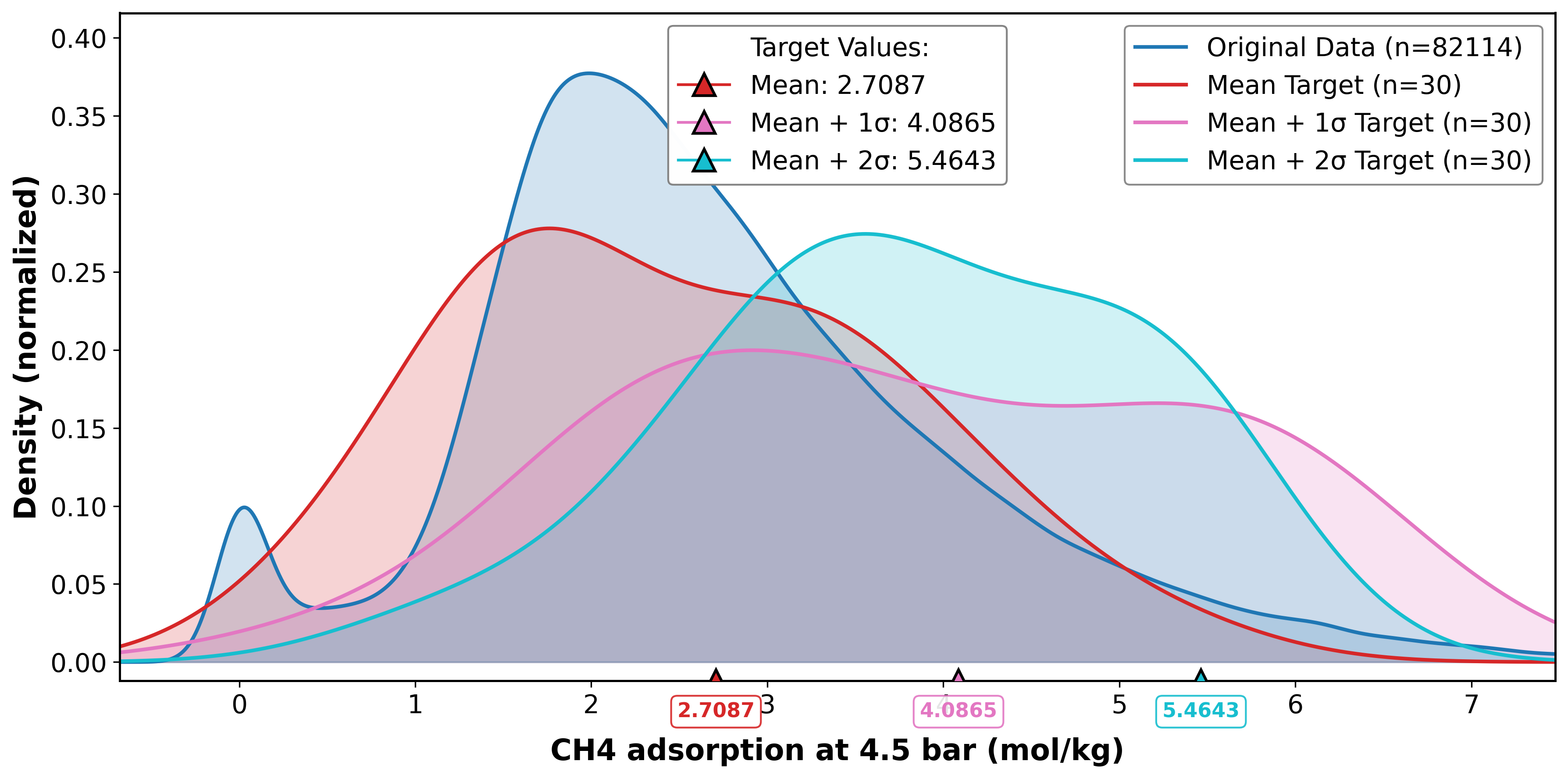}
    \caption{Normalized density distributions of CH$_4$ adsorption at 4.5 bar.}
    \label{fig:ch4_p4_distribution}
\end{figure}

\subsection{Carbon Dioxide Adsorption Optimization at Multiple Pressures}

Similarly, we evaluated CO$_2$ adsorption optimization at five pressure conditions: 0.01, 0.05, 0.1, 0.5 and 2.5 bar.

\subsubsection{CO$_2$ Adsorption at 0.01 bar}

\begin{table}[H]
\centering
\caption{Performance Metrics for RL-Based MOF Generation - CO$_2$ Adsorption at 0.01 bar}
\label{tab:co2_p1_performance}
\begin{tabular}{lccc}
\toprule
\textbf{Target} & \textbf{Validity (\%)} & \textbf{Novelty (\%)} & \textbf{Diversity (\%)} \\
\midrule
Mean & 54.23 & 93.75 & 100 \\
Mean + 1$\sigma$ & 44.87 & 88.57 & 99 \\
Mean + 2$\sigma$ & 58.82 & 100 & 99 \\
\bottomrule
\end{tabular}
\end{table}

\begin{table}[H]
\centering
\caption{Statistical Properties of Generated Structures - CO$_2$ Adsorption at 0.01 bar}
\label{tab:co2_p1_stats}
\begin{tabular}{lcccc}
\toprule
\textbf{Dataset} & \textbf{Mean} & \textbf{Std Dev} \\
\midrule
Original Data & 0.102 & 0.218  \\
Fine-tuned & 2.849 & 0.521 \\
RL (Mean) & 0.228 & 0.466  \\
RL (Mean + 1$\sigma$) & 0.052 & 0.048  \\
RL (Mean + 2$\sigma$) & 0.142 & 0.174 \\
\bottomrule
\end{tabular}
\end{table}

\begin{figure}[H]
    \centering
    \includegraphics[width=0.8\textwidth]{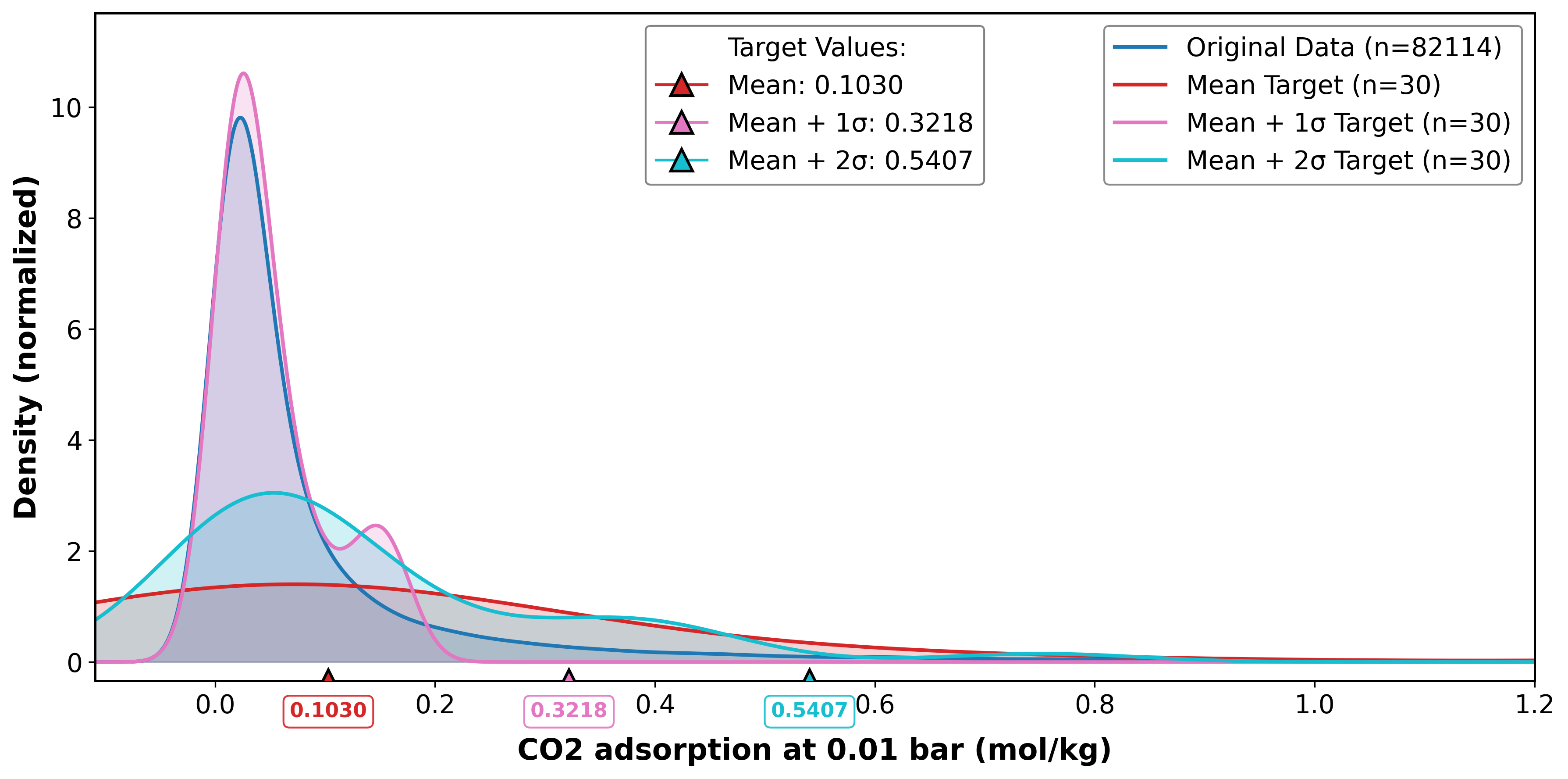}
    \caption{Normalized density distributions of CO$_2$ adsorption at 0.01 bar.}
    \label{fig:co2_p1_distribution}
\end{figure}

\subsubsection{CO$_2$ Adsorption at 0.05 bar}

\begin{table}[H]
\centering
\caption{Performance Metrics for RL-Based MOF Generation - CO$_2$ Adsorption at 0.05 bar}
\label{tab:co2_p2_performance}
\begin{tabular}{lccc}
\toprule
\textbf{Target} & \textbf{Validity (\%)} & \textbf{Novelty (\%)} & \textbf{Diversity (\%)} \\
\midrule
Mean & 90 & 83.33 & 100 \\
Mean + 1$\sigma$ & 43.90 & 88.88 & 86 \\
Mean + 2$\sigma$ & 70.21 & 90.90 & 100 \\
\bottomrule
\end{tabular}
\end{table}

\begin{table}[H]
\centering
\caption{Statistical Properties of Generated Structures - CO$_2$ Adsorption at 0.05 bar}
\label{tab:co2_p2_stats}
\begin{tabular}{lcccc}
\toprule
\textbf{Dataset} & \textbf{Mean} & \textbf{Std Dev} \\
\midrule
Original Data & 0.358 & 0.488  \\
Fine-tuned & 1.023 & 2.452 \\
RL (Mean) & 0.334 & 0.397  \\
RL (Mean + 1$\sigma$) & 0.814 & 0.765  \\
RL (Mean + 2$\sigma$) & 0.468 & 0.345 \\
\bottomrule
\end{tabular}
\end{table}

\begin{figure}[H]
    \centering
    \includegraphics[width=0.8\textwidth]{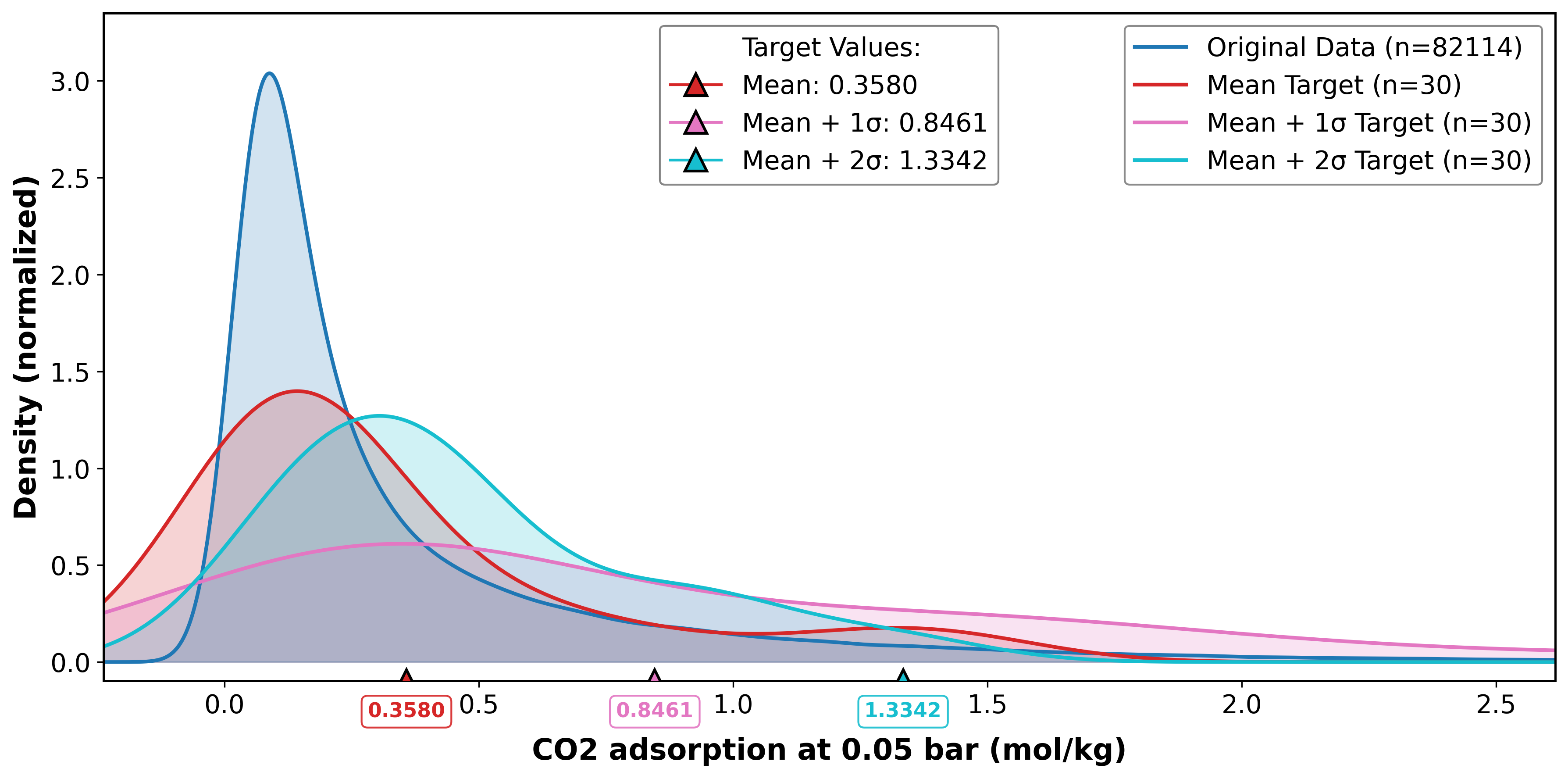}
    \caption{Normalized density distributions of CO$_2$ adsorption at 0.05 bar.}
    \label{fig:co2_p2_distribution}
\end{figure}

\subsubsection{CO$_2$ Adsorption at 0.1 bar}

\begin{table}[H]
\centering
\caption{Performance Metrics for RL-Based MOF Generation - CO$_2$ Adsorption at 0.1 bar}
\label{tab:co2_p3_performance}
\begin{tabular}{lccc}
\toprule
\textbf{Target} & \textbf{Validity (\%)} & \textbf{Novelty (\%)} & \textbf{Diversity (\%)} \\
\midrule
Mean & 42.85 & 83.33 & 99 \\
Mean + 1$\sigma$ & 50 & 83.33 & 99 \\
Mean + 2$\sigma$ & 100 & 100 & 100 \\
\bottomrule
\end{tabular}
\end{table}

\begin{table}[H]
\centering
\caption{Statistical Properties of Generated Structures - CO$_2$ Adsorption at 0.1 bar}
\label{tab:co2_p3_stats}
\begin{tabular}{lcccc}
\toprule
\textbf{Dataset} & \textbf{Mean} & \textbf{Std Dev} \\
\midrule
Original Data & 0.599 & 0.682 \\
Fine-tuned & -0.177 & 0.746 \\
RL (Mean) & 0.576 & 0.657 \\
RL (Mean + 1$\sigma$) & 0.637 & 0.563  \\
RL (Mean + 2$\sigma$) & 0.432 & 0.522 \\
\bottomrule
\end{tabular}
\end{table}

\begin{figure}[H]
    \centering
    \includegraphics[width=0.8\textwidth]{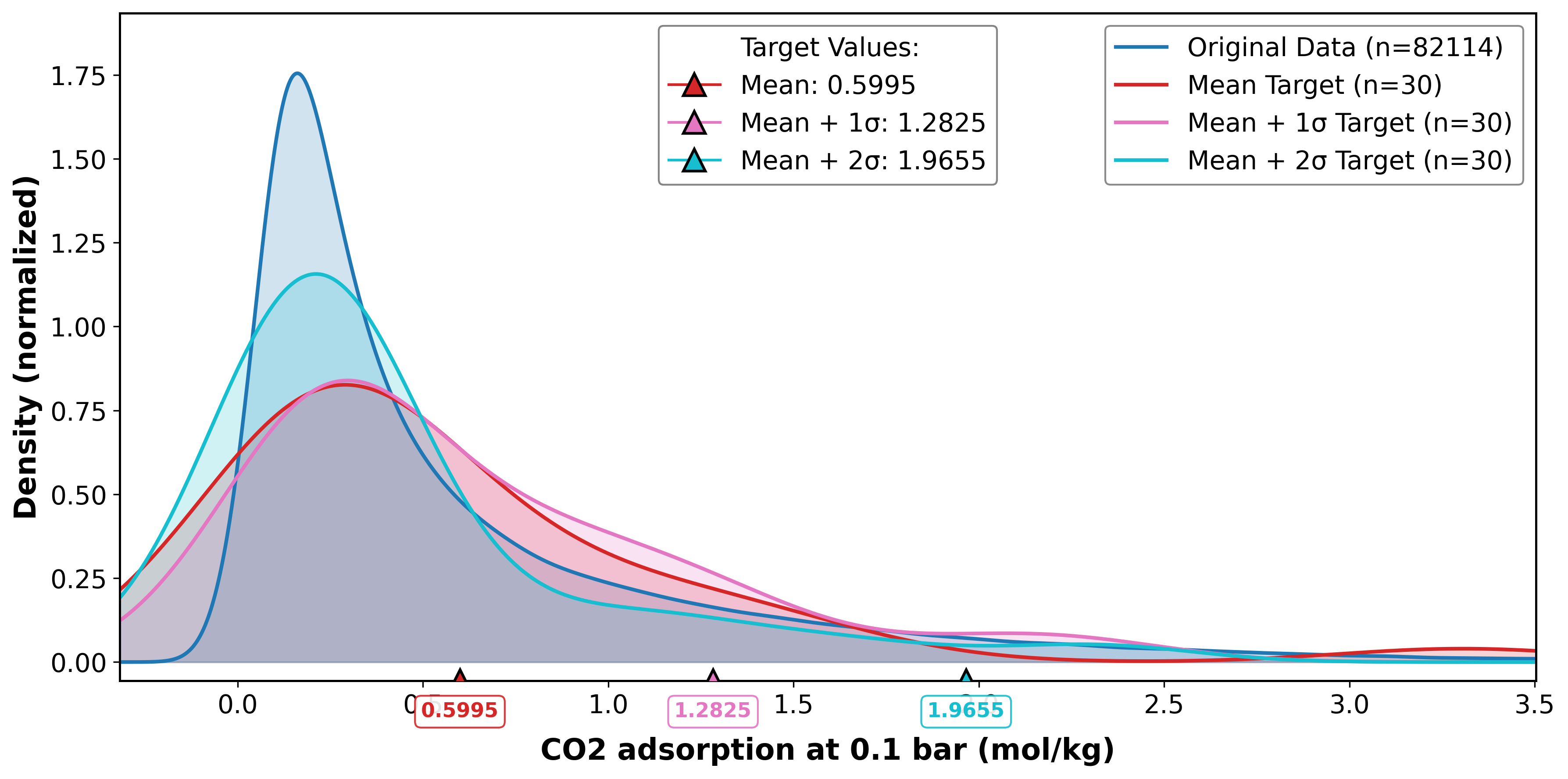}
    \caption{Normalized density distributions of CO$_2$ adsorption at 0.1 bar.}
    \label{fig:co2_p3_distribution}
\end{figure}

\subsubsection{CO$_2$ Adsorption at 0.5 bar}

\begin{table}[H]
\centering
\caption{Performance Metrics for RL-Based MOF Generation - CO$_2$ Adsorption at 0.5 bar}
\label{tab:co2_p5_performance}
\begin{tabular}{lccc}
\toprule
\textbf{Target} & \textbf{Validity (\%)} & \textbf{Novelty (\%)} & \textbf{Diversity (\%)} \\
\midrule
Mean & 95.12 & 76.9 & 100 \\
Mean + 1$\sigma$ & 65 & 76.92 & 100 \\
Mean + 2$\sigma$ & 15.81 & 81.08 & 100 \\
\bottomrule
\end{tabular}
\end{table}

\begin{table}[H]
\centering
\caption{Statistical Properties of Generated Structures - CO$_2$ Adsorption at 0.5 bar}
\label{tab:co2_p5_stats}
\begin{tabular}{lcccc}
\toprule
\textbf{Dataset} & \textbf{Mean} & \textbf{Std Dev} \\
\midrule
Original Data & 1.849 & 1.403 \\
Fine-tuned & 1.943 & 2.268 \\
RL (Mean) & 1.324 & 0.680 \\
RL (Mean + 1$\sigma$) & 2.383 & 1.117 \\
RL (Mean + 2$\sigma$) & 1.696 & 1.006 \\
\bottomrule
\end{tabular}
\end{table}

\begin{figure}[H]
    \centering
    \includegraphics[width=0.8\textwidth]{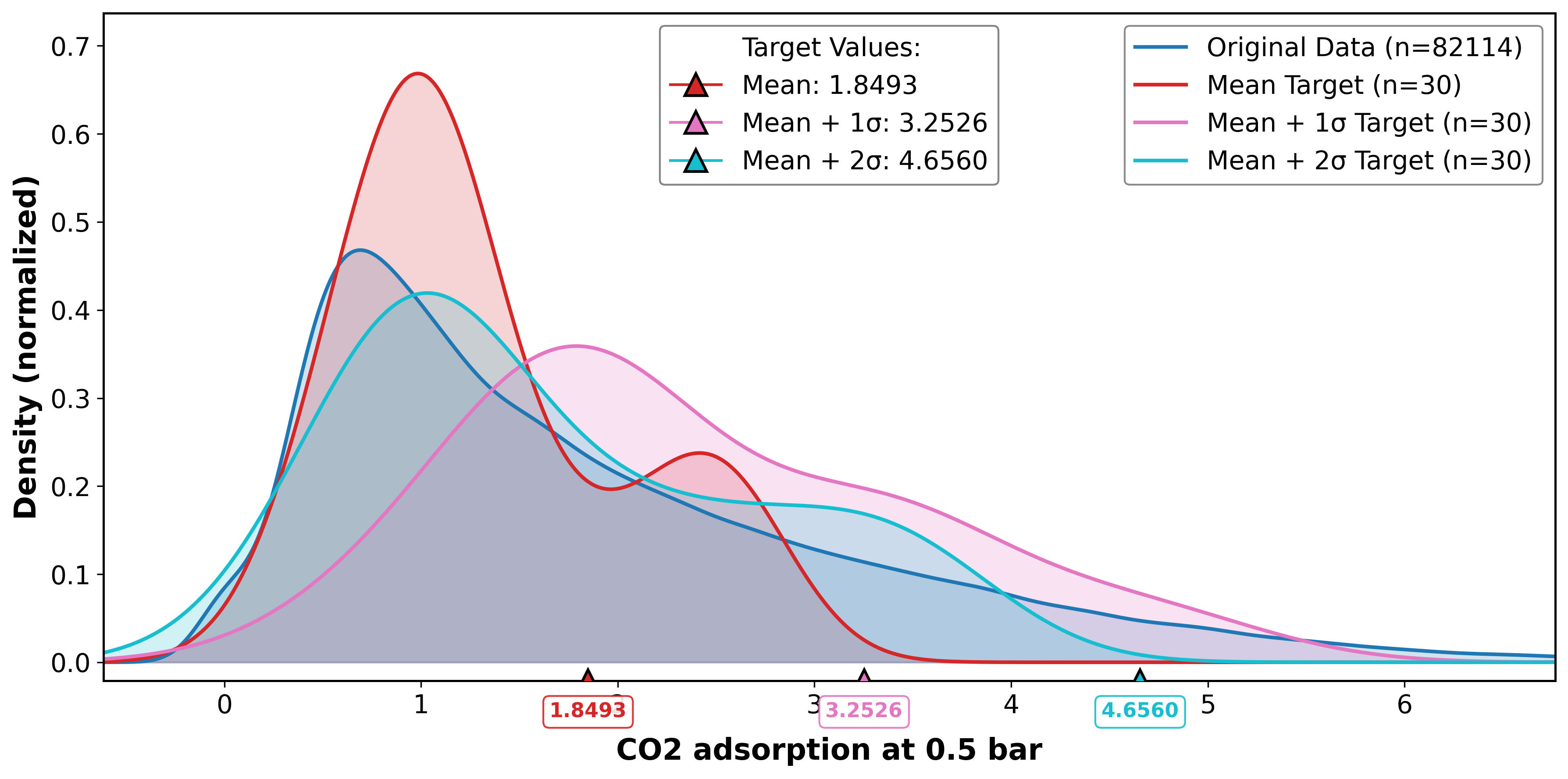}
    \caption{Normalized density distributions of CO$_2$ adsorption at 0.5 bar.}
    \label{fig:co2_p5_distribution}
\end{figure}

\subsubsection{CO$_2$ Adsorption at 2.5 bar}

\begin{table}[H]
\centering
\caption{Performance Metrics for RL-Based MOF Generation - CO$_2$ Adsorption at 2.5 bar}
\label{tab:co2_p4_performance}
\begin{tabular}{lccc}
\toprule
\textbf{Target} & \textbf{Validity (\%)} & \textbf{Novelty (\%)} & \textbf{Diversity (\%)} \\
\midrule
Mean & 57.42 & 51.72 & 99.33 \\
Mean + 1$\sigma$ & 66.55 & 62.5 & 99 \\
Mean + 2$\sigma$ & 63.63 & 85.71 & 99 \\
\bottomrule
\end{tabular}
\end{table}

\begin{table}[H]
\centering
\caption{Statistical Properties of Generated Structures - CO$_2$ Adsorption at 2.5 bar}
\label{tab:co2_p4_stats}
\begin{tabular}{lcccc}
\toprule
\textbf{Dataset} & \textbf{Mean} & \textbf{Std Dev} \\
\midrule
Original Data & 5.138 & 2.706  \\
Fine-tuned & 4.421 & 1.885 \\
RL (Mean) & 5.944 & 3.032  \\
RL (Mean + 1$\sigma$) &  6.320 & 2.535  \\
RL (Mean + 2$\sigma$) & 5.463 & 2.006 \\
\bottomrule
\end{tabular}
\end{table}

\begin{figure}[H]
    \centering
    \includegraphics[width=0.8\textwidth]{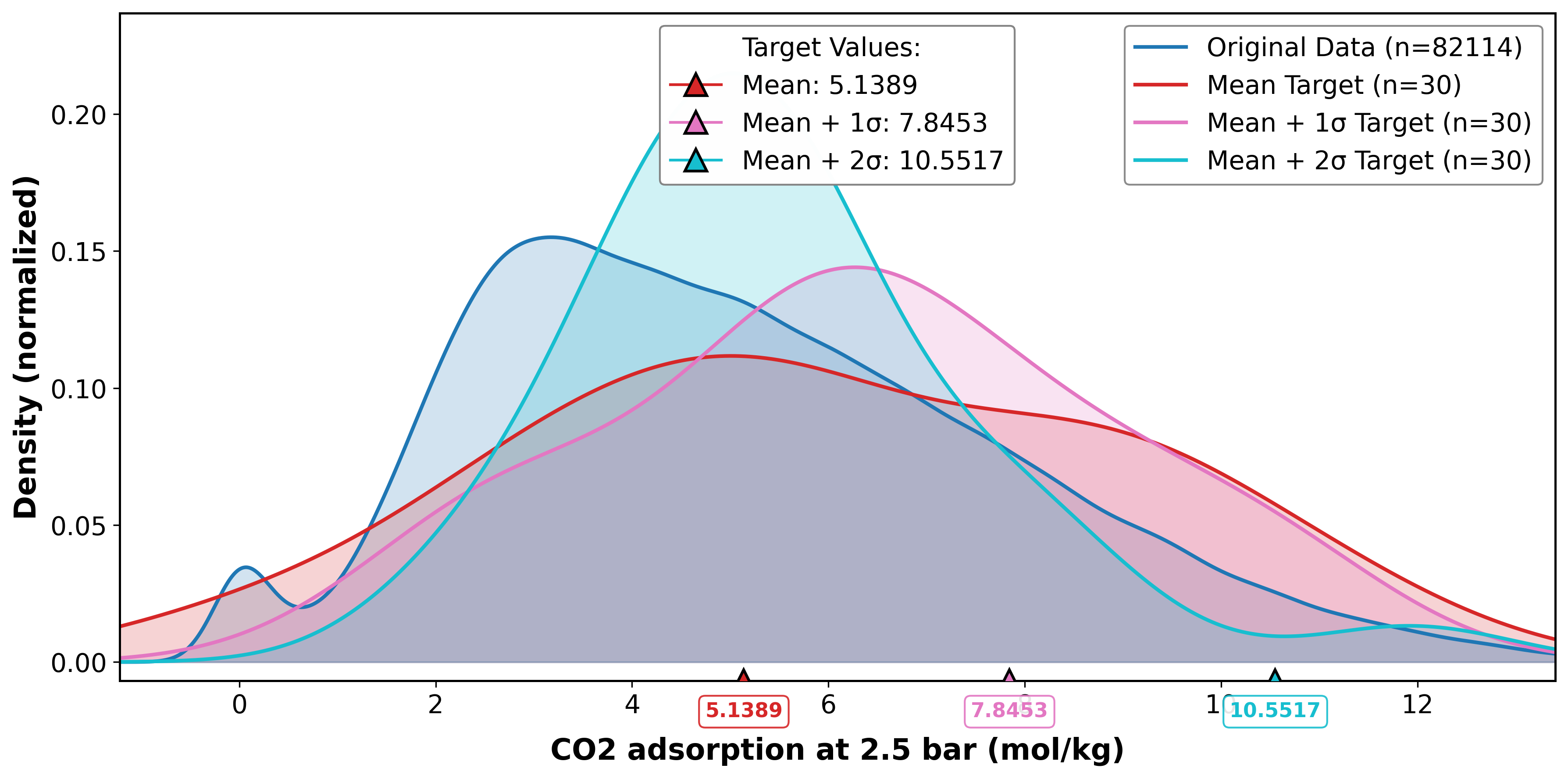}
    \caption{Normalized density distributions of CO$_2$ adsorption at 2.5 bar.}
    \label{fig:co2_p4_distribution}
\end{figure}

\subsection{Band gap (eV)}

We also evaluated band gap energy optimization (eV).

\begin{table}[H]
\centering
\caption{Performance Metrics for RL-Based MOF Generation - Band gap (eV)}
\label{tab:bandgap_performance}
\begin{tabular}{lccc}
\toprule
\textbf{Target} & \textbf{Validity (\%)} & \textbf{Novelty (\%)} & \textbf{Diversity (\%)} \\
\midrule
Mean & 100 & 93.75 & 93.75 \\
Mean + 1$\sigma$ & 64 & 93.75 & 100 \\
Mean + 2$\sigma$ & 35.63 & 100 & 83.0 \\
Mean - 1$\sigma$ & 39.75 & 90.90 & 100 \\
\bottomrule
\end{tabular}
\end{table}

\begin{table}[H]
\centering
\caption{Statistical Properties of Generated Structures - Band gap (eV)}
\label{tab:bandgap_stats}
\begin{tabular}{lcccc}
\toprule
\textbf{Dataset} & \textbf{Mean} & \textbf{Std Dev} \\
\midrule
Original Data & 1.944 & 1.027 \\
Fine-tuned & 1.961 & 0.984 \\
RL (Mean - 1$\sigma$) & 1.523 & 0.769 \\
RL (Mean) & 1.289 & 0.577 \\
RL (Mean + 1$\sigma$) & 1.767 & 0.520 \\
RL (Mean + 2$\sigma$) & 2.044 & 0.83 \\
\bottomrule
\end{tabular}
\end{table}

\begin{figure}[H]
    \centering
    \includegraphics[width=0.8\textwidth]{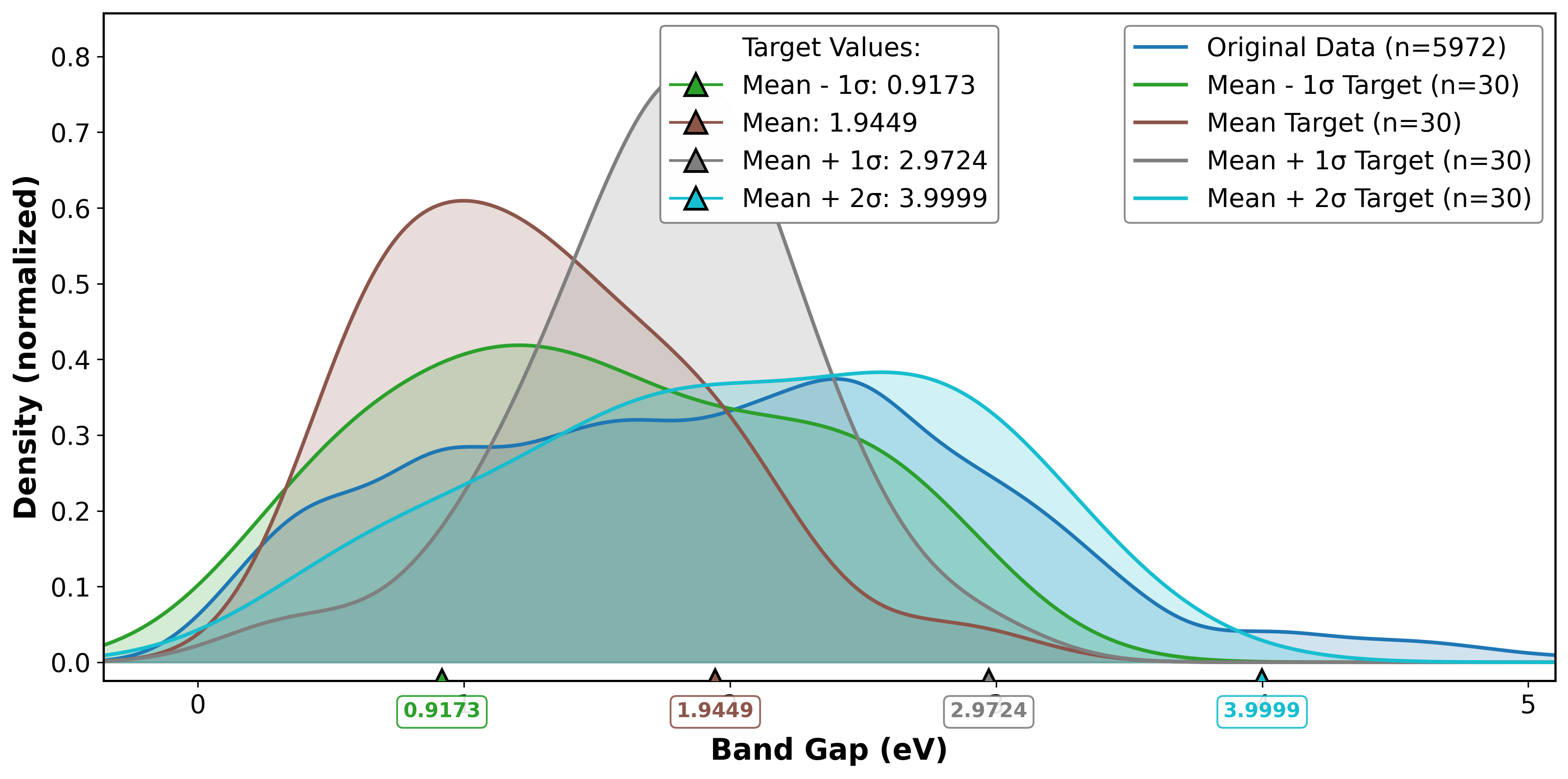}
    \caption{Normalized density distributions of Band gap (eV).}
    \label{fig:bandgap_distribution}
\end{figure}

\subsection{Summary of Multi-Pressure Performance}

\begin{table}[H]
\centering
\caption{Summary of Average Performance Metrics Across All Pressure Conditions}
\label{tab:pressure_summary}
\begin{tabular}{lcccc}
\toprule
\textbf{Gas} & \textbf{Pressure Range} & \textbf{Avg Validity (\%)} & \textbf{Avg Novelty (\%)} & \textbf{Avg Diversity (\%)} \\
\midrule
CH$_4$ & 0.05-4.5 bar & 55.8 ± 17.9 & 73.9 ± 11.1 & 98.1 ± 2.7 \\
CO$_2$ & 0.01-2.5 bar & 61.2 ± 21.2 & 83.1 ± 12.4 & 98.6 ± 3.4 \\
\bottomrule
\end{tabular}
\end{table}

The comprehensive evaluation across multiple pressure conditions demonstrates the robustness of our reinforcement learning approach for MOF property optimization. Consistent performance across different operating pressures validates the framework's applicability to diverse applications.

\end{appendices}

\bibliography{si}